%% file: neurips_2025.tex
\documentclass{article}

\PassOptionsToPackage{numbers, compress}{natbib}

 \usepackage[main, final]{neurips_2025}

\usepackage[utf8]{inputenc} 
\usepackage[T1]{fontenc}    
\usepackage{hyperref}       
\usepackage{url}            
\usepackage{booktabs}       
\usepackage{amsfonts}       
\usepackage{nicefrac}       
\usepackage{microtype}      
\RequirePackage[dvipsnames]{xcolor}
\usepackage{amssymb}
\usepackage{amsmath}
\usepackage{wrapfig}
\usepackage{algorithm}
\usepackage{algpseudocode}
\usepackage{subcaption}
\usepackage{multirow} 
\usepackage{placeins}
\usepackage{graphicx}
\usepackage{array} 
\usepackage{caption} 
\usepackage{colortbl} 
\usepackage{tabularx}
\usepackage{geometry}
\usepackage{diagbox} 
\usepackage{arydshln}
\usepackage{pifont}
\definecolor{grayblue}{rgb}{0.3, 0.75, 0.8}
\definecolor{darkblue}{rgb}{0.0, 0.0, 0.55}
\definecolor{darkred}{rgb}{0.5, 0.0, 0.0}
\definecolor{darkgreen}{rgb}{0.0, 0.4, 0.0}

\title{Spurious-Aware Prototype Refinement for\\
Reliable Out-of-Distribution Detection}

\author{%
  \textbf{Reihaneh Zohrabi}\thanks{Equal contribution}~$^1$ \quad
  \textbf{Hosein Hasani}\footnotemark[1]~$^2$ \quad
  \textbf{Mahdieh Soleymani Baghshah}$^2$ \quad
  \textbf{Anna Rohrbach}$^1$ \\
  \textbf{Marcus Rohrbach}$^1$ \quad
  \textbf{Mohammad Hossein Rohban}$^2$ \\
  $^1$TU Darmstadt \quad
  $^2$Sharif University of Technology \\
  \texttt{\{reihaneh.zohrabi, anna.rohrbach, marcus.rohrbach\}@tu-darmstadt.de} \\
  \texttt{\{hosein.hasani, soleymani, rohban\}@sharif.edu}
}

\begin{document}

\maketitle

\begin{abstract}
Out-of-distribution (OOD) detection is crucial for ensuring the reliability and safety of machine learning models in real-world applications, where they frequently face data distributions unseen during training.
Despite progress, existing methods are often vulnerable to spurious correlations that mislead models and compromise robustness. To address this, we propose SPROD, a novel prototype-based OOD detection approach that explicitly addresses the challenge posed by unknown spurious correlations.
Our post-hoc method refines class prototypes to mitigate bias from spurious features without additional data or hyperparameter tuning, and is broadly applicable across diverse backbones and OOD detection settings.
We conduct a comprehensive spurious correlation OOD detection benchmarking, comparing our method against existing approaches and demonstrating its superior performance across challenging OOD datasets, such as CelebA, Waterbirds, UrbanCars, Spurious Imagenet, and the newly introduced Animals MetaCoCo. On average, SPROD improves AUROC by 4.8\%  and FPR@95 by 9.4\% over the second best. 
\end{abstract}

\section{Introduction}
\label{sec:Introduction}

Machine learning systems in real-world applications often encounter out-of-distribution (OOD) inputs, which are samples from distributions different from the training data.
These inputs require cautious handling to prevent overconfident mispredictions during inference~\cite{nguyen2015deepneuralnetworkseasily}. This makes OOD detection crucial, as it aims to identify whether an input belongs to the known distribution or not. Yet, deep neural networks, widely used in vision tasks~\cite{DBLP:journals/corr/HeZR015, 10.1145/3065386,DBLP:journals/corr/abs-1804-02767, DBLP:journals/corr/RenHG015}, tend to make high-confidence predictions even on OOD inputs, demonstrating their inability to recognize data outside the training distribution as OOD~\cite{DBLP:journals/corr/NguyenYC14,MSP}.
The reliability of OOD detection is especially critical in applications like healthcare and autonomous driving, where overconfident predictions on unfamiliar data could have serious consequences~\cite{medical_ood, autonomous_driving}.

\begin{figure}[!t]
  \centering
  \includegraphics[width=0.72\linewidth]{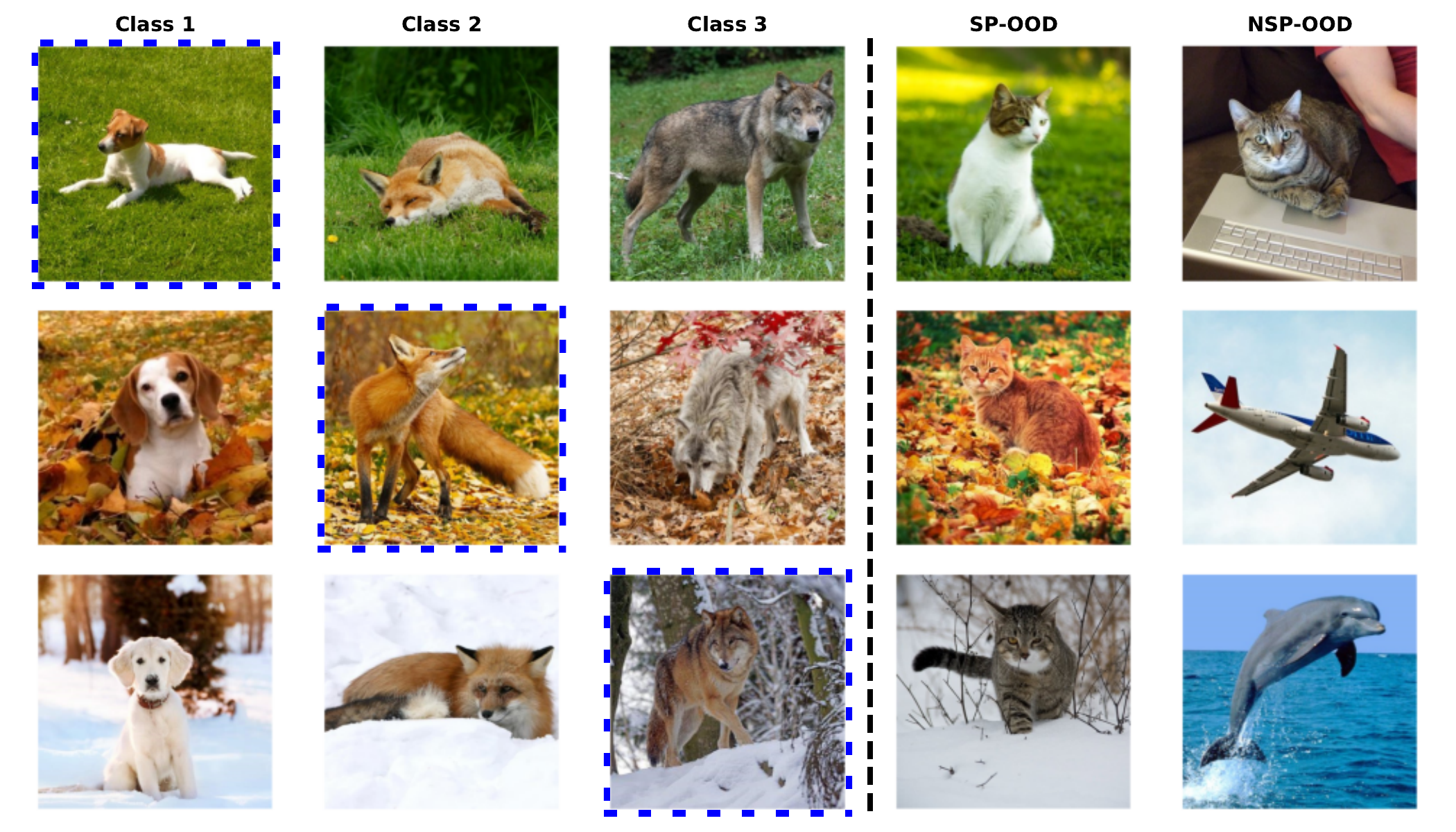}
  \vspace{-0.5mm}
   \caption{
   The challenge of spurious correlations in OOD detection. 
  ID classes (dog, fox, wolf) appear in correlated backgrounds (grass, autumn, snow), with majority groups relying on context shortcuts (blue frames). SP-OOD samples share the same contextual backgrounds, making detection more difficult. NSP-OOD samples differ in context and lack both spurious and core features.
    }
   \vspace{-1.mm}
\label{fig:intro:data_example}
\end{figure}

Recent research on OOD detection aims to ensure the reliable deployment of DNNs~\cite{MSP,ODIN,MD,OE,energy}.
Despite many effective methods, their robustness can be undermined by \emph{spurious correlations in the training data}~\cite{main_ref}. Studies indicate that models often rely on features that are statistically informative but not causally representative of the object itself~\cite{beery2018recognition,geirhos2018imagenet,DR}.
These misleading cues can act as shortcuts, allowing models to achieve high accuracy without learning the core, causally relevant features~\cite{shortcut}.
While spurious correlations have been well-explored in classification tasks~\cite{DR, Lastlayer, wg3}, their impact on OOD detection remains underexplored. 
Recently, \cite{main_ref} underscores the impact of spurious correlations on OOD detection and introduces a formalization that categorizes OOD samples into two types: spurious OODs (SP-OODs), which contain spurious attributes but lack core features, and non-spurious OODs (NSP-OODs), which lack both attributes and align with the traditional OOD setting.
Figure~\ref{fig:intro:data_example} shows an example of this problem setting.

Recent efforts to mitigate spurious correlations in OOD detection can be grouped into several broad categories. Outlier Exposure (OE) techniques reduce reliance on spurious correlations by incorporating synthetic OOD samples in their training~\cite{Ronf, kirby, imood, backmix}. Other methods focus on modifying training objectives to explicitly discourage models from depending on spurious features~\cite{rw, nsed}. However, these methods usually require retraining and generating additional OOD data. In contrast, several \emph{post-hoc} approaches have been proposed that bypass the limitations of training-heavy approaches and offer fast and light alternatives~\cite{pr, cover, MCM, dai-etal-2023-exploring, neglabel, CMA}. Despite their promise, these methods are often tested on limited synthetic datasets or specific backbones, and some rely on multiple modalities.

To address these limitations, we propose \textbf{S}purious-Aware \textbf{P}rototype \textbf{R}efinement for Reliable \textbf{O}ut-of-Distribution \textbf{D}etection (SPROD) for robust OOD detection, especially in the presence of unknown spurious correlations. It follows a three-stage process: (1) initial prototype construction, (2) classification-aware prototype calculation, and (3) group prototype refinement. SPROD can be easily applied to any pretrained feature extractor without fine-tuning on target datasets, offering a straightforward, hyperparameter-free approach that is both efficient and adaptable across diverse OOD detection tasks. Moreover, our work offers a comprehensive evaluation of OOD detection in the presence of spurious correlations, benchmarking existing methods across multiple challenging datasets, including Waterbirds~\cite{sagawa2020distributionallyrobustneuralnetworks}, CelebA~\cite{CelebA}, UrbanCars~\cite{UrbanCars}, Spurious ImageNet~\cite{neuhaus2023spuriousfeatureslargescale}, and the newly introduced Animals MetaCoCo. SPROD achieves state-of-the-art performance and consistently exhibits robust behavior across a wide range of benchmarks and experimental conditions. The main contributions of this work are as follows:

\begin{itemize}
    \item
    We propose SPROD, a post-hoc OOD detection method that directly addresses unknown spurious correlations by design and outperforms the state-of-the-art.
    We provide theoretical insight into how the proposed method mitigates spurious bias.
    \item
    SPROD is a fast, simple, and general approach applicable to diverse pretrained feature extractors and OOD detection settings.
    \item 
    SPROD does not assume access to group annotations to achieve robustness to spurious correlations and does not require either OOD or ID validation data for hyperparameter tuning. Moreover, it maintains strong performance even in low-training data regimes.
    \item 
    This work conducts and introduces comprehensive benchmarking across multiple SP-OOD datasets, including the newly introduced Animals MetaCoCo, a realistic, multiclass dataset with diverse spurious attributes.
    \item 
    Finally, our study sheds new light on key factors influencing SP-OOD detection, such as the impact of backbone fine-tuning and the choice of scoring mechanisms.
    
\end{itemize}

\section{Related Work}
\label{sec:RelatedWork}

OOD detection methods can be categorized into training-time and post-hoc approaches~\cite{openood}. Training-time methods leverage auxiliary OOD samples (Outlier Exposure)~\cite{OE,MCD, UDG} or apply regularization~\cite{devries2018learning, hendrycks2019using, huang2021mos, wei2022mitigating} to enhance OOD detection. Post-hoc methods, in contrast, derive OOD scores from base classifiers without modifying training~\cite{openood}. Overall, post-hoc methods offer simplicity and competitive performance~\cite{openood}, making them practical under limited data or training resources.
Among post-hoc methods, several approaches apply transformations to model logits to derive OOD scores. 
MSP~\cite{MSP} uses the maximum softmax probability, the energy-based method~\cite{energy} computes the log-sum-exp of logits, MLS~\cite{MLS} uses the maximum logit and introduces KL Matching (KLM) based on KL divergence, and GEN~\cite{Gen} employs generalized entropy of softmax outputs.

Another class of post-hoc methods detects OOD samples via feature-space distances. MDS~\cite{MD} fits class-conditional Gaussians to pre-logit features and computes Mahalanobis distances, refined by RMDS~\cite{RMD} with an unconditional Gaussian on ID data. KNN~\cite{knn} uses distances to nearest ID samples. SHE~\cite{SHE} scores samples by their distance to stored ID feature templates. NNGuide~\cite{NNGuide} leverages nearest-neighbor guidance to adjust test features toward the ID manifold. Relation~\cite{Relation} constructs a graph over training embeddings and detects outliers via relational anomalies. NECO~\cite{Neco} scores samples by their feature alignment with class weight vectors, leveraging neural collapse geometry. SCALE~\cite{Scale} separates ID and OOD samples by scaling penultimate-layer activations. FDBD~\cite{FDBD} measures features’ regularized mean distance to the classifier’s decision boundaries. NCI~\cite{NCI} scores samples by their distance to class weight vectors, filtered by feature norms.

Prototype-based methods shape class representations for distance-based OOD scoring. 
Classical approaches like MDS~\cite{MD} and its variants~\cite{RMD} are closely related, as they model each class by a centroid in feature space, optionally using class-conditional covariances to compute Mahalanobis distances. 
Recent works extend this via explicit training objectives.
CIDER~\cite{Cider} learns hyperspherical embeddings by jointly enforcing intra-class compactness and inter-class dispersion, thereby improving ID and OOD separability. 
PALM~\cite{palm} represents each class as a mixture of learnable prototypes and optimizes a maximum-likelihood and contrastive objective, updating prototypes and backbone features jointly during training. 
PROWL~\cite{prowl} also leverages prototype representations, but for pixel-level OOD detection in segmentation. 
While these methods share a prototypical framework, \emph{SPROD} differs as a post-hoc method operating on pretrained backbones and is explicitly designed to mitigate the negative effects of unknown spurious correlations. 
Appendix~\ref{sec:mixture_of_prototypes} further analyzes a variant, SPROD-KMeans, which connects to mixture-of-prototypes ideas while remaining fully post-hoc.

A few methods combine information from both feature and logit spaces. ReAct~\cite{ReAct} thresholds activations before applying energy-based scoring. ViM \cite{VIM} adds a virtual logit from the residual norm between input features and the ID subspace and applies softmax over extended logits.
ASH~\cite{ASH} prunes high-magnitude activations and rescales remaining features before logit computation, improving energy-based OOD separability.
Some methods also exploit gradient space for OOD scoring~\cite{gradnorm, DBLP:journals/corr/abs-2008-11600}. GradNorm~\cite{gradnorm} computes the KL divergence to a uniform distribution and uses the gradient norm (w.r.t. the penultimate layer) as the score.

Spurious correlations in training data degrade OOD detection performance, as shown in~\cite{main_ref}. Evaluations of popular methods~\cite{MSP,ODIN,MD,energy,Gram} reveal that as spurious correlations increase, detection performance drops, and SP-OOD samples become especially challenging to detect. Feature-based methods like MDS~\cite{MD} outperform others, especially for NSP samples. 
Recent work addresses spurious correlations in OOD detection through various strategies. OE methods synthesize OOD samples in ways that reduce reliance on background cues, encouraging models to focus on core semantic features~\cite{Ronf, kirby, imood, backmix}. Training-time regularization mitigates spurious cues by reweighting samples or augmenting non-semantic features~\cite{rw,nsed}. Post-hoc methods improve inference by modifying inputs to isolate semantics or reduce background influence~\cite{pr,cover}.

Recent advances in vision-language models~\cite{CLIP} have led to a category of zero-shot OOD detection methods~\cite{MCM, dai-etal-2023-exploring, neglabel, CMA} that use textual inputs, such as class names or attribute descriptions, to define ID data and identify OOD samples. While some report results on Waterbirds SP-OOD, this is not their main focus. Moreover, lacking training data with spurious correlations, their zero-shot setting does not fully capture the SP-OOD challenge. In contrast to these approaches, which rely on explicit text for OOD scoring, our method refines visual prototypes using only ID features and class labels from the training set.
A detailed review of studies regarding SP-OOD can be found in Appendix~\ref{sec:spood_literature}. 

\section{Preliminaries}

\subsection{Problem Setup}

This paper addresses Out-of-Distribution (OOD) detection under spurious correlations in a supervised classification setting. Let $\mathcal{X}$ be the input space and $\mathcal{Y} = \{1, \dots, C\}$ the label set. The in-distribution (ID) training data $\mathcal{D}_{\text{in}} = \{ (x_i, y_i) \}_{i=1}^N$ comprises samples from a joint distribution $P_{\mathcal{X}, \mathcal{Y}}$. A neural network $f_\theta$ maps each input $x_i$ to a feature embedding $h_i = f_\theta(x_i)$. This network is typically pretrained or fine-tuned on the training data. OOD detection aims to identify test samples from distributions not seen during training, including those from unseen classes.

Spurious correlations in OOD detection were first formalized by~\cite{main_ref}. According to this framework, each input can be decomposed into: 
\textbf{(i)} \textit{core features}, which are causally related to the label, and 
\textbf{(ii)} \textit{spurious features}, which are correlated with the label but not causally relevant. 
Imbalances in the training data often result in dominant core–spurious combinations (\textit{majority groups}), while rarer combinations (\textit{minority groups}) remain underrepresented. This bias encourages models to rely disproportionately on spurious cues. In controllable settings, the proportion of majority group samples within a class is captured by the \textbf{correlation rate}.
As illustrated in Figure~\ref{fig:intro:data_example}, this setup gives rise to two types of OOD examples: 
\textbf{Spurious OOD (SP-OOD)}, which share spurious features with ID data but differ in core features (e.g., a \textit{cat} on \textit{grass}, where \textit{grass} is spuriously associated with ID classes \textit{dog, wolf, fox}); and 
\textbf{Non-spurious OOD (NSP-OOD)}, which differ in both core and spurious features (e.g., a \textit{cat} on a \textit{laptop}). 

\subsection{Score Calculation}
\label{sec:motivation}

\begin{figure}[!t]
    \centering
    \begin{subfigure}[b]{0.26\textwidth}
        \centering
        \includegraphics[width=\textwidth]{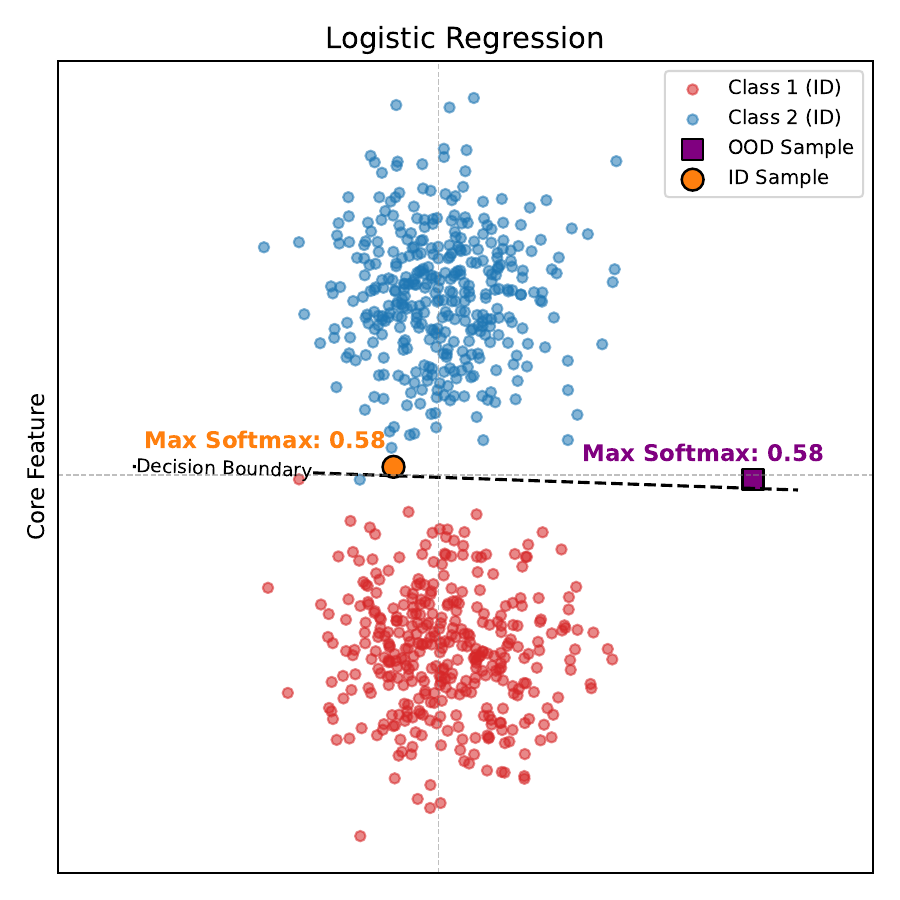}
        \vspace{-7mm}
        \caption{}
    \end{subfigure}
    \begin{subfigure}[b]{0.26\textwidth}
        \centering
        \includegraphics[width=\textwidth]{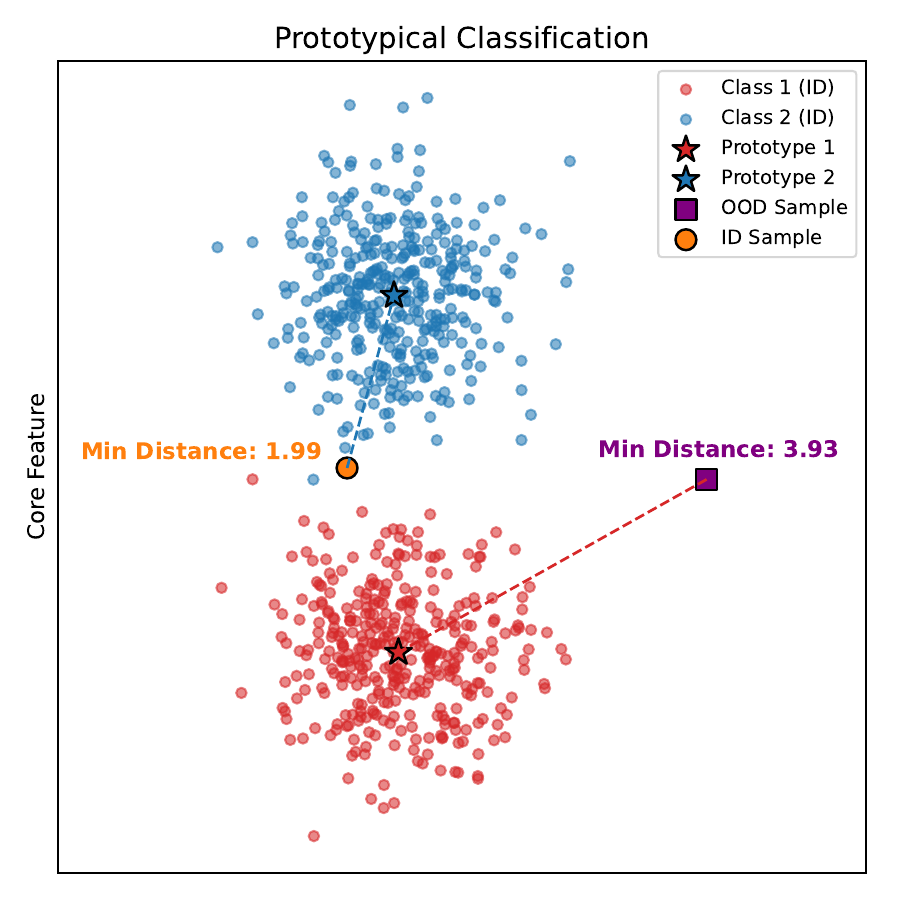}
        \vspace{-7mm}
        \caption{}
    \end{subfigure}
    \begin{subfigure}[b]{0.27\textwidth}
        \centering
        \includegraphics[width=\textwidth]{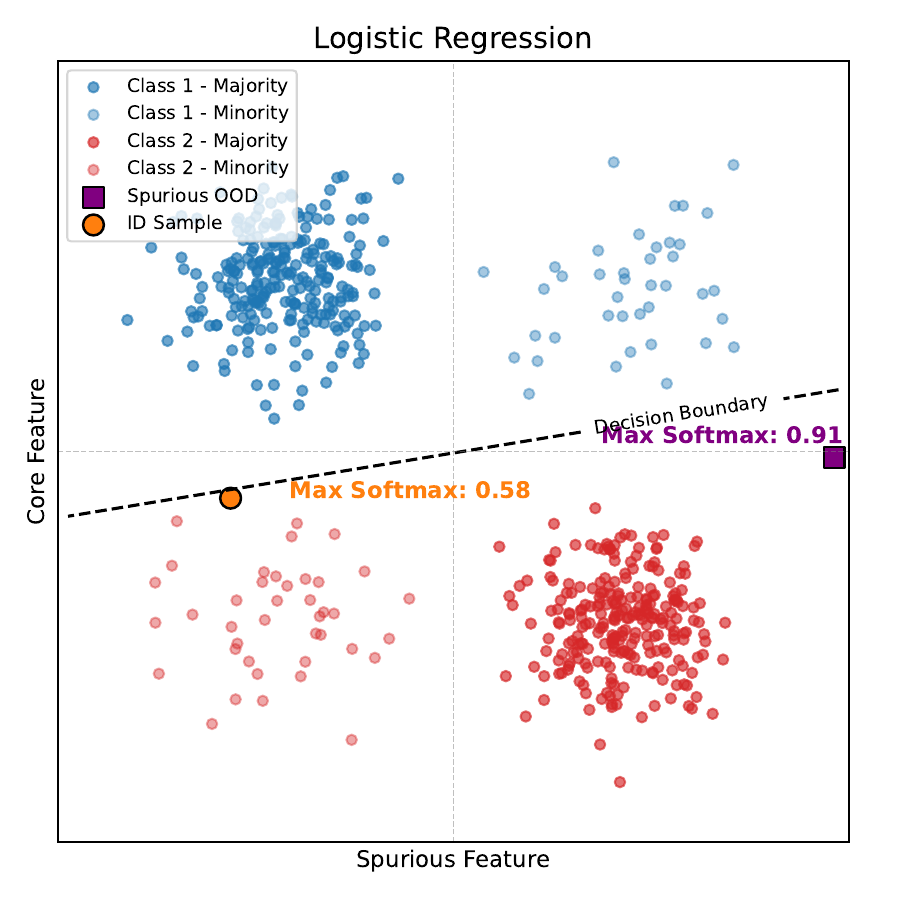}
        \vspace{-7mm}
        \caption{}\label{fig:toy:softmax_failure:sp}
    \end{subfigure}
    \vspace{-1mm}
    \caption{ 
\textbf{(a)} A far-OOD sample may receive a high softmax score, similar to a near-boundary ID sample.
\textbf{(b)} Distances to class prototypes offer a more consistent separation of OOD samples.
\textbf{(c)} In the SP-OOD setting, the problem is even more severe: A biased decision boundary causes the OOD sample to receive high softmax confidence, while a minority ID sample receives lower confidence.
}
\vspace{-0.5mm}
\label{fig:toy:softmax_failure}
\end{figure}

A key element in OOD detection is the scoring function $\mathcal{S}(x)$, which assigns a scalar value reflecting how likely an input belongs to the in-distribution (ID). This score is typically derived from a model's learned representations or its predictive outputs. OOD detection is performed based on this score function. An effective OOD method offers distinct and well-separated distributions of scores for ID and OOD samples.

Prior studies show that feature-based scores, derived from intermediate representations~\cite{ODIN, energy, knn}, typically outperform those based on output probabilities~\cite{MSP}, especially in the presence of spurious correlations~\cite{main_ref}.
Our experiments further support this trend in certain settings, suggesting that output-based methods (especially those relying on softmax probabilities) may be less reliable when model confidence is influenced by spurious features.
To better understand this difference, consider two probabilistic perspectives in classification: discriminative models directly estimate $p(y|x)$ and focus on separating classes, while generative models estimate $p(x|y)$ and capture how data $x$ is distributed for each class $y$. Most distance-based OOD detection methods can be viewed as approximating a generative approach in feature space.

While directly modeling $p(y|x)$ is generally effective for classifying ID samples in standard settings, discriminative approaches have been reported to be more sensitive under distribution shifts, such as in continual learning~\cite{generative_disc} or in the presence of spurious correlations~\cite{li2024generative}.
Furthermore, their utility for OOD detection can be limited, particularly when methods are optimized solely on ID data without exposure to OOD examples.
The softmax function forces a normalized probability distribution over the known classes, which can degrade OOD detection performance.
For instance, an OOD sample that lies far from all class distributions may receive a high softmax probability if it is only slightly distant from a decision boundary. Conversely, an ID sample situated near the boundary between multiple classes could receive a low maximum softmax probability, potentially leading to its misclassification as OOD.
In contrast, distance-based approaches, which rely on the class-conditional distribution $p(x|y)$, are by nature more robust in these scenarios.  
OOD samples that share few characteristics with any known class typically exhibit low likelihood under all class-conditional distributions $p(x|y)$, and can be reliably identified, regardless of their proximity to decision boundaries.

Figure~\ref{fig:toy:softmax_failure} highlights the limitations of softmax-based OOD scoring in a controlled toy dataset.
The challenge becomes more pronounced in the presence of spurious correlations, as shown in Figure~\ref{fig:toy:softmax_failure:sp}. In this scenario, a discriminative model, potentially biased by spurious features, may assign high confidence to an SP-OOD sample that shares these spurious cues with an ID class, while simultaneously assigning low confidence to a minority ID sample that lacks them. 
Motivated by these observations, we design a generative distance-based approach that is more robust to unknown spurious correlations.

\section{SPROD}
\label{sec:Method}

\begin{figure*}[!t]
  \centering
  \includegraphics[width=0.88\linewidth]{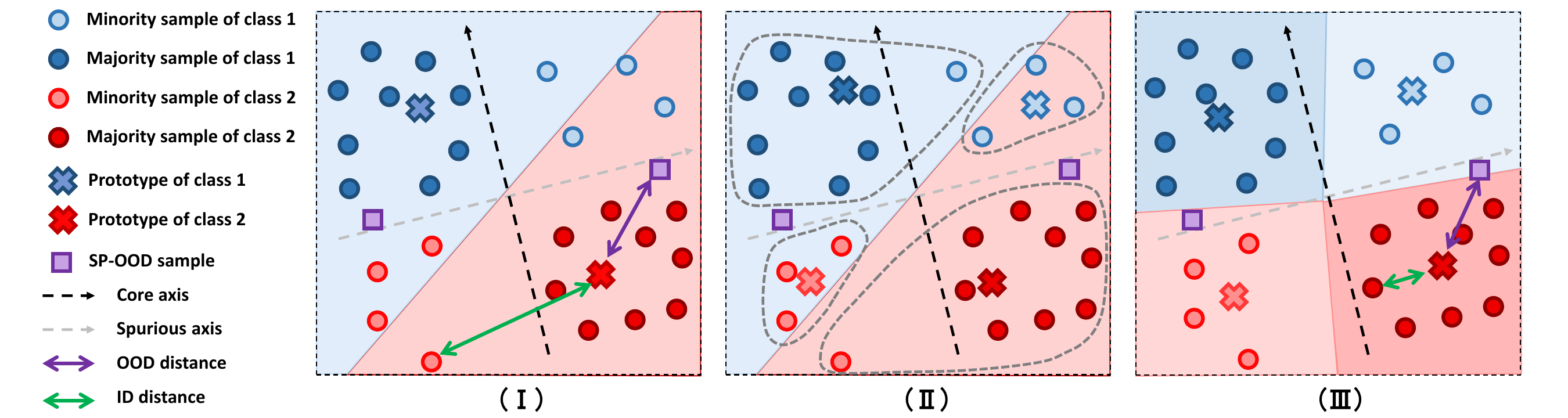}
  \vspace{-0.5mm}
   \caption{
    Overview of the three main stages of SPROD. In the first stage, class prototypes are computed, though they may be biased due to spurious correlations. In the second stage, group prototypes are constructed for the misclassified and correctly classified samples of each class. Finally, in the third stage, class samples are reassigned to their nearest group prototypes, and based on these assignments, refined minority and majority prototypes are recalculated.
    }
    \vspace{-0.5mm}
\label{fig:method:overview}
\end{figure*}

In this section, we introduce Spurious-Aware Prototype Refinement for Reliable Out-of-Distribution Detection (SPROD). SPROD adapts the prototypical framework~\cite{snell2017prototypical} for robust OOD detection by constructing class prototypes designed to be resilient to spurious correlations. The core method involves a three-stage process, which is shown in Figure~\ref{fig:method:overview} and detailed in the following subsections.

\subsection{Stage 1: Initial Prototype Construction}

Given a pretrained feature extractor $f_\theta$, we first obtain feature embeddings $h_i = f_\theta(x_i)$ for each training sample $x_i$.
To ensure uniformity in feature representation, these embeddings are normalized to have unit norm $z_i = h_i / \lVert h_i \rVert_2$.
For each ID class $c \in \mathcal{Y}$, an initial prototype $\mathbf{p}_c$ is computed as the mean of these normalized embeddings:
$
    \mathbf{p}_c = 1/N_c\sum_{i: y_i=c} z_i 
$, where $N_c$ is the number of samples in class $c$. Each $\mathbf{p}_c$ serves as an initial estimate of the class centroid in the normalized feature space.
A query sample $x_q$ (with normalized embedding $z_q$) is typically classified to the class $c$ whose prototype $\mathbf{p}_c$ is the closest (e.g., using Euclidean distance $d(z_q, \mathbf{p}_c)$). We use $d(z_q, \mathbf{p}_c)$ as the scoring function for OOD detection.
While our empirical results demonstrate the effectiveness of the na\"ive prototypical method in both SP-OOD and NSP-OOD settings, this approach remains vulnerable to biases from spurious correlations in the training data. As a result, the prototypes become skewed toward majority groups within each class (see part \uppercase\expandafter{\romannumeral 1\relax} of Figure~\ref{fig:method:overview}). This bias leads to a scenario where SP-OOD samples (represented by purple squares in Figure~\ref{fig:method:overview}) may be erroneously classified as ID due to their proximity to these biased prototypes. These limitations motivate the subsequent refinement stages.

\subsection{Stage 2: Classification-Aware Prototype Calculation}
To mitigate biases from spurious correlations present in the Stage 1 prototypes, we begin by analyzing how these initial prototypes classify the training data itself. 
Inspired by~\cite{wg3}
, our debiasing process starts with classifying the training samples based on the initial prototypes and partitioning the samples based on their prediction outcomes. For each class $c$, this identifies a set of correctly classified samples, $\mathcal{S}_c^{\text{corr}}$, and (multiple) sets of misclassified samples, $\{\mathcal{S}_{c \rightarrow m}^{\text{misc}}\}_{m\neq c}$, where $m$ is the incorrectly predicted class. The core assumption is that samples in $\mathcal{S}_{c \rightarrow m}^{\text{misc}}$ belong to subgroups of class $c$ that share spurious features with class $m$, leading to their misclassification.

Formally, for each class $c$, we compute the prototype $\mathbf{p}_c^{\text{corr}}$ by averaging over embeddings of correctly classified samples $\mathcal{S}_c^{\text{corr}}$. For the misclassified samples of class $c$, we compute the set of misclassified group prototypes $\{\mathbf{p}_{c \rightarrow m}^{\text{misc}}\}_{m\neq c}$ by averaging over samples in $\{\mathcal{S}_{c \rightarrow m}^{\text{misc}}\}_{m\neq c}$ individually. 
The number of misclassified group prototypes for class $c$, denoted by $C_c^{\text{misc}}$, corresponds to the number of other classes that training samples from class $c$ have been misclassified as during evaluation.

This procedure expands the number of prototypes per class from $1$ up to $C$ (total number of classes), helping incorporate diverse subgroup characteristics within each class.
However, this approach has a potential limitation.
Specifically, samples within the minority group may still contribute to the prototype for the correctly classified (majority) group if they were initially classified correctly, resulting in slightly biased prototypes (see part \uppercase\expandafter{\romannumeral 2\relax} of Figure~\ref{fig:method:overview}). Hence, we further refine the group prototypes in the third stage.

\subsection{Stage 3: Group Prototype Refinement}
    
In the third stage, we refine the group prototypes computed in Stage 2 to further reduce the remaining bias within them. Inspired by the reassignment step in K-means clustering, we first reassign samples within each class $\{z_i \mid y_i=c\}$ to either majority $\mathcal{S}_c^{\text{maj}}$ or minority groups $\{\mathcal{S}_{c \rightarrow m}^{\text{min}}\}_{m\neq c}$ based on their proximity to the corresponding prototypes ( $\mathbf{p}_c^{\text{corr}}$ for majority members and $\{\mathbf{p}_{c \rightarrow m}^{\text{misc}}\}_{m\neq c}$ for minority members). 
Following this reassignment, refined prototypes are computed as the mean of the updated group members:
\begin{equation}
\mathbf{p}_c^{\text{maj}} = \frac{1}{|\mathcal{S}_c^{\text{maj}}|} \sum_{z_i \in \mathcal{S}_c^{\text{maj}}} z_i, \quad
\mathbf{p}_{c \rightarrow m}^{\text{min}} = \frac{1}{|\mathcal{S}_{c \rightarrow m}^{\text{min}}|} \sum_{z_i \in \mathcal{S}_{c \rightarrow m}^{\text{min}}} z_i
\quad \forall m \neq c
\end{equation}

The refined prototypes $\mathbf{p}_c^{\text{maj}}$ and $\mathbf{p}_{c \rightarrow m}^{\text{min}}$, further reduce bias through the proposed refitting process.
During classification and OOD detection, the query embedding $z_q$ is compared to all group-specific prototypes, and the final prediction is based on the nearest prototype, regardless of group type.
This multiple-prototype approach reduces the likelihood of OOD samples being erroneously classified due to shared spurious attributes with any single prototype (see part \uppercase\expandafter{\romannumeral 3\relax} of Figure~\ref{fig:method:overview}).
For each sample, the OOD score is simply calculated based on the distance to the nearest group prototype.

\input{SPROD_algorithm}

The overall procedure of our prototype refinement strategy is outlined in Algorithm~\ref{alg:prototypical_ood} and a theoretical justification of how the proposed procedure mitigates spurious bias is provided in Appendix~\ref{sec:theory}.

\section{Experiments}
\label{sec:experimnts}

\subsection{Experimental Setup}
\label{sec:experimentalsetup}

We evaluate our method against a comprehensive suite of 19 post-hoc OOD detection approaches: MSP~\cite{MSP}, MDS~\cite{MD}, RMDS~\cite{RMD}, Energy~\cite{energy}, GradNorm~\cite{gradnorm}, ReAct~\cite{ReAct}, MaxLogit~\cite{MLS}, MLS \& KLM~\cite{MLS}, VIM~\cite{VIM}, KNN~\cite{knn}, SHE~\cite{SHE}, ASH~\cite{ASH}, NECO~\cite{Neco}, NNGuide~\cite{NNGuide}, Relation~\cite{Relation}, SCALE~\cite{Scale}, fDBD~\cite{FDBD}, and NCI~\cite{NCI}. We assess performance primarily using the Area Under the Receiver Operating Characteristic curve (AUROC), a threshold-independent metric, and the False Positive Rate at 95\% True Positive Rate (FPR@95). 
Additional metrics, including the Area Under the Precision-Recall curve (AUPR), are provided in Appendix~\ref{sec:backbone_ablation}. We repeat all experiments five times with different random seeds and report the mean and standard deviation.

In this experiments section, we focus on the more challenging SP-OOD scenarios, particularly those with the highest degree of spurious correlation in each dataset, as our method achieves near-perfect performance on far NSP-OOD samples. Detailed results for the NSP-OOD setting are provided in Appendix~\ref{sec:far_NSP_OOD_performance}. Further analyses of SPROD’s stages, along with broader evaluations across various transformer-based and convolutional backbones, are included in Appendix~\ref{sec:SPROD_ablation} and Appendix~\ref{sec:backbone_ablation}. For consistency, the results in this section primarily use the widely adopted ResNet-50~\cite{ResNet} backbone, with ResNet-18 additionally used in one analysis.
Beyond SP-OOD benchmarks, we also evaluate our approach under conventional (standard) OOD settings to further demonstrate its general applicability.

\subsection{Datasets}
\label{sec:datasets}

\begin{table*}[htb]
\caption{Overview of datasets used for SP-OOD evaluation. "\# Groups" denotes the number of distinct subpopulations based on class and spurious attribute combinations. "NA" indicates cases where such grouping is not explicitly defined.
}
\vspace{-2mm}
\centering
\fontsize{13}{15}\selectfont
\resizebox{0.94\textwidth}{!}{%
\begin{tabular}{llllcc}
\specialrule{1.5pt}{1pt}{1pt}
\textbf{Dataset} & \textbf{Type} & \textbf{\# Classes} & \textbf{\# Spurious Attr.} & \textbf{\# Groups} & \textbf{SP-OOD} \\
\specialrule{1.5pt}{1pt}{1pt}
Waterbirds (WB)~\cite{sagawa2020distributionallyrobustneuralnetworks}       & Synthetic     & ~~~~2 (Bird Type)         & 2 Backgrounds                   & 4                  & Places~\cite{Places} background               \\
CelebA (CA)~\cite{CelebA}           & Real-world    & ~~~~2 (Hair Color)        & 2 Genders                       & 4                  & Bald male (no hair)            \\
UrbanCars (UC)~\cite{UrbanCars}        & Synthetic    & ~~~~2 (Car Type)          & 2 Backgrounds $\times$ 2 Objects         & 8                  & Background / Background + Object \\
Animals MetaCoCo (AMC) [ours]  & Real-world    & ~~24 (Animal Type)      & 8 Backgrounds                   & NA             & Leave-2-out (class-based)      \\
Sp-ImageNet100 (SpI)~\cite{neuhaus2023spuriousfeatureslargescale}  & Real-world    & 100 (ImageNet classes)      & NA (spurious visual features)                   & NA             & Spurious ImageNet~\cite{neuhaus2023spuriousfeatureslargescale}       \\
\specialrule{1.5pt}{1pt}{1pt}
\end{tabular}}
\vspace{-2mm}
\label{table:datasets}
\vspace{-1mm}
\end{table*}
\vspace{1mm}

For evaluating SP-OOD detection, we utilize five diverse datasets, whose properties are summarized in Table~\ref{table:datasets}. Additional details and visual examples for all datasets, including the NSP datasets and their evaluation setup, are provided in Appendix~\ref{app:dataset_details}.
The datasets include \textbf{Waterbirds (WB)}~\cite{sagawa2020distributionallyrobustneuralnetworks} and \textbf{CelebA (CA)}~\cite{CelebA}. While widely adopted, these datasets present limitations in terms of scale and realism (CelebA, in particular, is noted for its label noise~\cite{10.1007/978-3-031-20713-6_10}), making them insufficient for comprehensive evaluation. To address this, our evaluation incorporates three additional datasets designed to test robustness under diverse conditions, including multi-class scenarios, multiple spurious attributes, and real-world complexity. 

\textbf{UrbanCars (UC)}~\cite{UrbanCars} is a binary classification dataset (urban vs. country cars) with two spurious attributes: background and a co-occurring object, both correlated with the class, making it a challenging multi-spurious benchmark. Next, we introduce \textbf{Animals MetaCoCo (AMC)}, a realistic SP-OOD benchmark created by subsampling and curating animal categories from MetaCoCo~\cite{MetaCoCo}. It contains 26 distinct classes, each with 8 different background types serving as imbalance shortcut attributes. We use a class-level leave-2-out setting, where two classes are held out as SP-OOD and the rest are treated as ID. The significant similarity in background distributions between the ID and OOD splits makes Animals MetaCoCo a particularly challenging multi-class benchmark for SP-OOD detection (Figure~\ref{fig:intro:data_example}). 
Last, we also use the \textbf{Spurious ImageNet} dataset introduced in~\cite{neuhaus2023spuriousfeatureslargescale} as SP-OOD data. This dataset consists of real-world images that contain only spurious features, such as bird feeders or graffiti, without the actual class objects, yet are consistently misclassified as specific ImageNet classes. These OOD samples are constructed for a subset of 100 ImageNet classes identified to rely on harmful spurious correlations. We refer to this subset of classes as Sp-ImageNet100 (SpI), a name we use for clarity, and treat it as the ID data.

For conventional OOD evaluations, we follow the same settings as in OpenOOD~\cite{openood}. Specifically, we use \textbf{CIFAR-10}~\cite{cifar}, \textbf{CIFAR-100}~\cite{cifar}, and \textbf{ImageNet-1k}~\cite{Imagenet}, along with their respective near-OOD and far-OOD datasets, to ensure consistency with widely adopted benchmarks in the field.
\subsection{Results}
\label{sec:results}

\input{backbone_tables/Main_ResNet-50}

\begin{figure*}[t]
  \centering

  \begin{subfigure}[b]{0.9\textwidth}
    \centering
    \includegraphics[width=0.49\textwidth]{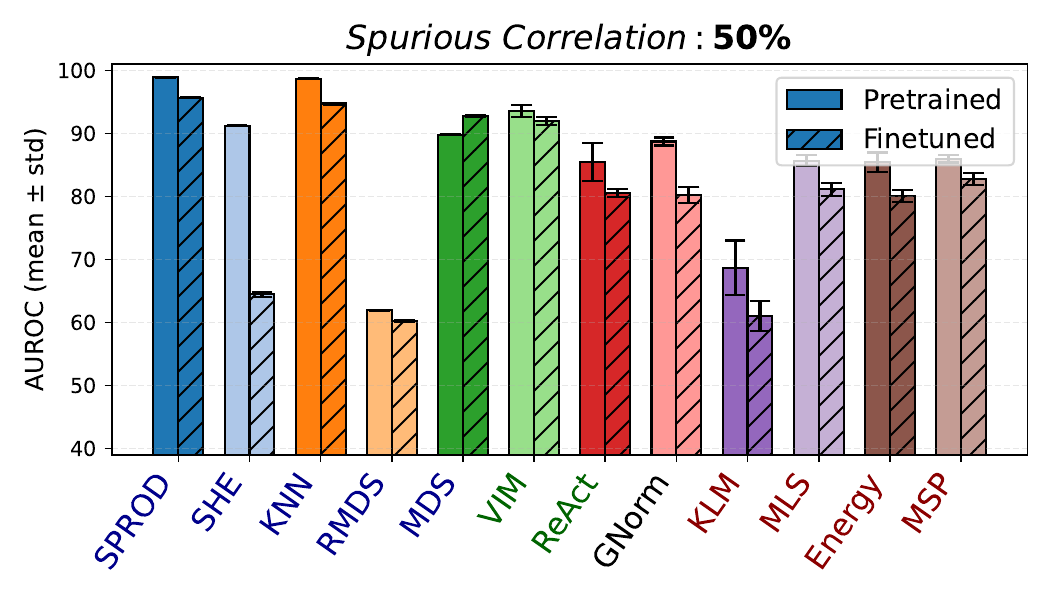}
    \includegraphics[width=0.49\textwidth]{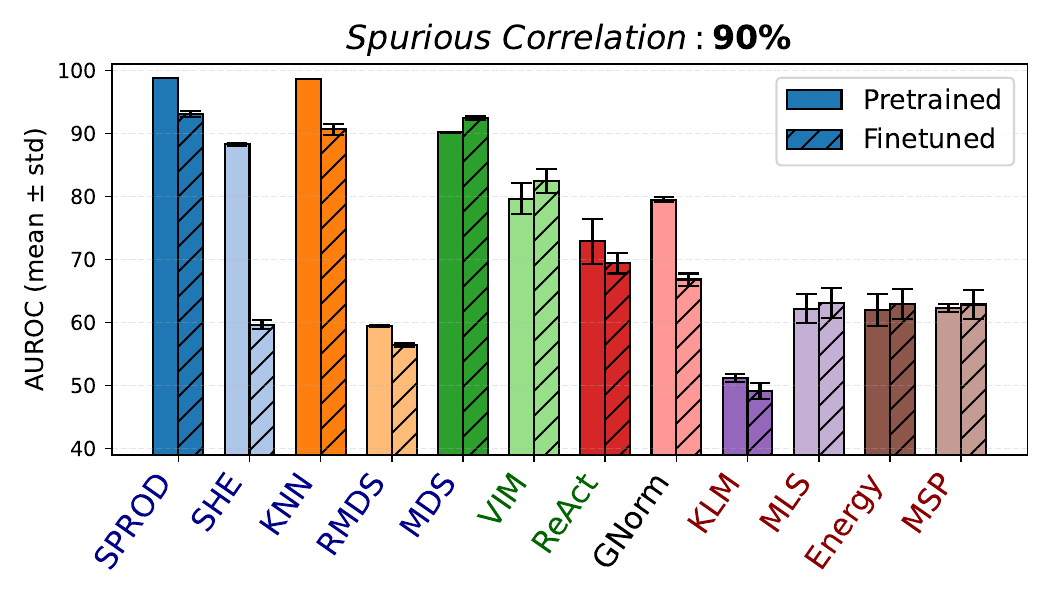}
    \vspace{-3.mm}
    \caption{ResNet-50 backbone.}
    \label{fig:wb_finetune_50}
  \end{subfigure}
  \hfill
  \begin{subfigure}[b]{0.9\textwidth}
    \centering
    \includegraphics[width=0.49\textwidth]{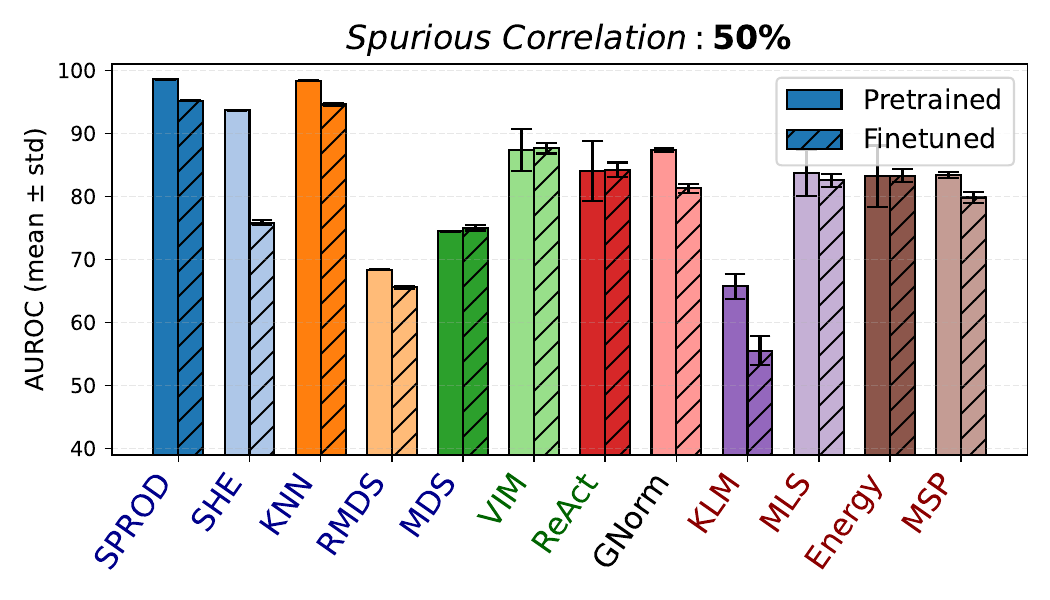}
    \includegraphics[width=0.49\textwidth]{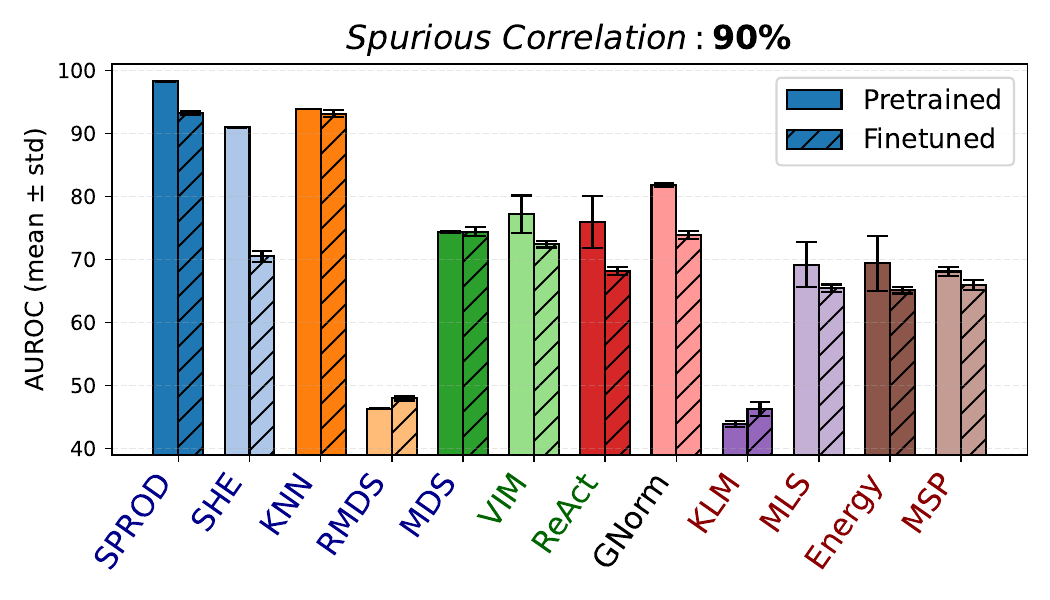}
    \vspace{-3.mm}
    \caption{ResNet-18 backbone.}
    \vspace{-1.mm}
    \label{fig:wb_finetune_18}
  \end{subfigure}

  \caption{Effect of backbone fine-tuning and spurious correlation on SP-OOD detection using the Waterbirds dataset. Left: ResNet-50; right: ResNet-18. Each pair shows results under 50\% (left) and 90\% (right) spurious correlation in ID data. Fine-tuned models are marked with a hatch texture.}
  \label{fig:wb_finetune}
\end{figure*}

The primary results, summarized in Table~\ref{table:main_result}, demonstrate the efficacy of SPROD across various datasets using a ResNet-50 backbone, evaluated with both AUROC and FPR@95 metrics. Additional results on other backbone architectures are provided in Appendix~\ref{sec:backbone_ablation}. SPROD consistently achieves superior performance compared to the 19 post-hoc baselines in all conducted experiments. In contrast, competing methods exhibit more variable performance, excelling in only a subset of the experimental settings. Generally, feature-based OOD detection methods tend to outperform output-based methods; however, this advantage appears to diminish on larger-scale datasets with multiple classes.
Overall, the average performance across all datasets reveals a notable margin by which SPROD surpasses the other evaluated baselines: Specifically, SPROD reaches an average of 85.1\% AUROC, 4.8\% higher (absolute) than the second best, KNN. For FPR@95, SPROD reaches 49.1\% error rate on average, 9.3\% better than the second best, KNN.
All other compared approaches are more than 20\% worse.

Next, we investigate the impact of backbone fine-tuning and correlation rate on OOD detection performance, using the Waterbirds dataset with ResNet-50 (Figure~\ref{fig:wb_finetune_50}) and ResNet-18 (Figure~\ref{fig:wb_finetune_18}) backbones. Fine-tuning the backbone on this dataset generally degrades OOD detection performance, a finding that contrasts with some common assumptions. This negative effect appears to be more pronounced for feature-based methods. As expected, increasing the spurious correlation rate of ID data from 50\% to 90\% leads to a general decline in performance across methods; however, this degradation is more noticeable for output-based techniques. Furthermore, the results suggest that employing a lighter, less expressive backbone (ResNet-18 compared to ResNet-50) does not lead to a substantial performance drop in OOD detection. Across these variations, SPROD demonstrates consistent robustness and maintains its performance as the spurious correlation rate increases.

\begin{figure}
    \centering
    \begin{subfigure}{0.57\textwidth}
        \centering
        \begin{subfigure}{0.49\textwidth}
            \includegraphics[width=\linewidth]{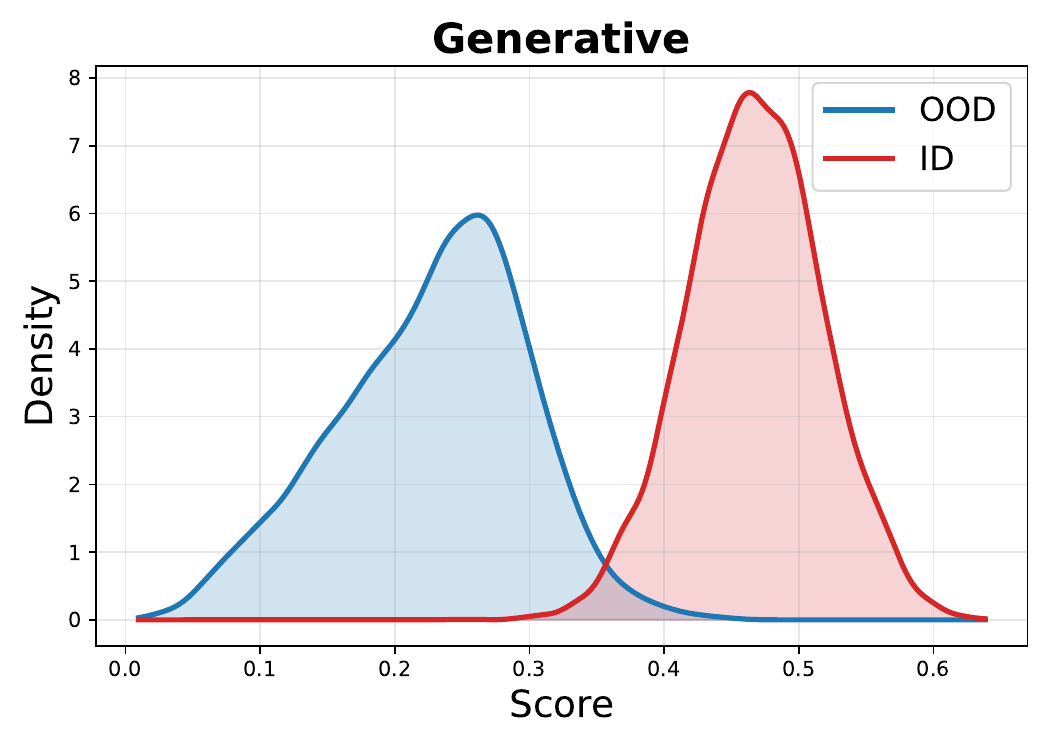}
        \end{subfigure}
        \begin{subfigure}{0.49\textwidth}
            \includegraphics[width=\linewidth]{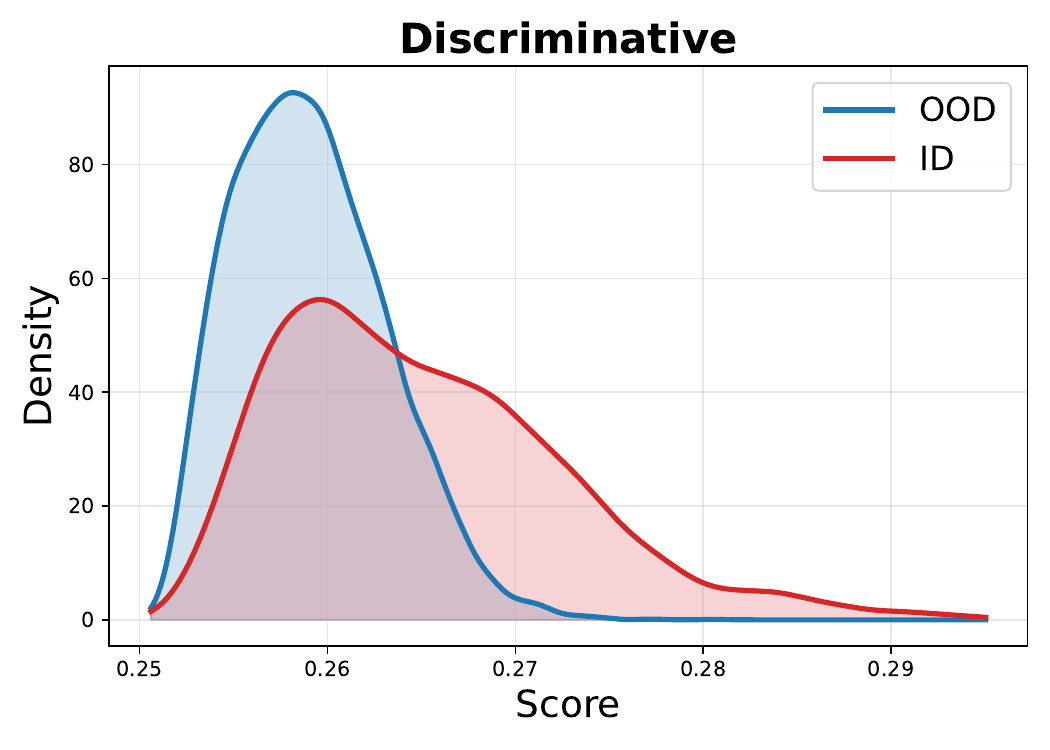}
        \end{subfigure}
        \vspace{-6.mm}
        \caption{}\label{fig:gen_disc:hist}
        \vspace{-2.mm}
    \end{subfigure}
    \begin{subfigure}{0.38\textwidth}
        \centering
        \includegraphics[width=\linewidth]{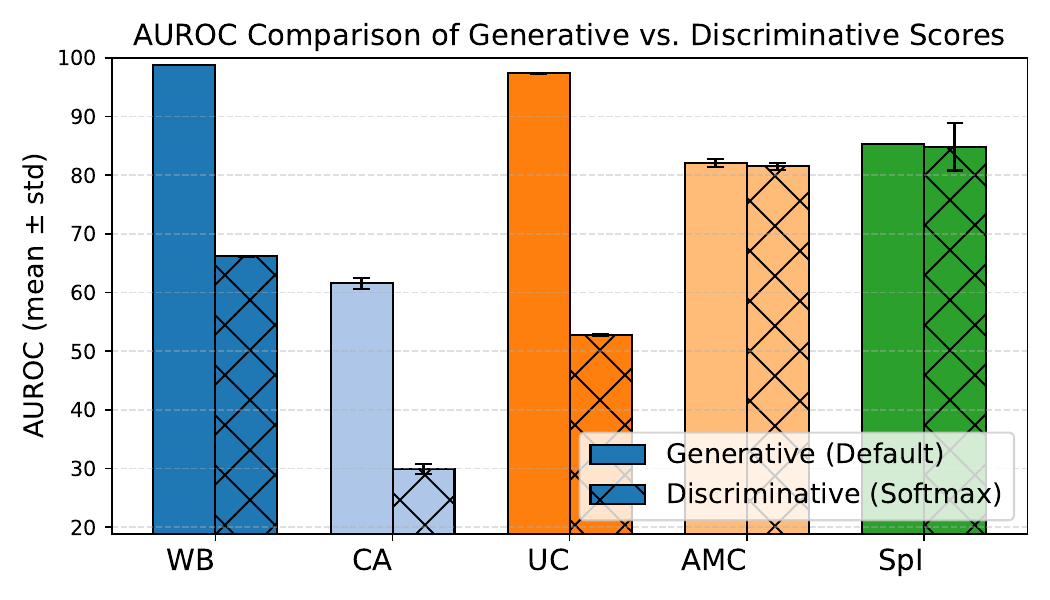}
        \vspace{-6.mm}        \caption{}\label{fig:gen_disc:benchmark}
        \vspace{-2.mm}
    \end{subfigure}
    
    \caption{
    Comparison of generative and discriminative scoring for OOD detection using SPROD.
    (a) Histograms of ID and OOD sample scores using the distance-based generative approach and the softmax-based discriminative approach, both computed with SPROD on the Waterbirds dataset. (b) Performance comparison between the generative (distance-based) and discriminative (softmax-based) scoring variants of SPROD across the five SP-OOD benchmark datasets.}
    \label{fig:gen_disc}
\end{figure}

To investigate the impact of the scoring mechanism, as discussed in Section~\ref{sec:motivation}, we conduct an ablation study using SPROD as the base method. This controlled experiment compares the effectiveness of deriving OOD scores in a discriminative manner $p(y|z)$ versus a generative manner $p(z|y)$. Both baselines utilize the same samples, feature embeddings, and refined prototypes from SPROD. For the discriminative score, we apply a softmax function to the negative distances between a sample's embedding $z$ and the class prototypes. The default SPROD approach, which uses the negative of the minimum distance to class prototypes, serves as the generative baseline (aligning with a log-likelihood under an exponential family distribution assumption).

Figure~\ref{fig:gen_disc:hist} presents histograms of the scores generated by both approaches, showing that the generative scoring method yields more distinctly separated distributions for ID and OOD samples. Furthermore, Figure~\ref{fig:gen_disc:benchmark} compares the OOD detection performance of these two scoring variants across our five SP-OOD benchmark settings. The results indicate that applying the softmax function for discriminative scoring substantially degrades performance, particularly in WaterBirds, CelebA and UrbanCars datasets. Conversely, on Animals MetaCoCo and Spurious ImageNet, the performance degradation from using softmax is less pronounced. This observation aligns with the trends in Table~\ref{table:main_result}, where traditional output-based methods tend to perform relatively better in these datasets.

\input{backbone_tables/General_ResNet-50}

To further assess the generality of SPROD beyond spurious correlation settings, we evaluate it on conventional OOD benchmarks using the standardized OpenOOD protocol~\cite{openood}. We compare SPROD with the top-performing post-hoc methods reported in the OpenOOD study. Table~\ref{tab:standard_ood_results} shows that SPROD attains performance comparable to or exceeding these methods and achieves the highest average score across all settings, confirming that SPROD performs reliably on standard OOD benchmarks and is not limited to spurious correlation scenarios.

\begin{figure}[htbp]
    \centering

    \begin{subfigure}[b]{0.32\textwidth}
        \includegraphics[width=\textwidth]{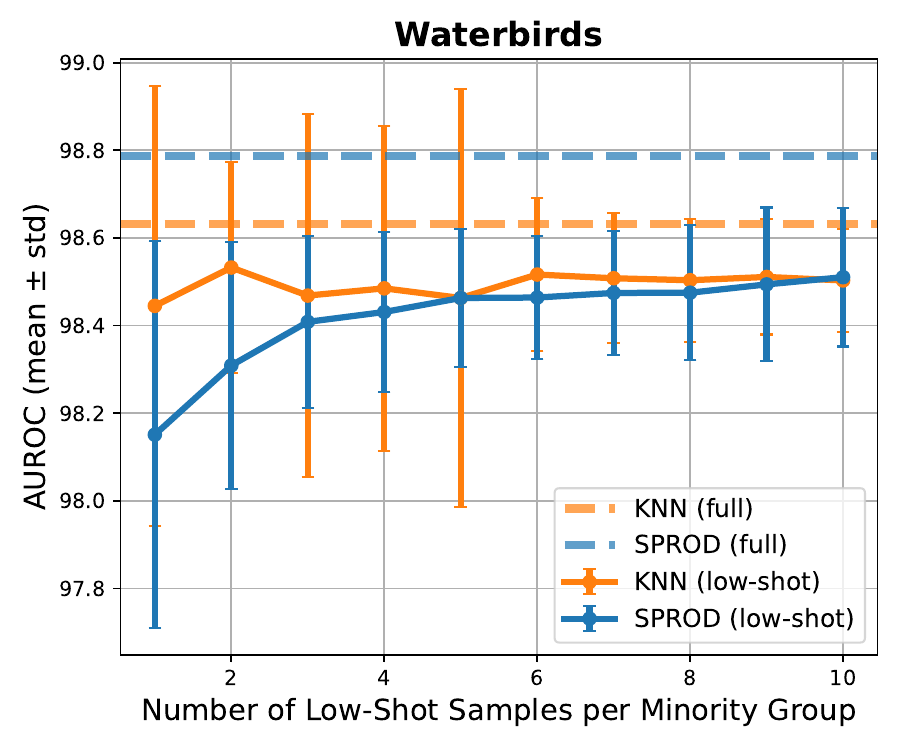}
        \label{fig:dataset1}
    \end{subfigure}
    \hfill
    \begin{subfigure}[b]{0.32\textwidth}
        \includegraphics[width=\textwidth]{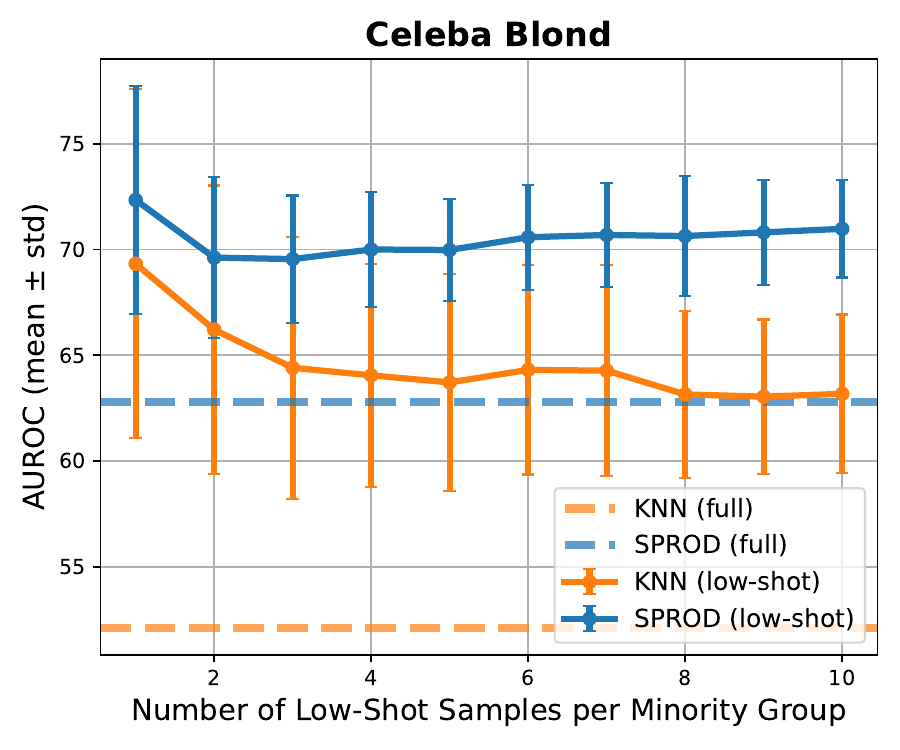}
        \label{fig:dataset2}
    \end{subfigure}
    \hfill
    \begin{subfigure}[b]{0.32\textwidth}
        \includegraphics[width=\textwidth]{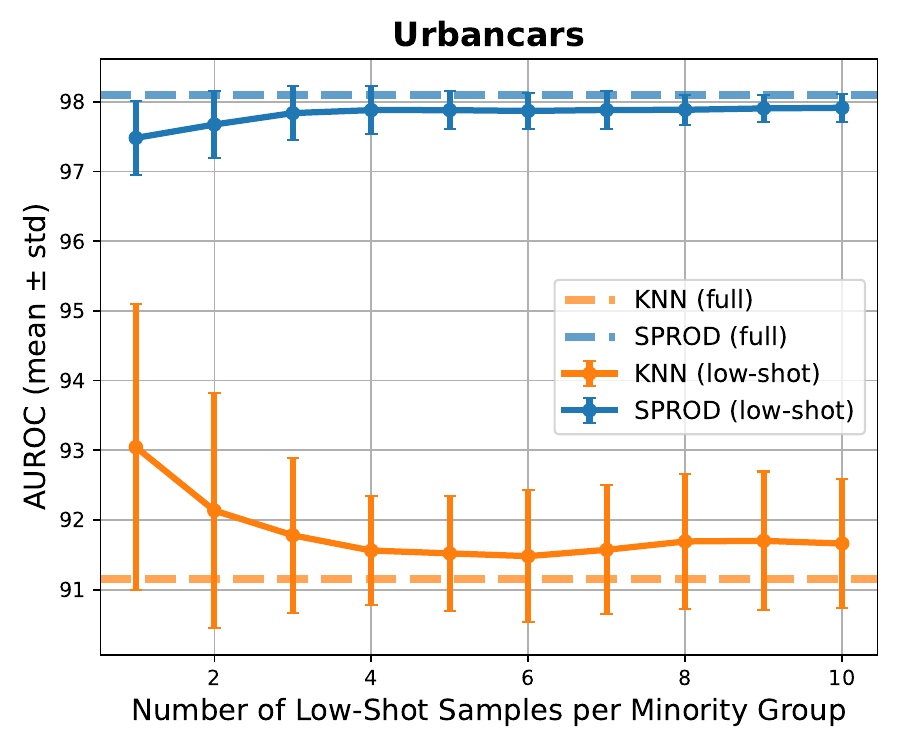}
        \label{fig:dataset3}
    \end{subfigure}
    \vspace{-5.mm}

    \caption{Performance of SPROD and KNN in low-shot SP-OOD settings across three datasets. Dashed lines indicate performance with the full training set, while solid lines show performance using varying numbers of samples per minority group.}
    \label{fig:lowshot_ood}
\end{figure}

While the simplicity of its prototypical framework makes SPROD a computationally efficient post-hoc method, its sample efficiency in SP-OOD settings also deserves investigation. To this end, we evaluate SPROD alongside the second top-performing method from Table~\ref{table:main_result} under low-shot conditions.  In this setup, we reduce the number of ID training samples while carefully preserving the original level of spurious correlation present in the experimental setting.
The results, presented in Figure~\ref{fig:lowshot_ood}, demonstrate that both SPROD and KNN maintain strong performance even in this data-scarce regime, highlighting the sample efficiency of these post-hoc approaches. Interestingly, for the CelebA dataset, both methods exhibit improved OOD detection performance in the low-shot setting compared to when trained on the full dataset.

Zero-shot OOD detection methods have reported results on Waterbirds using CLIP-B/16, but since they operate in a zero-shot setting, spurious correlations in the training set are not meaningful for them. Still, we compare: MCM~\cite{MCM} and CMA~\cite{CMA} achieve 98.36 and 98.62 AUROC using text inputs, while our text-free, vision-only method outperforms both with 99.01. COVER~\cite{cover} also evaluates this setting but over a different operating range, achieving lower AUROC scores (90.52 vs. 90.31 for MCM), highlighting our method's superior performance. Additional comparisons with existing SP-OOD methods (specifically CLIP-based approaches) are provided in Appendix~\ref{sec:SPOOD_methods}.

\section{Discussion}
\label{sec:discussion}

This paper introduces SPROD, a prototype-based method enhancing out-of-distribution (OOD) detection robustness against unknown spurious correlations. SPROD refines class prototypes by identifying and then adjusting for potential subgroups influenced by spurious features, thereby aligning representations with core, invariant class characteristics. A key strength is SPROD's efficiency and adaptability: as a post-hoc method, it integrates with various pre-trained feature extraction models without requiring retraining, additional OOD data, or hyperparameter tuning.

Experimental results consistently demonstrate the superiority of SPROD across ten convolutional and transformer-based backbones and five diverse SP-OOD benchmarks, including the newly introduced Animals MetaCoCo dataset. Our evaluations also reveal that fine-tuning the feature backbone on ID data can degrade SP-OOD detection performance. Furthermore, investigations into scoring mechanisms highlighted the advantages of distance-based approaches over softmax-based scoring for SP-OOD detection, particularly affirming the design choices in SPROD.

SPROD offers a scalable solution for improving robustness in OOD detection. While demonstrating significant advancements, SPROD's current formulation relies on class labels from the ID training data to construct class-conditional prototypes. This reliance aligns with the typical assumptions of the spurious correlation setting, which presumes access to data samples with spuriously correlated labels.
SPROD is not always the single best-performing method on every metric or in every setting (particularly in the conventional setting), but it remains competitive with the best-performing approaches. Such trade-offs are common in robustness research, where minor reductions in overall accuracy are often accepted to attain higher worst-group performance.
Future work could explore theoretical justifications for the robustness of the generative-like scoring mechanism or investigate more expressive approaches for modeling class-conditional distributions.

\newpage

\section*{Acknowledgments}

The research at TU Darmstadt was partially funded by an \textbf{Alexander von Humboldt Professorship in Multimodal Reliable AI}, sponsored by Germany’s Federal Ministry for Education and Research.



\bibliography{bibfile.bib}

\newpage
\appendix
\input{supplementary}

\clearpage
\FloatBarrier



\end{document}

%% file: SPROD_algorithm.tex

\begin{algorithm}[!htb]
\small
\caption{Spurious-Aware Prototype Refinement}
\label{alg:prototypical_ood}
\begin{algorithmic}[1]
\State \textbf{Input:} Training samples $\{(x_i, y_i)\}_{i=1}^{N}$, feature extractor $f_\theta$
\State \textbf{Output:} Refined class prototypes 
\State Get feature embedding $h_i = f_\theta(x_i)$ and $z_i = \frac{h_i}{\lVert h_i\rVert_2} \forall x_i$
\State \textbf{for} each class $c = 1, \dots, C$: \Comment{\textcolor{grayblue}{\textbf{Stage 1: Constructing class prototypes}}}
\State $\qquad \mathbf{p}_c = \frac{1}{N_c}\sum_{i: y_i=c} z_i$
    
\State Classify all training samples using initial prototypes
\State \textbf{for} each class $c = 1, \dots, C$: \Comment{\textcolor{grayblue}{\textbf{Stage 2: Augmenting class prototypes}}}
\State $\qquad$ Separate samples into correctly classified $\mathcal{S}_c^{\text{corr}}$ and misclassified $\{\mathcal{S}_{c \rightarrow m}^{\text{misc}}\}_{m\neq c}$
\State $\qquad$ Compute $\mathbf{p}_c^{\text{corr}}$ based on $\mathcal{S}_c^{\text{corr}}$ and $\{\mathbf{p}_{c \rightarrow m}^{\text{misc}}\}_{m\neq c}$ based on $\{\mathcal{S}_{c \rightarrow m}^{\text{misc}}\}_{m\neq c}$
\State \textbf{for} each class $c = 1, \dots, C$: \Comment{\textcolor{grayblue}{\textbf{Stage 3: Refining group prototypes}}}
    \State $\qquad$ Construct majority $\mathcal{S}_c^{\text{maj}}$ and minority $\{\mathcal{S}_{c \rightarrow m}^{\text{min}}\}_{m\neq c}$ groups based on proximity to $\mathbf{p}_c^{\text{corr}}$ or $\{\mathbf{p}_{c \rightarrow m}^{\text{misc}}\}_{m\neq c}$
    \State $\qquad$ Compute $\mathbf{p}_c^{\text{maj}}$ based on $\mathcal{S}_c^{\text{maj}}$
    and $\{\mathbf{p}_{c \rightarrow m}^{\text{min}}\}_{m\neq c}$ based on $\{\mathcal{S}_{c \rightarrow m}^{\text{min}}\}_{m\neq c}$
    
\end{algorithmic}
\end{algorithm}

%% file: backbone_tables/Main_ResNet-50.tex
    \begin{table}[t]
    \centering
    \caption{
    Comparative performance of post-hoc OOD detection methods on SP-OOD benchmarks using a ResNet-50 backbone. Left: AUROC scores (higher is better); Right: FPR@95 scores (lower is better). Feature-based methods are indicated in blue, output-based methods in red, gradient-based in black, and hybrid methods in green. For each experiment, the top-performing method is shown in \textbf{bold}, and the second-best is \underline{underlined}.
    }
    \label{table:main_result}
    \begin{minipage}{0.493\textwidth}
      \centering
      {\scriptsize\textbf{AUROC}$\uparrow$} \\
      \resizebox{\linewidth}{!}{
\begin{tabular}{lcccccc}
\toprule
Method & WB & CA & UC & AMC & SpI & Avg. \\
\midrule
\textcolor{darkred}{MSP}\cite{MSP} & $62.3_{\textcolor{gray}{\pm0.6}}$ & $46.0_{\textcolor{gray}{\pm1.4}}$ & $38.5_{\textcolor{gray}{\pm0.3}}$ & $79.7_{\textcolor{gray}{\pm0.4}}$ & $83.1_{\textcolor{gray}{\pm0.3}}$ & $61.9$ \\
\textcolor{darkred}{Energy}\cite{energy} & $62.0_{\textcolor{gray}{\pm2.6}}$ & $45.4_{\textcolor{gray}{\pm3.4}}$ & $38.4_{\textcolor{gray}{\pm2.1}}$ & $79.9_{\textcolor{gray}{\pm0.6}}$ & $80.6_{\textcolor{gray}{\pm0.4}}$ & $61.3$ \\
\textcolor{darkred}{MLS}\cite{MLS} & $62.2_{\textcolor{gray}{\pm2.3}}$ & $45.3_{\textcolor{gray}{\pm3.2}}$ & $38.4_{\textcolor{gray}{\pm1.4}}$ & $80.2_{\textcolor{gray}{\pm0.6}}$ & $81.9_{\textcolor{gray}{\pm0.3}}$ & $61.6$ \\
\textcolor{darkred}{KLM}\cite{MLS} & $51.2_{\textcolor{gray}{\pm0.7}}$ & $41.7_{\textcolor{gray}{\pm2.5}}$ & $57.0_{\textcolor{gray}{\pm0.2}}$ & $74.2_{\textcolor{gray}{\pm0.6}}$ & $79.6_{\textcolor{gray}{\pm0.8}}$ & $60.7$ \\
\textcolor{darkred}{GEN}\cite{Gen} & $62.3_{\textcolor{gray}{\pm0.6}}$ & $46.0_{\textcolor{gray}{\pm1.4}}$ & $38.5_{\textcolor{gray}{\pm0.3}}$ & $80.2_{\textcolor{gray}{\pm0.0}}$ & $80.8_{\textcolor{gray}{\pm0.4}}$ & $61.6$ \\
GNorm\cite{gradnorm} & $79.5_{\textcolor{gray}{\pm0.4}}$ & $38.0_{\textcolor{gray}{\pm1.3}}$ & $46.6_{\textcolor{gray}{\pm0.4}}$ & $74.2_{\textcolor{gray}{\pm0.5}}$ & $85.2_{\textcolor{gray}{\pm0.2}}$ & $64.7$ \\
\textcolor{darkgreen}{ReAct}\cite{ReAct} & $72.9_{\textcolor{gray}{\pm3.6}}$ & $45.6_{\textcolor{gray}{\pm5.3}}$ & $41.3_{\textcolor{gray}{\pm3.1}}$ & $80.1_{\textcolor{gray}{\pm0.6}}$ & $83.6_{\textcolor{gray}{\pm0.7}}$ & $64.7$ \\
\textcolor{darkgreen}{VIM}\cite{VIM} & $79.6_{\textcolor{gray}{\pm2.5}}$ & $50.4_{\textcolor{gray}{\pm3.1}}$ & $60.7_{\textcolor{gray}{\pm1.7}}$ & $78.6_{\textcolor{gray}{\pm0.6}}$ & $77.4_{\textcolor{gray}{\pm0.9}}$ & $69.3$ \\
\textcolor{darkgreen}{ASH}\cite{ASH} & $78.5_{\textcolor{gray}{\pm3.2}}$ & $47.3_{\textcolor{gray}{\pm2.8}}$ & $39.6_{\textcolor{gray}{\pm1.7}}$ & $78.0_{\textcolor{gray}{\pm0.2}}$ & $\textbf{86.6}_{\textcolor{gray}{\pm0.7}}$ & $66.0$ \\
\textcolor{darkblue}{MDS}\cite{MD} & $90.2_{\textcolor{gray}{\pm0.1}}$ & $57.8_{\textcolor{gray}{\pm0.5}}$ & $91.8_{\textcolor{gray}{\pm0.1}}$ & $62.9_{\textcolor{gray}{\pm0.8}}$ & $58.4_{\textcolor{gray}{\pm0.1}}$ & $72.2$ \\
\textcolor{darkblue}{RMDS}\cite{RMD} & $59.4_{\textcolor{gray}{\pm0.1}}$ & $33.6_{\textcolor{gray}{\pm1.4}}$ & $47.4_{\textcolor{gray}{\pm0.2}}$ & $\underline{81.9}_{\textcolor{gray}{\pm0.4}}$ & $68.8_{\textcolor{gray}{\pm0.1}}$ & $58.2$ \\
\textcolor{darkblue}{KNN}\cite{knn} & $\underline{98.6}_{\textcolor{gray}{\pm0.0}}$ & $54.5_{\textcolor{gray}{\pm0.5}}$ & $91.1_{\textcolor{gray}{\pm0.1}}$ & $79.7_{\textcolor{gray}{\pm0.6}}$ & $77.4_{\textcolor{gray}{\pm0.0}}$ & $\underline{80.3}$ \\
\textcolor{darkblue}{SHE}\cite{SHE} & $88.3_{\textcolor{gray}{\pm0.2}}$ & $42.7_{\textcolor{gray}{\pm0.6}}$ & $73.2_{\textcolor{gray}{\pm0.1}}$ & $54.8_{\textcolor{gray}{\pm0.7}}$ & $83.0_{\textcolor{gray}{\pm0.1}}$ & $68.4$ \\
\textcolor{darkblue}{NECO}\cite{Neco} & $53.5_{\textcolor{gray}{\pm1.6}}$ & $39.5_{\textcolor{gray}{\pm3.2}}$ & $35.1_{\textcolor{gray}{\pm1.5}}$ & $80.2_{\textcolor{gray}{\pm0.1}}$ & $67.2_{\textcolor{gray}{\pm0.3}}$ & $55.1$ \\
\textcolor{darkblue}{NNGuide}\cite{NNGuide} & $70.6_{\textcolor{gray}{\pm2.9}}$ & $49.8_{\textcolor{gray}{\pm4.2}}$ & $43.6_{\textcolor{gray}{\pm2.1}}$ & $79.4_{\textcolor{gray}{\pm0.0}}$ & $85.1_{\textcolor{gray}{\pm0.8}}$ & $65.7$ \\
\textcolor{darkblue}{Relation}\cite{Relation} & $80.7_{\textcolor{gray}{\pm0.2}}$ & $\underline{60.4}_{\textcolor{gray}{\pm2.5}}$ & $\underline{96.0}_{\textcolor{gray}{\pm0.5}}$ & $74.5_{\textcolor{gray}{\pm0.3}}$ & $81.8_{\textcolor{gray}{\pm0.7}}$ & $78.7$ \\
\textcolor{darkblue}{SCALE}\cite{Scale} & $89.0_{\textcolor{gray}{\pm2.9}}$ & $44.9_{\textcolor{gray}{\pm3.2}}$ & $54.4_{\textcolor{gray}{\pm2.1}}$ & $78.4_{\textcolor{gray}{\pm0.4}}$ & $\underline{86.2}_{\textcolor{gray}{\pm0.5}}$ & $70.6$ \\
\textcolor{darkblue}{fDBD}\cite{FDBD} & $71.1_{\textcolor{gray}{\pm0.5}}$ & $51.3_{\textcolor{gray}{\pm1.3}}$ & $47.4_{\textcolor{gray}{\pm0.2}}$ & $79.9_{\textcolor{gray}{\pm0.0}}$ & $84.2_{\textcolor{gray}{\pm0.3}}$ & $66.8$ \\
\textcolor{darkblue}{NCI}\cite{NCI} & $84.0_{\textcolor{gray}{\pm0.1}}$ & $46.4_{\textcolor{gray}{\pm2.4}}$ & $54.8_{\textcolor{gray}{\pm0.8}}$ & $78.5_{\textcolor{gray}{\pm0.1}}$ & $84.9_{\textcolor{gray}{\pm0.2}}$ & $69.7$ \\
\midrule
\textcolor{darkblue}{\textbf{SPROD}} & $\textbf{98.8}_{\textcolor{gray}{\pm0.0}}$ & $\textbf{61.6}_{\textcolor{gray}{\pm0.9}}$ & $\textbf{97.4}_{\textcolor{gray}{\pm0.0}}$ & $\textbf{82.4}_{\textcolor{gray}{\pm0.5}}$ & $85.3_{\textcolor{gray}{\pm0.0}}$ & $\textbf{85.1}$ \\
\bottomrule
\end{tabular}
}
    \end{minipage}
    \hfill
    \begin{minipage}{0.50\textwidth}
      \centering
      {\scriptsize\textbf{FPR@95}$\downarrow$} \\
      \resizebox{\linewidth}{!}{
\begin{tabular}{lcccccc}
\toprule
Method & WB & CA & UC & AMC & SpI & Avg. \\
\midrule
\textcolor{darkred}{MSP}\cite{MSP} & $87.9_{\textcolor{gray}{\pm0.8}}$ & $98.7_{\textcolor{gray}{\pm0.5}}$ & $97.3_{\textcolor{gray}{\pm0.3}}$ & $83.8_{\textcolor{gray}{\pm0.7}}$ & $74.1_{\textcolor{gray}{\pm1.2}}$ & $88.4$ \\
\textcolor{darkred}{Energy}\cite{energy} & $89.2_{\textcolor{gray}{\pm3.2}}$ & $98.6_{\textcolor{gray}{\pm0.7}}$ & $95.5_{\textcolor{gray}{\pm3.1}}$ & $84.8_{\textcolor{gray}{\pm0.8}}$ & $76.3_{\textcolor{gray}{\pm0.9}}$ & $88.9$ \\
\textcolor{darkred}{MLS}\cite{MLS} & $88.1_{\textcolor{gray}{\pm2.0}}$ & $98.8_{\textcolor{gray}{\pm0.6}}$ & $96.7_{\textcolor{gray}{\pm2.0}}$ & $84.4_{\textcolor{gray}{\pm0.8}}$ & $74.6_{\textcolor{gray}{\pm0.9}}$ & $88.5$ \\
\textcolor{darkred}{KLM}\cite{MLS} & $89.1_{\textcolor{gray}{\pm0.7}}$ & $98.7_{\textcolor{gray}{\pm0.5}}$ & $97.1_{\textcolor{gray}{\pm0.3}}$ & $80.5_{\textcolor{gray}{\pm0.8}}$ & $76.1_{\textcolor{gray}{\pm1.7}}$ & $88.3$ \\
\textcolor{darkred}{GEN}\cite{Gen} & $87.9_{\textcolor{gray}{\pm0.8}}$ & $98.7_{\textcolor{gray}{\pm0.5}}$ & $97.3_{\textcolor{gray}{\pm0.3}}$ & $84.8_{\textcolor{gray}{\pm0.1}}$ & $76.3_{\textcolor{gray}{\pm0.7}}$ & $89.0$ \\
GNorm\cite{gradnorm} & $84.2_{\textcolor{gray}{\pm0.7}}$ & $98.8_{\textcolor{gray}{\pm0.4}}$ & $97.1_{\textcolor{gray}{\pm0.1}}$ & $84.2_{\textcolor{gray}{\pm0.6}}$ & $54.7_{\textcolor{gray}{\pm0.6}}$ & $83.8$ \\
\textcolor{darkgreen}{ReAct}\cite{ReAct} & $86.9_{\textcolor{gray}{\pm7.0}}$ & $96.3_{\textcolor{gray}{\pm2.4}}$ & $95.5_{\textcolor{gray}{\pm3.2}}$ & $83.9_{\textcolor{gray}{\pm0.8}}$ & $57.5_{\textcolor{gray}{\pm1.6}}$ & $84.0$ \\
\textcolor{darkgreen}{VIM}\cite{VIM} & $61.4_{\textcolor{gray}{\pm3.5}}$ & $96.2_{\textcolor{gray}{\pm0.4}}$ & $69.0_{\textcolor{gray}{\pm1.5}}$ & $86.6_{\textcolor{gray}{\pm0.7}}$ & $79.5_{\textcolor{gray}{\pm0.5}}$ & $78.5$ \\
\textcolor{darkgreen}{ASH}\cite{ASH} & $85.2_{\textcolor{gray}{\pm7.0}}$ & $96.9_{\textcolor{gray}{\pm1.4}}$ & $96.1_{\textcolor{gray}{\pm1.5}}$ & $87.9_{\textcolor{gray}{\pm0.4}}$ & $52.9_{\textcolor{gray}{\pm3.1}}$ & $83.8$ \\
\textcolor{darkblue}{MDS}\cite{MD} & $49.2_{\textcolor{gray}{\pm0.2}}$ & $96.0_{\textcolor{gray}{\pm0.5}}$ & $39.0_{\textcolor{gray}{\pm0.3}}$ & $93.0_{\textcolor{gray}{\pm0.3}}$ & $90.5_{\textcolor{gray}{\pm0.1}}$ & $73.5$ \\
\textcolor{darkblue}{RMDS}\cite{RMD} & $91.7_{\textcolor{gray}{\pm0.2}}$ & $99.6_{\textcolor{gray}{\pm0.1}}$ & $95.3_{\textcolor{gray}{\pm0.1}}$ & $83.4_{\textcolor{gray}{\pm0.9}}$ & $88.1_{\textcolor{gray}{\pm0.1}}$ & $91.6$ \\
\textcolor{darkblue}{KNN}\cite{knn} & $\underline{4.8}_{\textcolor{gray}{\pm0.1}}$ & $\underline{94.4}_{\textcolor{gray}{\pm1.0}}$ & $42.5_{\textcolor{gray}{\pm0.2}}$ & $79.9_{\textcolor{gray}{\pm1.1}}$ & $70.4_{\textcolor{gray}{\pm0.2}}$ & $\underline{58.4}$ \\
\textcolor{darkblue}{SHE}\cite{SHE} & $33.2_{\textcolor{gray}{\pm0.5}}$ & $96.4_{\textcolor{gray}{\pm0.5}}$ & $76.5_{\textcolor{gray}{\pm0.2}}$ & $93.9_{\textcolor{gray}{\pm0.3}}$ & $\underline{52.6}_{\textcolor{gray}{\pm0.8}}$ & $70.5$ \\
\textcolor{darkblue}{NECO}\cite{Neco} & $90.5_{\textcolor{gray}{\pm2.2}}$ & $98.8_{\textcolor{gray}{\pm0.6}}$ & $96.7_{\textcolor{gray}{\pm1.9}}$ & $\underline{78.2}_{\textcolor{gray}{\pm0.0}}$ & $89.9_{\textcolor{gray}{\pm0.8}}$ & $90.8$ \\
\textcolor{darkblue}{NNGuide}\cite{NNGuide} & $77.7_{\textcolor{gray}{\pm6.0}}$ & $97.6_{\textcolor{gray}{\pm1.2}}$ & $91.6_{\textcolor{gray}{\pm3.6}}$ & $86.3_{\textcolor{gray}{\pm0.1}}$ & $\textbf{52.2}_{\textcolor{gray}{\pm1.4}}$ & $81.1$ \\
\textcolor{darkblue}{Relation}\cite{Relation} & $73.8_{\textcolor{gray}{\pm0.5}}$ & $95.4_{\textcolor{gray}{\pm0.2}}$ & $\underline{24.2}_{\textcolor{gray}{\pm1.7}}$ & $84.6_{\textcolor{gray}{\pm0.2}}$ & $78.0_{\textcolor{gray}{\pm1.1}}$ & $71.2$ \\
\textcolor{darkblue}{SCALE}\cite{Scale} & $61.0_{\textcolor{gray}{\pm22.5}}$ & $98.7_{\textcolor{gray}{\pm0.6}}$ & $94.6_{\textcolor{gray}{\pm3.1}}$ & $87.7_{\textcolor{gray}{\pm0.3}}$ & $53.0_{\textcolor{gray}{\pm1.5}}$ & $79.0$ \\
\textcolor{darkblue}{fDBD}\cite{FDBD} & $85.5_{\textcolor{gray}{\pm0.8}}$ & $98.6_{\textcolor{gray}{\pm0.5}}$ & $96.1_{\textcolor{gray}{\pm0.4}}$ & $85.4_{\textcolor{gray}{\pm0.2}}$ & $70.8_{\textcolor{gray}{\pm1.0}}$ & $87.3$ \\
\textcolor{darkblue}{NCI}\cite{NCI} & $41.1_{\textcolor{gray}{\pm0.1}}$ & $99.4_{\textcolor{gray}{\pm0.3}}$ & $92.2_{\textcolor{gray}{\pm0.6}}$ & $85.8_{\textcolor{gray}{\pm0.3}}$ & $63.8_{\textcolor{gray}{\pm0.9}}$ & $76.5$ \\
\midrule
\textcolor{darkblue}{\textbf{SPROD}} & $\textbf{4.7}_{\textcolor{gray}{\pm0.1}}$ & $\textbf{93.7}_{\textcolor{gray}{\pm0.9}}$ & $\textbf{19.0}_{\textcolor{gray}{\pm0.4}}$ & $\textbf{69.5}_{\textcolor{gray}{\pm1.2}}$ & $58.0_{\textcolor{gray}{\pm0.1}}$ & $\textbf{49.0}$ \\
\bottomrule
\end{tabular}
}
    \end{minipage}
    \end{table}

%% file: backbone_tables/General_ResNet-50.tex
\begin{table}[!t]
\centering
\caption{Comparison with the best-performing methods reported in the OpenOOD~\cite{openood} on conventional (standard) OOD datasets (AUROC$\uparrow$). The highest scores are highlighted in \textbf{bold}.}
\label{tab:standard_ood_results}
\begin{minipage}{0.97\textwidth}
  \centering
  \resizebox{\linewidth}{!}{%
\begin{tabular}{lccccccc}
\toprule
\textbf{Method} & \textbf{CIFAR-10 Near} & \textbf{CIFAR-10 Far} & \textbf{CIFAR-100 Near} & \textbf{CIFAR-100 Far} & \textbf{ImageNet Near} & \textbf{ImageNet Far} & \textbf{Avg (\%)} \\
\midrule
RMDS~\cite{RMD} & 89.80 & 92.20 & 80.15 & \textbf{82.92} & 76.99 & 86.38 & 84.74 \\
MLS~\cite{MLS} & 87.52 & 91.10 & 81.05 & 79.67 & 76.46 & 89.57 & 84.23 \\
VIM~\cite{VIM} & 88.68 & \textbf{93.48} & 74.98 & 81.70 & 72.08 & 92.68 & 83.27 \\
KNN~\cite{knn} & \textbf{90.64} & 92.96 & 80.18 & 82.40 & 71.10 & 90.18 & 84.58 \\
ASH~\cite{ASH} & 75.12 & 78.49 & 78.20 & 80.58 & \textbf{78.17} & \textbf{95.74} & 81.05 \\
\specialrule{1pt}{1pt}{1pt}
\textbf{SPROD} & 89.04 & 91.78 & \textbf{81.80} & 79.93 & 75.06 & 95.29 & \textbf{85.15} \\
\bottomrule
\end{tabular}
  }
\end{minipage}
\end{table}

%% file: supplementary.tex
\clearpage

\definecolor{fandango}{rgb}{0.71, 0.2, 0.54}
\definecolor{mediumorchid}{rgb}{0.73, 0.33, 0.83}
\definecolor{palatinatepurple}{rgb}{0.41, 0.16, 0.38}
\definecolor{fluorescentorange}{rgb}{1.0, 0.75, 0.0}
\definecolor{lapislazuli}{rgb}{0.15, 0.38, 0.61}
\definecolor{blush}{rgb}{0.87, 0.36, 0.51}
\definecolor{cinereous}{rgb}{0.6, 0.51, 0.48}

\setlength{\arrayrulewidth}{0.3mm} 
\arrayrulecolor[HTML]{000000}

\definecolor{limegreen}{HTML}{32CD32} 
\definecolor{lightblue}{HTML}{4169E1} 
\definecolor{forestgreen}{HTML}{228B22} 
\definecolor{firebrick}{HTML}{B22222}

\section{Theoretical Analysis of SPROD}
\label{sec:theory}

SPROD employs a two-step prototype refinement strategy to approximate group-specific representations within each class. The first step creates new prototypes for misclassified training samples, motivated by the observation that minority group instances (those that deviate from dominant spurious patterns) are more prone to misclassification. However, some correctly classified minority samples may still be incorrectly assigned to biased prototypes.

To address this, the second step reassigns each training sample to its nearest prototype and recalculates prototypes based on these updated assignments. This reassignment helps produce cleaner subgroup representations by reducing the influence of spurious-majority samples on minority feature structure.

By applying these two steps, SPROD aims to obtain nearly pure group-specific prototypes. In the remainder of this section, we provide a theoretical formulation to explain why such subgroup-specific prototypes improve robustness to spurious OOD (SP-OOD) samples.

\subsubsection{Feature Decomposition}

Without loss of generality, we assume a binary classification problem with classes $c \in \{0,1\}$. Each input sample $x_i$ is mapped to a normalized feature embedding $z_i \in \mathbb{R}^d$. We consider a general decomposition of $z_i$ into three semantically distinct components: spurious, core, and irrelevant features. We model $z_i$ as a linear combination of three functionally distinct components that span the embedding space:

\[
z_i = \sum_{j=1}^{n_{u_c^{\text{sp}}}} \alpha_{i,j}^{\text{sp}} \, \vec{\mathbf{u}}_{c,j}^{\text{sp}} + 
       \sum_{j=1}^{n_{u_c^{\text{core}}}} \beta_{i,j}^{\text{core}} \, \vec{\mathbf{u}}_{c,j}^{\text{core}} + 
       \sum_{j=1}^{n_{u^{\text{irr}}}} \gamma_{i,j}^{\text{irr}} \, \vec{\mathbf{u}}_{j}^{\text{irr}},
\]

where the sets $\{\vec{\mathbf{u}}_{c,j}^{\text{sp}}\}$, $\{\vec{\mathbf{u}}_{c,j}^{\text{core}}\}$, and $\{\vec{\mathbf{u}}_{j}^{\text{irr}}\}$ form orthonormal bases for the spurious, core, and irrelevant subspaces, respectively. Each set of coefficients $\{\alpha_{i,j}^{\text{sp}}\}$, $\{\beta_{i,j}^{\text{core}}\}$, and $\{\gamma_{i,j}^{\text{irr}}\}$ is specific to instance $i$.

These subspaces span mutually exclusive semantic roles:
\begin{itemize}
    \item Spurious subspace: captures features correlated with the label during training but not causally predictive.
    \item Core subspace: captures features that are causally predictive of the label.
    \item Irrelevant subspace: captures features unrelated to the task.
\end{itemize}

By definition, the basis vectors across subspaces are orthogonal: if an irrelevant basis vector were correlated with a core basis, it would be predictive and thus belong in the core subspace.

This decomposition can be expressed compactly in matrix form:
\[
z_i = U_c^{\text{sp}} \, \vec{\boldsymbol{\alpha}}_i^{\text{sp}} + 
      U_c^{\text{core}} \, \vec{\boldsymbol{\beta}}_i^{\text{core}} + 
      U^{\text{irr}} \, \vec{\boldsymbol{\gamma}}_i^{\text{irr}},
\]
where each $U^{(\cdot)} \in \mathbb{R}^{d \times n_u^{(\cdot)}}$ is a matrix of orthonormal basis vectors for the corresponding subspace, and each coefficient vector (e.g., $\vec{\boldsymbol{\alpha}}_i^{\text{sp}} \in \mathbb{R}^{n_{u_c^{\text{sp}}}}$) represents the coordinates of $z_i$ in that subspace.

\subsubsection{Distributional Assumptions}

We assume the coefficient vectors for each instance $z_i$ are drawn from class-conditional distributions:

\[
\vec{\boldsymbol{\alpha}}_i^{\text{sp}} \sim P_c^{\text{sp}}, \quad 
\vec{\boldsymbol{\beta}}_i^{\text{core}} \sim P_c^{\text{core}},
\]

where $P_c^{\text{sp}}$ and $P_c^{\text{core}}$ are distributions over the spurious and core subspace coefficients for class $c$.

The irrelevant component is shared across classes:

\[
\vec{\boldsymbol{\gamma}}_i^{\text{irr}} \sim P^{\text{irr}}.
\]

\vspace{1em}
\subsubsection{Group Definitions}

Within each class $c$, we define majority and minority groups based on the alignment of spurious features.

\textbf{Majority group} samples $z_i \in \mathcal{S}_c^{\text{maj}}$ satisfy:

\[
\vec{\boldsymbol{\alpha}}_i^{\text{sp}} \sim P_c^{\text{sp}}, \quad 
\vec{\boldsymbol{\beta}}_i^{\text{core}} \sim P_c^{\text{core}}.
\]

\textbf{Minority group} samples $z_i \in \mathcal{S}_c^{\text{min}}$ satisfy:

\[
\vec{\boldsymbol{\alpha}}_i^{\text{sp}} \sim P_{1-c}^{\text{sp}}, \quad 
\vec{\boldsymbol{\beta}}_i^{\text{core}} \sim P_c^{\text{core}}.
\]

Let $C_c$ denote the number of samples in class $c$, and $|\mathcal{S}_c^{\text{maj}}|$ the number of majority samples. Then:

\[
r_c = \frac{|\mathcal{S}_c^{\text{maj}}|}{C_c} \in [0,1]
\]

quantifies the class-conditional spurious correlation strength.

\vspace{1em}
\subsubsection{OOD Sample Definition}

OOD samples may contain components beyond the core, spurious, and irrelevant subspaces. To model this, we extend the feature decomposition to include an additional subspace $U^{\text{ext}}$, capturing external factors not observed during training. We define an OOD sample as:

\[
z_{\text{OOD}} = 
U_c^{\text{sp}} \vec{\boldsymbol{\alpha}}_{\text{OOD}}^{\text{sp}} + 
U_c^{\text{irr}} \vec{\boldsymbol{\gamma}}_{\text{OOD}}^{\text{irr}} + 
U^{\text{ext}} \vec{\boldsymbol{\delta}}_{\text{OOD}}^{\text{ext}},
\]

where $\vec{\boldsymbol{\alpha}}_{\text{OOD}}^{\text{sp}} \sim P_c^{\text{sp}}$, 
$\vec{\boldsymbol{\gamma}}_{\text{OOD}}^{\text{irr}} \sim P^{\text{irr}}$, 
and $\vec{\boldsymbol{\delta}}_{\text{OOD}}^{\text{ext}}$ is unconstrained.
Hard OOD samples may also include core components drawn from a shifted distribution $\vec{\boldsymbol{\beta}}_{\text{OOD}}^{\text{core}} \sim Q^{\text{core}} \neq P^{\text{core}}_c$. Although it is possible to analyze the general form with core and external components, for simplicity we focus on near-OOD samples with no core or external components, i.e.,

\[
z_{\text{OOD}} = 
U_c^{\text{sp}} \vec{\boldsymbol{\alpha}}_{\text{OOD}}^{\text{sp}} + 
U_c^{\text{irr}} \vec{\boldsymbol{\gamma}}_{\text{OOD}}^{\text{irr}}, 
\quad \text{with } \vec{\boldsymbol{\beta}}_{\text{OOD}}^{\text{core}} = \vec{0},\;
\vec{\boldsymbol{\delta}}_{\text{OOD}}^{\text{ext}} = \vec{0}.
\]

We define spurious-OOD groups based on the source of the spurious component:

\[
\mathcal{S}_c^{\text{OOD}} = \left\{
z_{\text{OOD}} \,\middle|\,
\vec{\boldsymbol{\alpha}}_{\text{OOD}}^{\text{sp}} \sim P_c^{\text{sp}},\;
\vec{\boldsymbol{\gamma}}_{\text{OOD}}^{\text{irr}} \sim P^{\text{irr}}
\right\}, \quad c \in \{0,1\}.
\]

These near-OOD samples resemble class $c$ only through spurious features and match the in-distribution (ID) irrelevant component distribution.

\subsubsection{Prototype Calculation}

We define the expected coefficient vectors for each subspace:

\[
\vec{\boldsymbol{\mu}}_c^{\text{core}} = \mathbb{E}_{\vec{\boldsymbol{\beta}} \sim P_c^{\text{core}}}[\vec{\boldsymbol{\beta}}], \quad
\vec{\boldsymbol{\mu}}_c^{\text{sp}} = \mathbb{E}_{\vec{\boldsymbol{\alpha}} \sim P_c^{\text{sp}}}[\vec{\boldsymbol{\alpha}}], \quad
\vec{\boldsymbol{\mu}}^{\text{irr}} = \mathbb{E}_{\vec{\boldsymbol{\gamma}} \sim P^{\text{irr}}}[\vec{\boldsymbol{\gamma}}].
\]

Using these, we define the majority and minority subgroup prototypes for class $c \in \{0,1\}$ as:

\[
\vec{\mathbf{p}}_c^{\text{maj}} = 
U_c^{\text{sp}} \vec{\boldsymbol{\mu}}_c^{\text{sp}} +
U_c^{\text{core}} \vec{\boldsymbol{\mu}}_c^{\text{core}} +
U^{\text{irr}} \vec{\boldsymbol{\mu}}^{\text{irr}},
\]

\[
\vec{\mathbf{p}}_c^{\text{min}} = 
U_c^{\text{sp}} \vec{\boldsymbol{\mu}}_{1-c}^{\text{sp}} +
U_c^{\text{core}} \vec{\boldsymbol{\mu}}_c^{\text{core}} +
U^{\text{irr}} \vec{\boldsymbol{\mu}}^{\text{irr}}.
\]

The overall class prototype is a convex combination of subgroup prototypes:

\[
\vec{\mathbf{p}}_c = r_c \vec{\mathbf{p}}_c^{\text{maj}} + (1 - r_c) \vec{\mathbf{p}}_c^{\text{min}},
\]

where $r_c \in [0,1]$ denotes the proportion of majority group samples in class $c$.

\subsubsection{Bias in Prototype Distances Under Strong Spurious Correlation}

In standard prototype-based OOD detection, the class prototype $\vec{\mathbf{p}}_c$ is a convex combination of majority and minority subgroup prototypes. In spurious-dominated regimes where $r_c \approx 1$, we have:
\[
\vec{\mathbf{p}}_c \approx \vec{\mathbf{p}}_c^{\text{maj}} = 
U_c^{\text{sp}} \vec{\boldsymbol{\mu}}_c^{\text{sp}} +
U_c^{\text{core}} \vec{\boldsymbol{\mu}}_c^{\text{core}} +
U^{\text{irr}} \vec{\boldsymbol{\mu}}^{\text{irr}}.
\]

\vspace{0.8em}
\noindent
\textbf{Case 1: Spurious-OOD Sample.}
Let $z_{\text{OOD}} \in \mathcal{S}_c^{\text{OOD}}$ be an OOD sample with spurious alignment to class $c$, but no core features:
\[
z_{\text{OOD}} = 
U_c^{\text{sp}} \vec{\boldsymbol{\alpha}}_{\text{OOD}} +
U^{\text{irr}} \vec{\boldsymbol{\gamma}}_{\text{OOD}}, 
\quad \text{with } \vec{\boldsymbol{\alpha}}_{\text{OOD}} \sim P_c^{\text{sp}}, \;
\vec{\boldsymbol{\gamma}}_{\text{OOD}} \sim P^{\text{irr}}.
\]

The expected squared distance to the biased class prototype becomes:
\[
\mathbb{E}\left[\|z_{\text{OOD}} - \vec{\mathbf{p}}_c\|^2\right] =
\underbrace{\|\vec{\boldsymbol{\mu}}_c^{\text{core}}\|^2}_{\text{core bias}} +
\underbrace{\operatorname{Tr}(\Sigma_c^{\text{sp}})}_{\text{spurious variance}} +
\underbrace{\operatorname{Tr}(\Sigma^{\text{irr}})}_{\text{irrelevant variance}}.
\]

\vspace{0.8em}
\noindent
\textbf{Case 2: Minority In-Distribution Sample.}
Now consider an ID sample from the minority group:
\[
z_{\text{min}} = 
U_c^{\text{sp}} \vec{\boldsymbol{\alpha}}_{\text{min}} +
U_c^{\text{core}} \vec{\boldsymbol{\beta}}_{\text{min}} +
U^{\text{irr}} \vec{\boldsymbol{\gamma}}_{\text{min}},
\]
with:
\[
\vec{\boldsymbol{\alpha}}_{\text{min}} \sim P_{1-c}^{\text{sp}}, \quad
\vec{\boldsymbol{\beta}}_{\text{min}} \sim P_c^{\text{core}}, \quad
\vec{\boldsymbol{\gamma}}_{\text{min}} \sim P^{\text{irr}}.
\]

Its expected squared distance to the majority prototype is:
\[
\mathbb{E}\left[\|z_{\text{min}} - \vec{\mathbf{p}}_c^{\text{maj}}\|^2\right] =
\underbrace{\|\vec{\boldsymbol{\mu}}_{1-c}^{\text{sp}} - \vec{\boldsymbol{\mu}}_c^{\text{sp}}\|^2}_{\text{spurious bias}} +
\underbrace{\operatorname{Tr}(\Sigma_c^{\text{core}})}_{\text{core variance}} +
\underbrace{\operatorname{Tr}(\Sigma^{\text{irr}})}_{\text{irrelevant variance}}.
\]

\vspace{0.8em}
\noindent
\textbf{Key Insight:}  
Although $z_{\text{min}}$ is an ID sample, its distance to the biased prototype includes a \emph{spurious bias} term, while the OOD sample differs only in the \emph{core} direction. Depending on the relative magnitudes of the core and spurious bias terms, this highlights the potential for erroneous OOD detection when prototype estimates are biased toward spurious features.

For example, in the Waterbird dataset, the background (water or land) represents the spurious feature dimension characterized by distinct mean vectors $\vec{\boldsymbol{\mu}}_c^{\text{sp}}$ and $\vec{\boldsymbol{\mu}}_{1-c}^{\text{sp}}$. The difference $\|\vec{\boldsymbol{\mu}}_c^{\text{sp}} - \vec{\boldsymbol{\mu}}_{1-c}^{\text{sp}}\|$ may exceed the core bias in OOD distances, causing minority samples with conflicting backgrounds to lie farther from prototypes than some OOD samples.

\subsubsection{Why SPROD Mitigates Distance Bias}

If prototypes can be accurately estimated to match each group's distribution, then each ID sample $z_i \in \mathcal{S}_c^g$ can be compared to its corresponding prototype $\hat{\mathbf{p}}(z_i) = \vec{\mathbf{p}}_c^g$, where $g \in \{\text{maj}, \text{min}\}$.

Because the prototype shares the same spurious basis alignment, the spurious bias is eliminated. The expected squared distance becomes:
\[
\mathbb{E}\left[\|z_i - \hat{\mathbf{p}}(z_i)\|^2\right] = 
\operatorname{Tr}(\Sigma_c^{\text{core}}) + 
\operatorname{Tr}(\Sigma^{\text{irr}}).
\]

In contrast, the standard prototype-based approach introduces a systematic bias for minority samples:
\[
\mathbb{E}\left[\|z_{\text{min}} - \vec{\mathbf{p}}_c^{\text{maj}}\|^2\right] =
\textcolor{darkred}{\|\vec{\boldsymbol{\mu}}_{1-c}^{\text{sp}} - \vec{\boldsymbol{\mu}}_c^{\text{sp}}\|^2} +
\operatorname{Tr}(\Sigma_c^{\text{core}}) + \operatorname{Tr}(\Sigma^{\text{irr}}).
\]

\vspace{0.8em}
\noindent
For an SP-OOD sample, the prototype it is compared against (either group) contains a core component it lacks. Hence the expected distance becomes:
\[
\mathbb{E}\left[\|z_{\text{OOD}} - \hat{\mathbf{p}}\|^2\right] = 
\textcolor{darkblue}{\|\vec{\boldsymbol{\mu}}_c^{\text{core}}\|^2} +
\operatorname{Tr}(\Sigma_c^{\text{core}}) +
\operatorname{Tr}(\Sigma^{\text{irr}}).
\]

This is strictly greater than the refined ID distance, which includes neither the \textcolor{darkred}{spurious bias} nor the \textcolor{darkblue}{core bias} term:

\[
\mathbb{E}\left[\|z_{\text{OOD}} - \hat{\mathbf{p}}\|^2\right] 
> \mathbb{E}\left[\|z_i - \hat{\mathbf{p}}(z_i)\|^2\right].
\]

\vspace{0.8em}
\noindent
\textbf{Conclusion:}
SPROD removes the \textcolor{darkred}{spurious bias} term for ID samples and reduces overlap with OOD samples by accounting for their missing core features. This systematic (in expectation) bias elimination leads to tighter ID clusters and more reliable OOD detection.

\section{Spurious Correlation in OOD Detection: Literature Review}
\label{sec:spood_literature}
This work~\cite{main_ref} was the first to introduce the problem of spurious correlations in OOD detection, showing that detection performance degrades as spurious correlation increases, especially for SP-OOD samples. It also encouraged the community to report results of OOD methods under this challenging setting.
Recently, several works have addressed the challenge of \textbf{spurious correlations} in \textit{out-of-distribution (OOD)} detection, employing different strategies that can be broadly categorized into \textit{outlier exposure based}, \textit{training-time regularization}, and \textit{post-hoc methods}.

\textbf{Outlier exposure based methods} include \textbf{RONF}~\cite{Ronf}, which improves synthetic outlier generation and model fine-tuning using only ID data. It introduces \textit{Boundary Feature Mixup} to create more realistic virtual outliers by interpolating near decision boundaries, and \textit{Optimal Parameter Learning} to suppress spurious feature learning during training. At inference, it uses a custom \textit{Energy with Energy Discrepancy} score to better separate ID from OOD samples without relying on external OOD datasets. Similarly, \textbf{KIRBY}~\cite{kirby} generates hard negative samples by removing class-discriminative regions identified via Class Activation Maps (CAM) and inpainting these regions with background-like content, creating semantically degraded but visually plausible images as surrogate OOD examples. A lightweight rejection network trained on features from both clean and modified images enables strong OOD detection without requiring real OOD data or backbone retraining. Although not directly targeting spurious correlations, \textbf{ImOOD}~\cite{imood} evaluates robustness under spurious settings, focusing on long-tailed datasets where class imbalance biases OOD detection toward frequent (head) classes. It learns a bias correction term to shift OOD scores per input, improving separation especially for rare (tail) classes.

\textbf{Training-time regularization methods} seek to reduce reliance on spurious cues during model training. \textbf{BackMix}~\cite{backmix} regularizes models by mixing foreground objects with different backgrounds, breaking spurious correlations between objects and backgrounds. Using CAM to estimate foreground regions, it replaces background patches with those from other images while preserving labels, thus improving robustness primarily against background spuriousity. \textbf{RW}~\cite{rw} introduces a nuisance-aware training framework that reweights the training loss to reduce correlations between class labels and nuisance attributes. It further applies a penalty based on the \textit{Hilbert-Schmidt Independence Criterion (HSIC)} to explicitly remove nuisance information from learned features, enhancing semantic OOD detection under shared nuisance conditions. \textbf{NsED}~\cite{nsed} decomposes images into semantic (phase) and style (amplitude) components via Discrete Fourier Transform, generating augmented samples by mixing amplitude spectra across images. Training with a robust loss that minimizes worst-case classification error over these augmented samples results in features less sensitive to style variations.

\textbf{Post-hoc methods} operate after model training to improve OOD detection. \textbf{Projection Regret (PR)}~\cite{pr} uses partial diffusion and denoising to project inputs onto the ID manifold, measuring semantic novelty by comparing the original image with its projection. A second projection step recursively removes background bias, isolating true semantic differences and enabling object-level OOD detection focused on meaningful changes rather than background similarity. \textbf{CoVer}~\cite{cover} enhances detection by evaluating model confidence across multiple corrupted versions of the same input (e.g., fog, blur, contrast shifts). Instead of relying on a single prediction, it averages confidence scores over these variants, improving robustness.

Finally, several \textbf{zero-shot OOD detection} approaches, while not explicitly targeting spurious correlations, report results under limited spurious settings. \textbf{Maximum Concept Matching (MCM)}~\cite{MCM} leverages CLIP’s vision-language embeddings for zero-shot detection without fine-tuning or extra data. Each ID class is represented by a text prompt encoded as a concept prototype, and test images are scored based on cosine similarity with these prototypes. A softmax scaling sharpens distinctions between ID and OOD samples. Another zero-shot multimodal method~\cite{dai-etal-2023-exploring} improves performance by prompting a large language model to generate rich descriptors for each ID class, filtered by consistency on retrieval tasks to reduce hallucinations. It combines CLIP-based similarity scores between test images, filtered descriptors, and detected object labels to compute a robust matching score. \textbf{NegLabel}~\cite{neglabel} defines negative labels semantically distant from ID classes and computes softmax-based similarity scores that favor ID labels while penalizing negatives, optionally averaging over label groups to reduce noise. Lastly, \textbf{CMA}~\cite{CMA} embeds both ID class labels and neutral, class-agnostic Agents in a shared semantic space, leveraging a triangular similarity relationship to reduce confidence on OOD images while maintaining high confidence on ID data.

Together, these approaches represent diverse strategies to tackle the spurious correlation problem in OOD detection, from data augmentation and loss regularization to post-hoc corrections and zero-shot semantic reasoning.

While existing approaches have made significant strides in mitigating spurious correlations in OOD detection, each comes with certain trade-offs and constraints. Outlier exposure and training-time regularization methods often involve non-trivial training overhead, including access to auxiliary data or retraining backbone models with certain objectives, which limits their practicality in deployment settings. Moreover, some of these methods, such as BackMix~\cite{backmix}, primarily address specific types of spurious features (e.g., background), which may not generalize to more complex real-world scenarios involving multiple or less structured spurious cues. Post-hoc methods offer a more efficient alternative by avoiding retraining and operating directly on pre-trained models. However, many still rely on careful tuning of multiple hyperparameters or complex transformation pipelines (e.g., CoVer’s corruption sets~\cite{cover} or PR’s iterative projections~\cite{pr}), making them sensitive to design choices and dataset specifics. Furthermore, several zero-shot approaches utilize vision-language models like CLIP, benefiting from rich semantic priors but introducing dependencies on text prompts and external descriptors, which can be limiting when only visual features are available or when text-image alignment is imperfect.

\section{Comparison with SP-OOD Methods}
\label{sec:SPOOD_methods}

Most existing OOD detection methods are not directly comparable to SPROD, as they often rely on different assumptions, such as access to auxiliary OOD data, or necessitate retraining the feature backbone with specific objectives. These design choices contrast with SPROD’s post-hoc nature and its objectives of simplicity, efficiency, and broad applicability without model retraining. Nonetheless, to provide a broader context, we include results from a selection of methods that, while not perfectly aligned, have been evaluated under similar SP-OOD conditions. For fairness and consistency, we report the results as published in their original papers, rather than re-implementing them, thereby representing each method according to its best-reported performance.
As shown in Table~\ref{tab:ood_comparisons}, under the most comparable conditions presented, SPROD achieves highly competitive performance, notably without requiring model retraining (when using a ResNet-18 backbone, as applicable to some comparisons) or access to an auxiliary text modality (in the case of methods leveraging CLIP).

\input{appendix_tables/related_comparisons}

\begin{figure}[!t]
    \centering
    \begin{subfigure}[b]{0.42\textwidth}
        \centering
        \includegraphics[width=\textwidth]{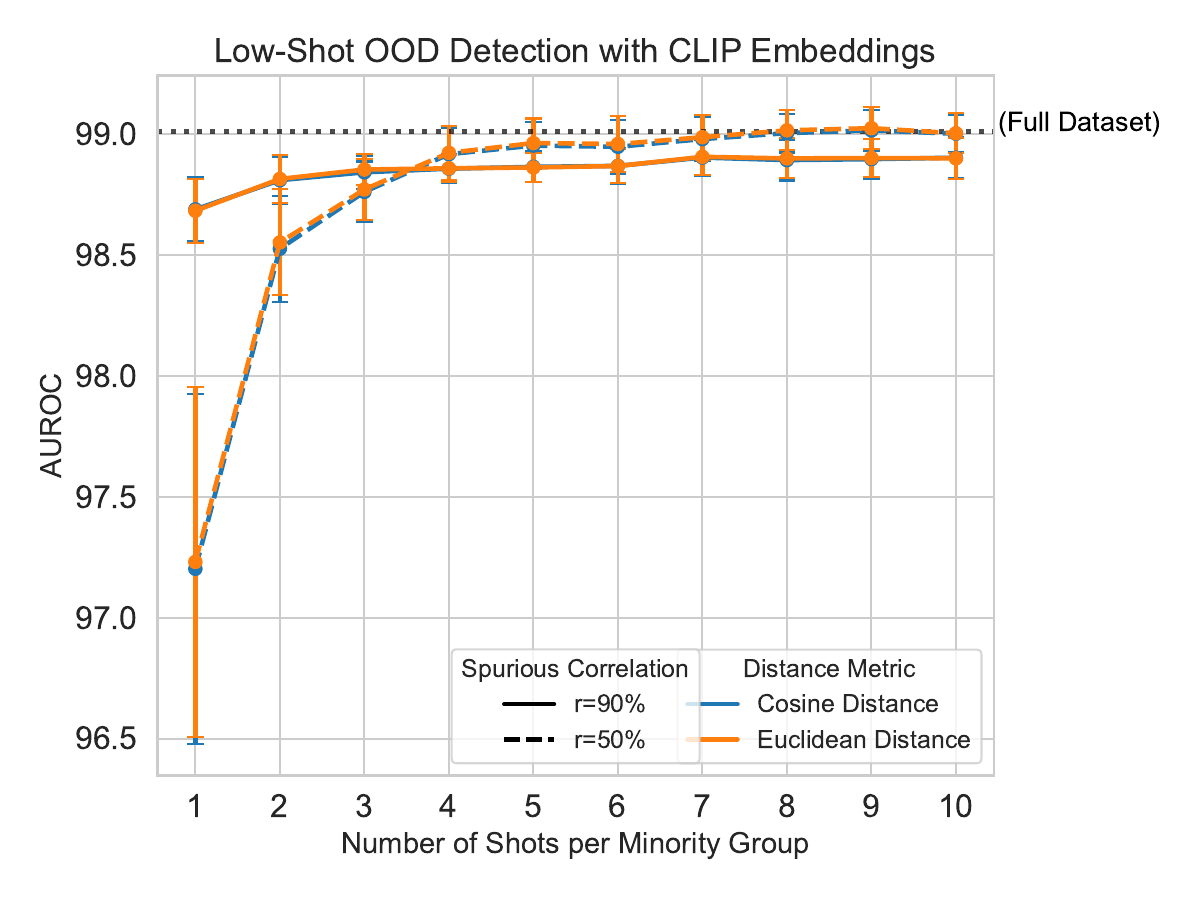}
        \vspace{-7mm}
        \caption{{\footnotesize AUROC}}
    \end{subfigure}
    \begin{subfigure}[b]{0.42\textwidth}
        \centering
        \includegraphics[width=\textwidth]{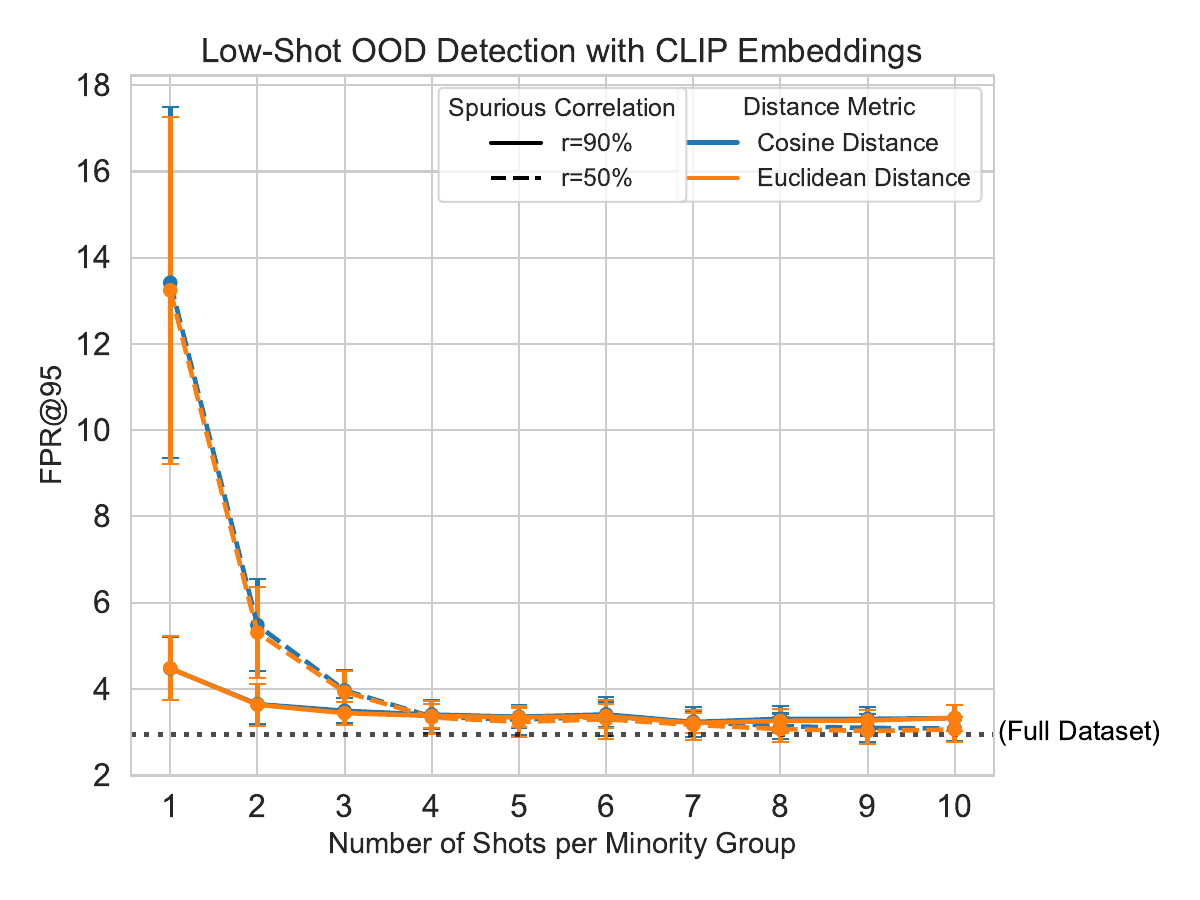}
        \vspace{-7mm}
        \caption{{\footnotesize FPR@95}}
    \end{subfigure}
    \vspace{-1mm}
    \caption{ 
Low-shot SP-OOD detection performance of SPROD on the Waterbirds dataset using features from a frozen CLIP ViT-B/16 vision encoder. Performance (AUROC and FPR@95) is shown as a function of the number of samples per minority group for two spurious correlation rates (50\% and 90\%) and two distance metrics (Euclidean and Cosine). Dashed lines indicate performance with the full training set.
}
\vspace{-0.5mm}
\label{fig:clip:lowshot}
\end{figure}

\begin{figure}[!t]
    \centering
    \begin{subfigure}[b]{0.42\textwidth}
        \centering
        \includegraphics[width=\textwidth]{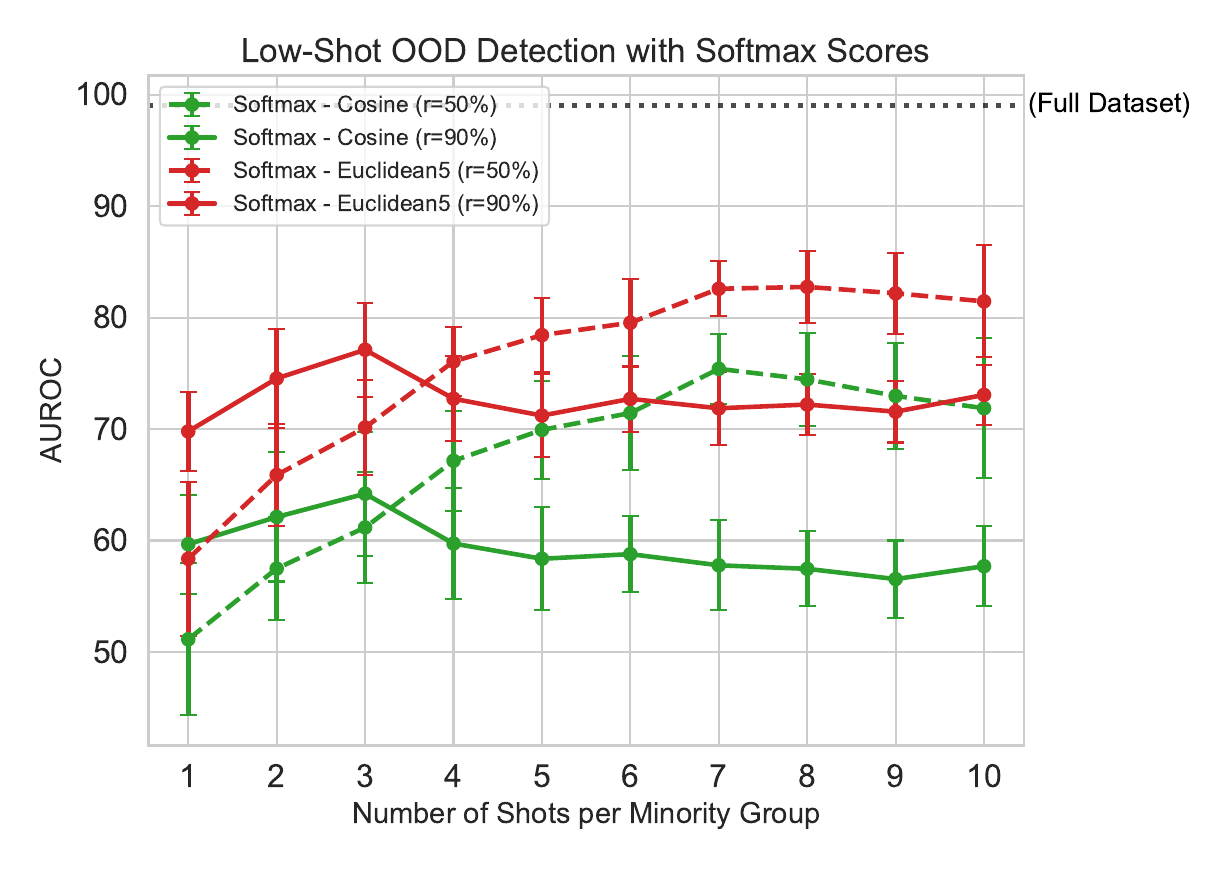}
        \vspace{-7mm}
        \caption{{\footnotesize AUROC}}
    \end{subfigure}
    \begin{subfigure}[b]{0.42\textwidth}
        \centering
        \includegraphics[width=\textwidth]{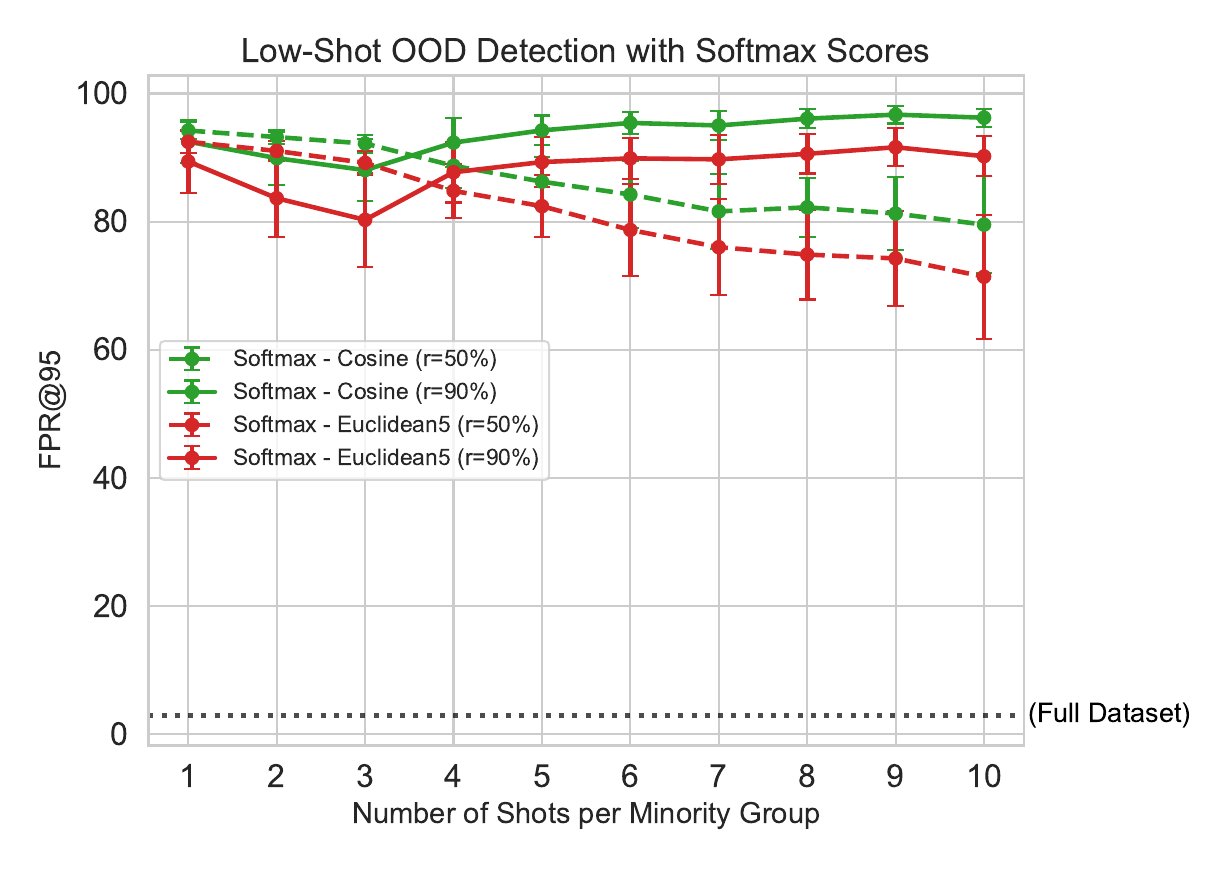}
        \vspace{-7mm}
        \caption{{\footnotesize FPR@95}}
    \end{subfigure}
    \vspace{-1mm}
    \caption{ 
Impact of softmax normalization on low-shot SP-OOD detection performance of SPROD on the Waterbirds dataset using features from a frozen CLIP ViT-B/16 vision encoder. OOD scores are derived by applying softmax to the negative distances. Performance (AUROC and FPR@95) is shown as a function of the number of samples per minority group for two spurious correlation rates (50\% and 90\%) and two distance metrics (Euclidean and Cosine). Dashed lines indicate performance with the full training set using softmax-normalized scores.
}
\vspace{-0.5mm}
\label{fig:clip:lowshot:softmax}
\end{figure}

To further investigate the sample efficiency of SPROD,
we conduct a low-shot experiment using feature embeddings from a frozen CLIP ViT-B/16 vision encoder, evaluated on the Waterbirds dataset. Two experimental settings for spurious correlation are considered: a 50\% correlation rate (where majority and minority group sample sizes are equal) and a 90\% correlation rate (where majority group samples are nine times more numerous than minority group samples). Within each setting, we vary the number of available samples per minority group and evaluate performance using both Euclidean and cosine distance metrics for score calculation.
The results are presented in Figure~\ref{fig:clip:lowshot}. Initially, with very few samples per minority group, the 90\% correlation setting exhibits slightly higher performance, which may be due to the larger initial population of majority group samples aiding prototype stability.
However, as the number of samples per minority group increases (e.g., to four samples), the 50\% correlation setting surpasses the 90\% setting in performance. Notably, the performance in both low-shot variants quickly becomes competitive with the performance achieved using the full dataset, underscoring the sample efficiency of the prototypical approach, even with CLIP features.

We extend this low-shot analysis by examining the impact of applying a softmax normalization to the distance-based OOD scores. Figure~\ref{fig:clip:lowshot:softmax} illustrates the results of this modification. A considerable performance degradation is observed when softmax normalization is applied to the distance scores, further highlighting the potential sensitivity of softmax-based scoring mechanisms.
In this softmax-normalized setting, Euclidean distance appears to yield relatively better performance compared to cosine distance, an observation that differs from the direct distance-based scoring results shown in Figure~\ref{fig:clip:lowshot}, where both distance metrics performed equally.

\section{Dataset Details and Examples}
\label{app:dataset_details}
Datasets available for studying spurious correlations are generally limited. In the context of OOD detection, this limitation becomes even more pronounced, as the task requires datasets with SP-OOD samples. As a result, the datasets employed must be thoughtfully considered to ensure meaningful evaluation.
In this study, our goal is to address a broader range of spurious features present in data, beyond just background features. A distinctive aspect of our experimental design is the inclusion of three additional settings to explore:
\begin{enumerate}
\item \textbf{Multi-spurious feature setting}: Scenarios where multiple spurious features (e.g., both background and co-occurring objects) are simultaneously present, increasing the complexity of the detection task. To the best of our knowledge, this is the first time such a setting has been explored in the context of OOD detection.
\item \textbf{Multi-class setting}: Scenarios with more than two classes, where inter-class relationships and spurious correlations introduce additional challenges.
\item \textbf{Realistic dataset setting}: Beyond existing datasets in the literature, where spurious correlations are often predefined and controlled, we also focus on scenarios leveraging realistic datasets such as AnimalsMetaCoCo and Sp-ImageNet100~\cite{neuhaus2023spuriousfeatureslargescale}. These datasets include diverse spurious correlations and more closely mimic real-world data distributions.\end{enumerate}
These aspects are relatively underexplored in the context of SP-OOD detection. By incorporating these settings, we aim to demonstrate the effectiveness of our proposed approach. 

\begin{figure}[htbp]
\centering
\begin{tabular}{c c :c}
    \begin{subfigure}{0.21\textwidth}
        \centering
        \setlength{\fboxsep}{0pt} 
        \fboxrule=0.1mm
        \fcolorbox{limegreen}{white}{\includegraphics[width = \linewidth]{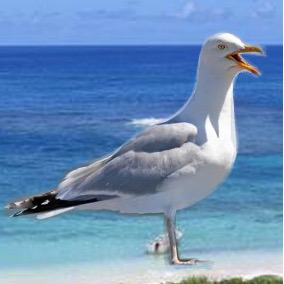}} 
        \vspace{-0.4cm}
        \caption*{\textbf{y: \textcolor{lightblue}{waterbird} \textbf{e}: \textcolor{lightblue}{water}}}
    \end{subfigure} 
    & 
    \begin{subfigure}{0.21\textwidth}
        \centering
        \setlength{\fboxsep}{0pt} 
        \fboxrule=0.1mm
        \fcolorbox{red}{white}{\includegraphics[width = \linewidth]{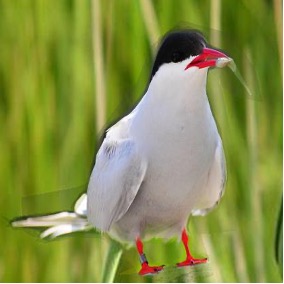}}
        \vspace{-0.4cm}
        \caption*{\textbf{y: \textcolor{lightblue}{waterbird} \textbf{e}: \textcolor{forestgreen}{land}}}
    \end{subfigure} 
    & 
    \begin{subfigure}{0.21\textwidth}
        \centering
        \setlength{\fboxsep}{0pt} 
        \fboxrule=0.1mm
        \fcolorbox{white}{white}{\includegraphics[width = \linewidth]{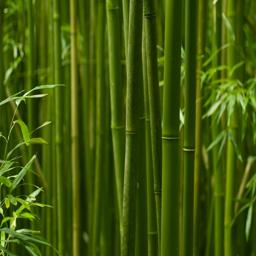}} 
        \vspace{-0.4cm}
        \caption*{\textbf{Spurious OOD-\textcolor{forestgreen}{land}}} 
    \end{subfigure} 
    \\
        \begin{subfigure}{0.21\textwidth}
        \centering
        \setlength{\fboxsep}{0pt} 
        \fboxrule=0.1mm
        \fcolorbox{red}{white}{\includegraphics[width = \linewidth]{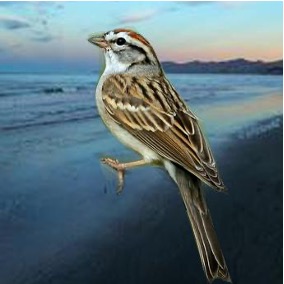}} 
        \vspace{-0.4cm}
        \caption*{\textbf{y: \textcolor{forestgreen}{landbird} \textbf{e}: \textcolor{lightblue}{water}}}
    \end{subfigure} 
    &
    \begin{subfigure}{0.21\textwidth}
        \centering
        \setlength{\fboxsep}{0pt} 
        \fboxrule=0.1mm
        \fcolorbox{limegreen}{white}{\includegraphics[width = \linewidth]{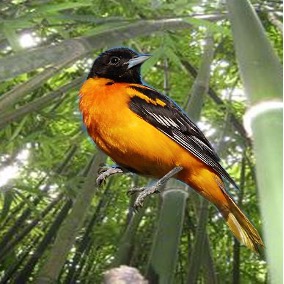}}
        \vspace{-0.4cm}
        \caption*{\textbf{y: \textcolor{forestgreen}{landbird} \textbf{e}: \textcolor{forestgreen}{land}}}
    \end{subfigure} 
    & 

    \begin{subfigure}{0.21\textwidth}
        \centering
        \setlength{\fboxsep}{0pt} 
        \fboxrule=0.1mm
        \fcolorbox{white}{white}{\includegraphics[width = \linewidth]{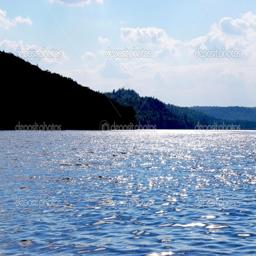}} 
        \vspace{-0.4cm}
        \caption*{\textbf{Spurious OOD-\textcolor{lightblue}{water}}} 
    \end{subfigure} 
\\
\end{tabular}

\caption{Representative examples from the Waterbirds~\cite{DR} dataset. The dataset consists of four groups: \textit{(Waterbird, Water)}, \textit{(Waterbird, Land)}, \textit{(Landbird, Water)}, and \textit{(Landbird, Land)}. Minority groups, indicated with red borders, are underrepresented, while majority groups, indicated with green borders, are more prevalent. Spurious OOD samples include only background features (land or water) without core bird-related features.}
\label{fig:dataset:waterbirds}
\end{figure}

\begin{figure}[htbp]
\centering
\begin{tabular}{c c :c}
    \begin{subfigure}{0.215\textwidth}
        \centering
        \setlength{\fboxsep}{0pt} 
        \fboxrule=0.1mm
        \fcolorbox{limegreen}{white}{\includegraphics[width = \linewidth, height = 0.18\textheight]{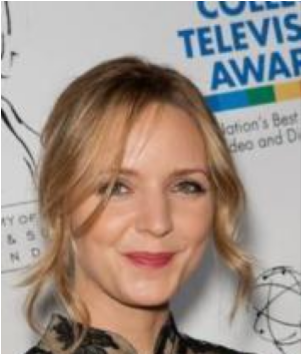}} 
        \vspace{-0.4cm}
        \caption*{\textbf{y: \textcolor{fluorescentorange}{blond hair} \textbf{e}: \textcolor{blush}{female}}}
    \end{subfigure} 
    & 

    \begin{subfigure}{0.215\textwidth}
        \centering
        \setlength{\fboxsep}{0pt} 
        \fboxrule=0.1mm
        \fcolorbox{red}{white}{\includegraphics[width = \linewidth, height = 0.18\textheight]{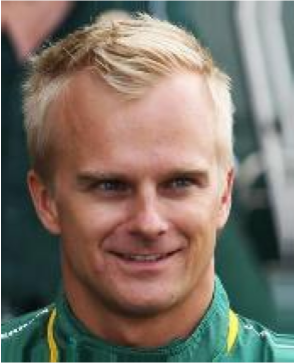}} 
        \vspace{-0.4cm}
        \caption*{\textbf{y: \textcolor{fluorescentorange}{blond hair} \textbf{e}: \textcolor{lapislazuli}{male}}}
    \end{subfigure} 
    & 
    \begin{subfigure}{0.215\textwidth}
        \centering
        \setlength{\fboxsep}{0pt} 
        \fboxrule=0.1mm
        \fcolorbox{white}{white}{\includegraphics[width = \linewidth, height = 0.18\textheight]{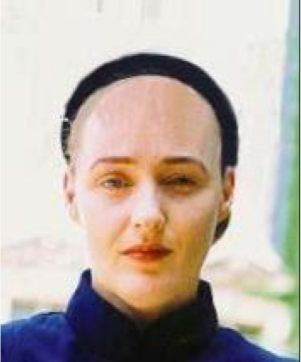}} 
        \vspace{-0.4cm}
        \caption*{\textbf{Spurious OOD-\textcolor{blush}{female}}} 
    \end{subfigure} 
    \\
    \begin{subfigure}{0.215\textwidth}
        \centering
        \setlength{\fboxsep}{0pt} 
        \fboxrule=0.1mm
        \fcolorbox{red}{white}{\includegraphics[width = \linewidth, height = 0.18\textheight]{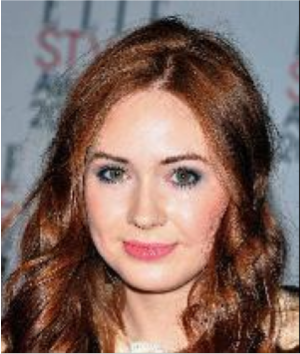}}
        \vspace{-0.4cm}
        \caption*{\textbf{y: \textcolor{cinereous}{non-blond hair} \textbf{e}: \textcolor{blush}{female}}}
    \end{subfigure} 
    &
    \begin{subfigure}{0.215\textwidth}
        \centering
        \setlength{\fboxsep}{0pt} 
        \fboxrule=0.1mm
        \fcolorbox{limegreen}{white}{\includegraphics[width = \linewidth, height = 0.18\textheight]{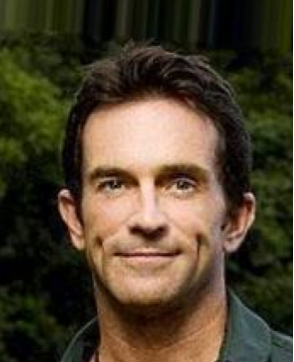}}
        \vspace{-0.4cm}
        \caption*{\textbf{y: \textcolor{cinereous}{non-blond hair} \textbf{e}: \textcolor{lapislazuli}{male}}}
    \end{subfigure} 
    & 

    \begin{subfigure}{0.215\textwidth}
        \centering
        \setlength{\fboxsep}{0pt} 
        \fboxrule=0.1mm
        \fcolorbox{white}{white}{\includegraphics[width = \linewidth, height = 0.18\textheight]{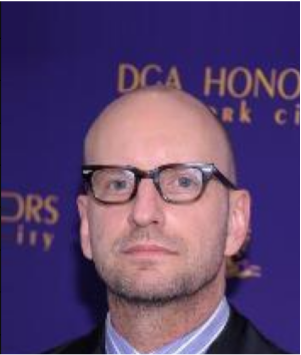}} 
        \vspace{-0.05cm}
        \caption*{\textbf{Spurious OOD-\textcolor{lapislazuli}{male}}} 
    \end{subfigure} 
\\
\end{tabular}

\caption{Representative examples from the CelebA~\cite{CelebA} dataset, which is divided into four groups: \textit{(Blond hair, Female)}, \textit{(Blond hair, Male)}, \textit{(Non-blond, Female)}, and \textit{(Non-blond, Male)}. Minority groups, marked with red borders, are underrepresented, while majority groups, highlighted with green borders, are more prevalent. Spurious OOD samples contain only spurious features (gender) without core hair color attributes.}
\label{fig:dataset:celeb}
\end{figure}

\begin{figure}[htbp]
\centering
\resizebox{\textwidth}{!}{ 
\begin{tabular}{cccc:cc}

    \begin{subfigure}{0.16\textwidth} 
        \centering
        \setlength{\fboxsep}{0pt} 
        \fboxrule=0.1mm
        \fcolorbox{limegreen}{white}{\includegraphics[width=\linewidth, height=0.12\textheight]{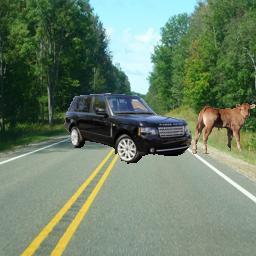}} 
        \vspace{-0.4cm}
        \caption*{\textbf{y: \textcolor{OliveGreen}{country car} \\ e1: \textcolor{YellowGreen}{country bg} \\ e2: \textcolor{JungleGreen}{country obj}}}
    \end{subfigure} 
    & 
    \begin{subfigure}{0.16\textwidth} 
        \centering
        \setlength{\fboxsep}{0pt} 
        \fboxrule=0.1mm
        \fcolorbox{red}{white}{\includegraphics[width=\linewidth, height=0.12\textheight]{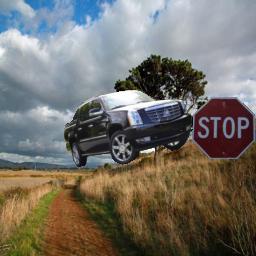}} 
        \vspace{-0.4cm}
        \caption*{\textbf{y: \textcolor{OliveGreen}{country car} \\ e1: \textcolor{YellowGreen}{country bg} \\ e2: \textcolor{fandango}{urban obj}}}
    \end{subfigure} 
    & 
    \begin{subfigure}{0.16\textwidth} 
        \centering
        \setlength{\fboxsep}{0pt} 
        \fboxrule=0.1mm
        \fcolorbox{red}{white}{\includegraphics[width=\linewidth, height=0.12\textheight]{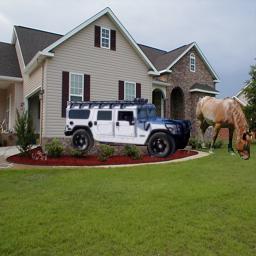}} 
        \vspace{-0.4cm}
        \caption*{\textbf{y: \textcolor{OliveGreen}{country car} \\ e1: \textcolor{mediumorchid}{urban bg} \\ e2: \textcolor{JungleGreen}{country obj}}}
    \end{subfigure} 
    & 
    \begin{subfigure}{0.16\textwidth} 
        \centering
        \setlength{\fboxsep}{0pt} 
        \fboxrule=0.1mm
        \fcolorbox{red}{white}{\includegraphics[width=\linewidth, height=0.12\textheight]{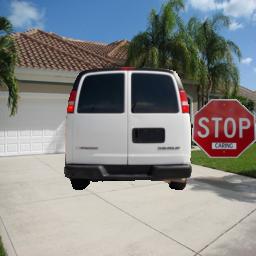}} 
        \vspace{-0.4cm}
        \caption*{\textbf{y: \textcolor{OliveGreen}{country car} \\ e1: \textcolor{mediumorchid}{urban bg} \\ e2: \textcolor{fandango}{urban obj}}}
    \end{subfigure} 
    & 
    \begin{subfigure}{0.16\textwidth} 
        \centering
        \setlength{\fboxsep}{0pt} 
        \fboxrule=0.1mm
        \fcolorbox{white}{white}{\includegraphics[width=\linewidth, height=0.12\textheight]{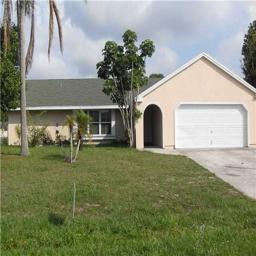}} 
        \vspace{-0.4cm}
        \caption*{\textbf{Spurious OOD - BG \\ \textcolor{mediumorchid}{urban bg}}} 
    \end{subfigure} 
    & 
    \begin{subfigure}{0.16\textwidth} 
        \centering
        \setlength{\fboxsep}{0pt} 
        \fboxrule=0.1mm
        \fcolorbox{white}{white}{\includegraphics[width=\linewidth, height=0.12\textheight]{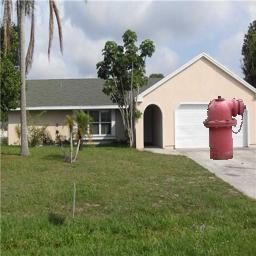}} 
        \vspace{-0.4cm}
        \caption*{\textbf{BG \& CoObj \\ \textcolor{mediumorchid}{urban bg} + \\ \textcolor{fandango}{urban obj}}} 
    \end{subfigure} 
    
    \\
    
    \begin{subfigure}{0.16\textwidth} 
        \centering
        \setlength{\fboxsep}{0pt} 
        \fboxrule=0.1mm
        \fcolorbox{red}{white}{\includegraphics[width=\linewidth, height=0.12\textheight]{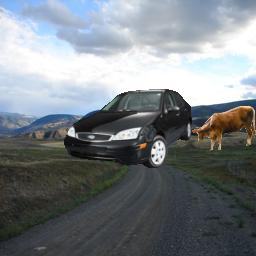}} 
        \vspace{-0.4cm}
        \caption*{\textbf{y: \textcolor{palatinatepurple}{urban car} \\ e1: \textcolor{YellowGreen}{country bg} \\ e2: \textcolor{JungleGreen}{country obj}}}
    \end{subfigure} 
    & 
    \begin{subfigure}{0.16\textwidth} 
        \centering
        \setlength{\fboxsep}{0pt} 
        \fboxrule=0.1mm
        \fcolorbox{red}{white}{\includegraphics[width=\linewidth, height=0.12\textheight]{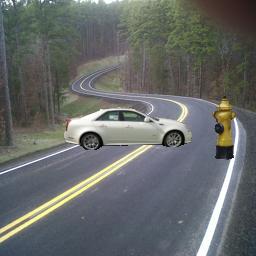}} 
        \vspace{-0.4cm}
        \caption*{\textbf{y: \textcolor{palatinatepurple}{urban car} \\ e1: \textcolor{YellowGreen}{country bg} \\ e2: \textcolor{fandango}{urban obj}}}
    \end{subfigure} 
    & 
    \begin{subfigure}{0.16\textwidth} 
        \centering
        \setlength{\fboxsep}{0pt} 
        \fboxrule=0.1mm
        \fcolorbox{red}{white}{\includegraphics[width=\linewidth, height=0.12\textheight]{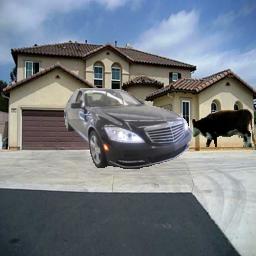}} 
        \vspace{-0.4cm}
        \caption*{\textbf{y: \textcolor{palatinatepurple}{urban car} \\ e1: \textcolor{mediumorchid}{urban bg} \\ e2: \textcolor{JungleGreen}{country obj}}}
    \end{subfigure} 
    & 
    \begin{subfigure}{0.16\textwidth} 
        \centering
        \setlength{\fboxsep}{0pt} 
        \fboxrule=0.1mm
        \fcolorbox{limegreen}{white}{\includegraphics[width=\linewidth, height=0.12\textheight]{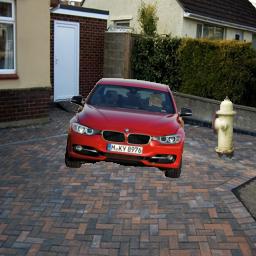}} 
        \vspace{-0.4cm}
        \caption*{\textbf{y: \textcolor{palatinatepurple}{urban car} \\ e1: \textcolor{mediumorchid}{urban bg} \\ e2: \textcolor{fandango}{urban obj}}}
    \end{subfigure} 
    & 
    \begin{subfigure}{0.16\textwidth} 
        \centering
        \setlength{\fboxsep}{0pt} 
        \fboxrule=0.1mm
        \fcolorbox{white}{white}{\includegraphics[width=\linewidth, height=0.12\textheight]{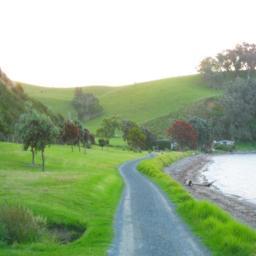}} 
        \vspace{-0.4cm}
        \caption*{\textbf{Spurious OOD - BG \\ \textcolor{YellowGreen}{country bg}}} 
    \end{subfigure} 
    & 
    \begin{subfigure}{0.16\textwidth} 
        \centering
        \setlength{\fboxsep}{0pt} 
        \fboxrule=0.1mm
        \fcolorbox{white}{white}{\includegraphics[width=\linewidth, height=0.12\textheight]{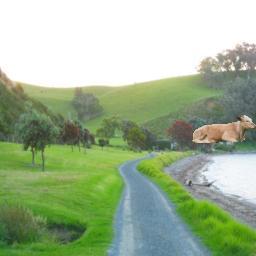}} 
        \vspace{-0.4cm}
        \caption*{\textbf{BG \& CoObj \\ \textcolor{YellowGreen}{country bg} + \\ \textcolor{JungleGreen}{country obj}}} 
    \end{subfigure} 

\end{tabular}
}
\caption{Representative examples from the UrbanCars~\cite{UrbanCars} dataset, which has 8 groups, each pairing country and urban car classes with two spurious features: background and co-occurring object, both of which can be urban or country related. The "all-country" and "all-urban" groups (green borders) dominate the dataset, while the remaining groups are significantly underrepresented (red borders). Spurious OOD samples, containing only spurious features (no cars), consist of combinations of background and co-occurring objects (BG \& CoObj) as well as background-only (BG) samples.}
\label{fig:dataset:urbancars}
\end{figure}

To evaluate the proposed approach, we used the following datasets:  
\begin{itemize}
    \item \textbf{Waterbirds}~\cite{DR}: This synthetic dataset is generated by combining the CUB~\cite{WahCUB_200_2011} (bird classes) and the Places~\cite{Places} (background scenes) datasets  for a binary classification task, with labels \( y \in \{\text{waterbird}, \text{landbird}\} \). Spurious correlations are introduced between the background \( e \in \{\text{water}, \text{land}\} \) and the label. The dataset consists of four groups, as depicted in Figure~\ref{fig:dataset:waterbirds}, with the minority and majority groups highlighted by red and green borders, respectively. Two different correlation values, \( r \in \{0.5, 0.9\} \), are employed, where \( r \) denotes the probability that the environment \( e \) aligns with the label \( y \). Specifically, we have:
\[
\scriptsize r = P(e = \text{water} \mid y = \text{waterbird}) = P(e = \text{land} \mid y = \text{landbird}).
\]

The distribution of samples within each group and class is provided in Table~\ref{tab:wb_data}. For SP-OOD, we select samples from the Places dataset~\cite{Places}, following the previous works~\cite{main_ref, rw}.

    \item \textbf{CelebA}~\cite{CelebA}: This dataset is used for a binary classification task with labels \( y \in \{\text{blond hair}, \text{non-blond hair}\} \), where spurious correlations with gender \( \in \{\text{male}, \text{female}\} \) are present. It is a real-world dataset, making it suitable for evaluating models in realistic settings. In the dataset, most females have blond hair, and most males have non-blond hair, forming the majority groups, as shown in Figure~\ref{fig:dataset:celeb} with colored borders. The minority groups are the opposite combinations. The distribution of groups across classes with varying correlation levels is presented in Table~\ref{tab:clb_data}. In this setting, SP-OOD samples are those with no hair, but still exhibiting the spurious gender-related features. For our SP-OOD samples, we used bald males, who lack core features but retain spurious (gender) cues.

    \item \textbf{UrbanCars}~\cite{UrbanCars}: This dataset, introduced in~\cite{UrbanCars}, is synthetically generated by combining the Stanford Cars dataset~\cite{6755945} (which includes both urban and country cars) with co-occurring objects from either urban or country environments, sourced from the LVIS dataset~\cite{gupta2019lvisdatasetlargevocabulary}. The co-occurring objects are positioned to the right of the cars, and both the car and the object are placed onto background images from the Places dataset~\cite{Places}, representing either urban or country scenes. The dataset is considered particularly challenging due to the presence of two spurious features. By varying the combinations of cars, co-occurring objects, and backgrounds, the dataset is divided into eight distinct groups, as illustrated in Figure~\ref{fig:dataset:urbancars}. Notably, the "all-country" and "all-urban" groups dominate the dataset, while the other groups are underrepresented, as highlighted by the border colors. The exact number of samples in each group of our generated dataset, which exhibits a 95\% correlation,  is detailed in Table~\ref{tab:urban_cars_data}, with two groups being especially underrepresented. For the SP-OOD analysis, we sample combinations of backgrounds and co-occurring objects, called the BG \& CoObj setting, as well as backgrounds alone, referred to as the BG setting. In the results section, we reported the more challenging scenario(BG \& CoObj) as it presented greater difficulty, as expected.

    \item \textbf{AnimalsMetaCoCo}: 
    To create a multi-class, multi-valued spurious attribute setting that reflects realistic and challenging scenarios, we selected a subset of 26 animal classes from the MetaCoCo dataset~\cite{MetaCoCo}. The samples were first cleaned through relabeling, removal of duplicates, and deletion of irrelevant images. Subsequently, we defined subattributes to create a total of 8 major concepts for modeling attribute imbalance and spurious features, as illustrated in the first Figure of the paper. We introduce this new dataset as \textbf{AnimalsMetaCoCo}, a refined subset tailored for multi-class, multi-valued spurious feature scenarios. Details of the classes, attributes, and group sample distributions are provided in Table~\ref{tab:animal_data}. This dataset represents a more realistic scenario, where each class is associated with at least one spurious feature similar to those present in ID data. 
    While the strength of spurious correlations may not be as pronounced as in the other three datasets, AnimalsMetaCoCo introduces several unique and challenging aspects. Specifically, its spurious attributes can take on multiple distinct values, each exhibiting varying degrees of imbalance across classes. Moreover, the class distributions themselves are inherently imbalanced. To construct SP-OOD scenarios, we adopt a leave-2-out strategy: in each round, two classes are treated as SP-OOD, characterized by semantic shifts while retaining at least one spurious attribute shared with ID data, and the remaining classes serve as ID. This setup significantly increases the difficulty of the detection task due to overlapping spurious patterns across environments.

    \item \textbf{Sp-ImageNet100}: 
    We evaluate models on the \textbf{Spurious ImageNet (SpI)} dataset introduced in ~\cite{neuhaus2023spuriousfeatureslargescale}. This dataset contains real-world images (from OpenImages~\cite{openimages} and Flickr) that include only spurious features, such as bird feeders or graffiti, without the actual class object. These images are consistently misclassified as specific ImageNet classes, revealing harmful spurious correlations.

SpI focuses on 100 ImageNet classes (we call Sp-ImageNet100) where such correlations are prevalent. The authors distinguish between two types of harmful spurious features:
\begin{itemize}
    \item \textbf{Spurious Class Extension:} A spurious feature alone causes a class prediction (e.g., bird feeder $\rightarrow$ hummingbird).
    \item \textbf{Spurious Shared Feature:} A feature shared across classes biases prediction toward one due to imbalance (e.g., water jet $\rightarrow$ fireboat over fountain).
\end{itemize}

We note that most spurious features in SpI are of the class extension type. This is less aligned with our goal of identifying underrepresented groups based on \textbf{shared spurious features} and misclassification signals~\cite{wg3}, and without group supervision.

Nevertheless, we include SpI as a challenging, naturally occurring OOD benchmark for evaluating model robustness to spurious correlations.

\end{itemize}

For NSP-OOD, we utilized the SVHN~\cite{SVHN}, LSUN~\cite{lsun}, and iSUN~\cite{isun} datasets, which are commonly employed in prior works~\cite{main_ref, rw}. These datasets are used consistently as NSP-OOD for all the mentioned ID datasets.

\input{appendix_tables/waterbirds_detail}

\input{appendix_tables/celeba_detail}
\input{appendix_tables/urbancars_detail}
\input{appendix_tables/animals_detail}

\section{Performance on Non-Spurious OOD Datasets}
\label{sec:far_NSP_OOD_performance}

\input{far_ood_tables/ResNet-50_AUROC}
\input{far_ood_tables/ResNet-50_AUPR}

We evaluate the performance of post-hoc methods in the non-spurious OOD (NSP-OOD) detection setting. For this evaluation, the OOD samples are drawn from a collection of standard benchmark datasets: SVHN~\cite{SVHN}, LSUN (resized)\cite{lsun}, and iSUN\cite{isun}. The ID context for each evaluation varies across the five datasets used in our primary spurious correlation experiments: Waterbirds (WB), CelebA (CA), UrbanCars (UC), Animals MetaCoCo (AMC), and Spurious ImageNet (SpI). Specifically, for each of these five ID contexts, the pretrained ResNet-50 backbone is employed, and the post-hoc OOD detection methods are subsequently set up using the features derived from this backbone.

The performance metrics reported in Tables \ref{tab:nspood_auroc_fpr} and \ref{tab:nspood_aupr} are the average performance when distinguishing each ID dataset from the NSP-OOD samples, averaged across SVHN, LSUN, and iSUN. 
This setup assesses the general OOD detection capability of the post-hoc methods when the primary challenge is not spurious correlations shared between ID and OOD, but the ID datasets still contain inherent biases.
As observed in the tables, distance-based methods such as SPROD and KNN consistently achieve near-perfect scores across all metrics (AUROC, FPR@95, AUPR-In, and AUPR-Out) and ID contexts. Other methods like VIM and MDS also show strong performance. 

\section{Ablation Study on SPROD Stages}
\label{sec:SPROD_ablation}

To understand the contribution of each stage in our proposed SPROD method, we conduct an ablation study. We evaluate the performance of:
\begin{itemize}
\item Stage 1: Initial Prototype Construction
\item Stage 2: Classification-Aware Prototype Calculation
\item Stage 3: Group Prototype Refinement
\end{itemize}
For this analysis, we select the Waterbirds dataset, as its synthetic nature allows for precise control over spurious correlations and clearly exemplifies the SP-OOD challenge by design. We evaluate on two versions of Waterbirds: one with a 50\% spurious correlation rate (where spurious features are less effective) and another with a 90\% correlation rate (representing a strong spurious bias).
Experiments are conducted using features from pretrained ResNet-50 and ResNet-18 backbones, without fine-tuning on Waterbirds, to isolate the effect of the prototype refinement stages.

\begin{figure}[!ht]
    \centering
    \begin{subfigure}[b]{0.34\textwidth}
        \centering
        \includegraphics[width=\textwidth]{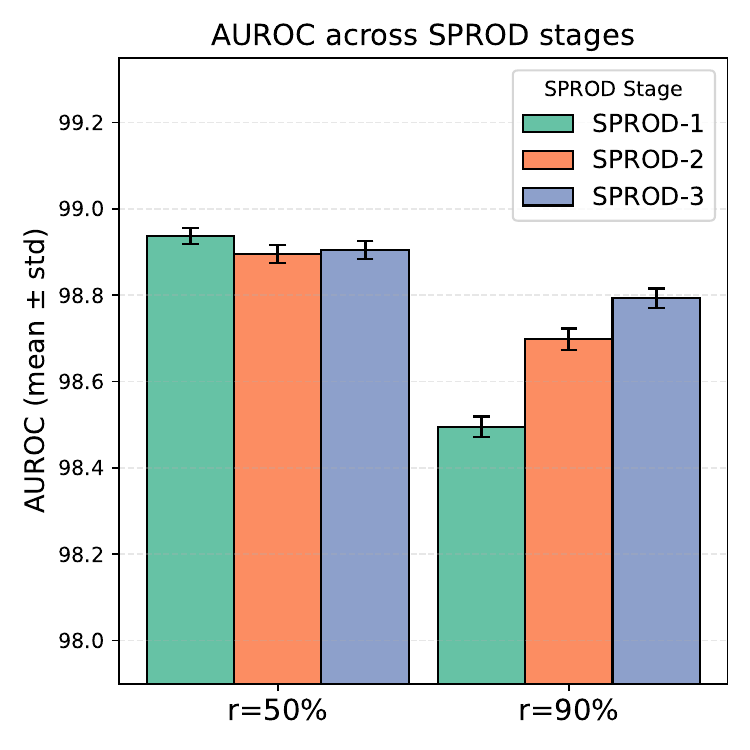}
        \vspace{-7mm}
        \caption{{\footnotesize AUROC}}
    \end{subfigure}
    \begin{subfigure}[b]{0.34\textwidth}
        \centering
        \includegraphics[width=\textwidth]{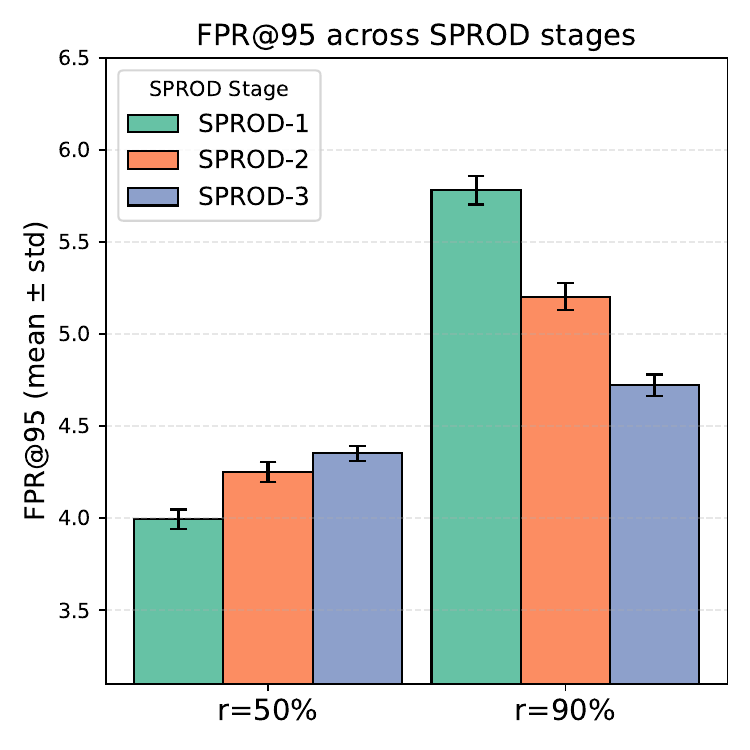}
        \vspace{-7mm}
        \caption{{\footnotesize FPR@95}}
    \end{subfigure}
    \vspace{-1mm}
    \caption{ 
Ablation study of SPROD stages on the Waterbirds dataset using ResNet-50 features. Results are shown for two spurious correlation settings: 50\% and 90\%.
}
\vspace{-0.5mm}
\label{fig:stages_ablation:resnet50}
\end{figure}

\begin{figure}[!ht]
    \centering
    \begin{subfigure}[b]{0.37\textwidth}
        \centering
        \includegraphics[width=\textwidth]{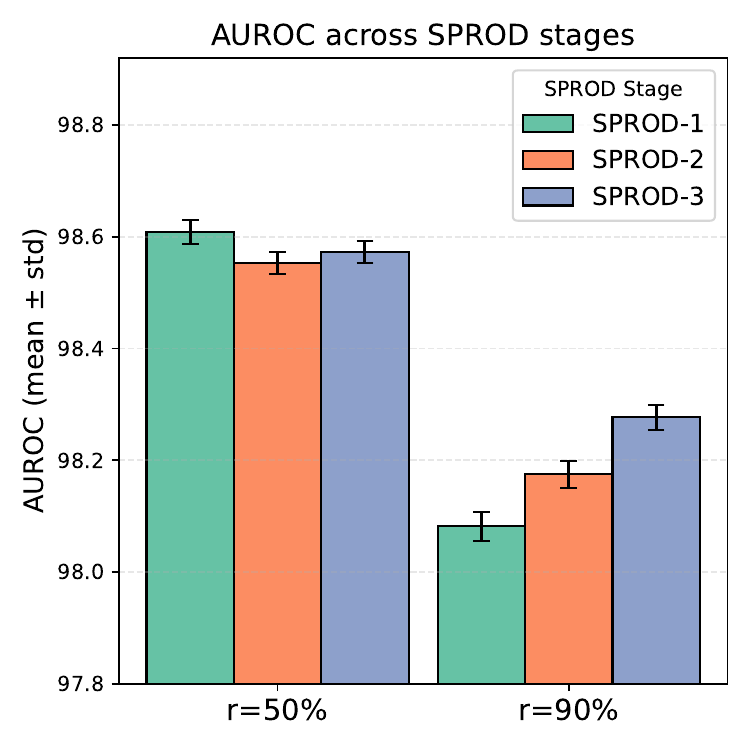}
        \vspace{-7mm}
        \caption{{\footnotesize AUROC}}
    \end{subfigure}
    \begin{subfigure}[b]{0.37\textwidth}
        \centering
        \includegraphics[width=\textwidth]{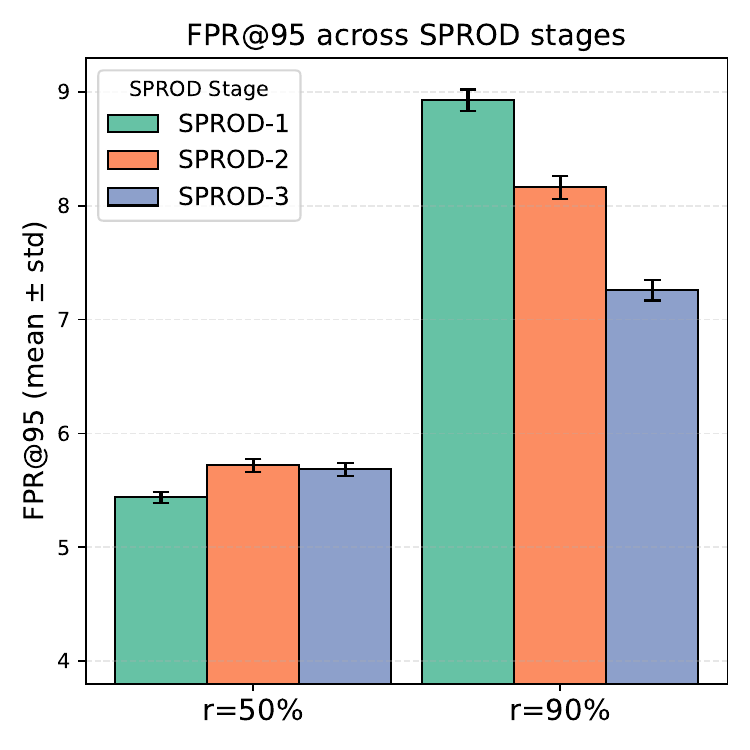}
        \vspace{-7mm}
        \caption{{\footnotesize FPR@95}}
    \end{subfigure}
    \vspace{-1mm}
    \caption{ 
Ablation study of SPROD stages on the Waterbirds dataset using ResNet-18 features. Results are shown for two spurious correlation settings: 50\% and 90\%.
}
\vspace{-0.5mm}
\label{fig:stages_ablation:resnet18}
\end{figure}

The results of this ablation study are presented in Figure~\ref{fig:stages_ablation:resnet50} and Figure~\ref{fig:stages_ablation:resnet18}.
As seen, the performance of the simple initial prototypical approach (Stage 1) performs competitively, especially on the 50\% correlation setting. This suggests that a basic prototypical method, which computes distances to class means in feature space, is a competitive baseline for OOD detection that has been somewhat overlooked in existing literature.
When the spurious correlation rate is increased to 90\%, we observe a general reduction in OOD detection performance across all three variants. This is expected, as stronger spurious correlations make it more difficult to distinguish true class features from misleading cues.
However, Stage 3 (the full SPROD method) is significantly more robust to the increase of spurious correlation. 

\section{Backbone Experiments}
\label{sec:backbone_ablation}

To assess the generality and robustness of SPROD across different neural network architectures, we evaluate its performance using a wide range of feature backbones. This analysis complements the main paper’s results, which primarily rely on ResNet-50~\cite{ResNet}, and demonstrates that our method remains effective across both convolutional and transformer-based representations.

We include a comprehensive selection of backbones commonly used in the OOD detection literature. This includes all major variants of the ResNet~\cite{ResNet} family, ResNet-18 (R18), ResNet-34 (R34), ResNet-50 (R50), and ResNet-101 (R101), owing to their widespread adoption and varying representational capacities. Alongside these, we evaluate several modern transformer-based architectures with diverse embedding sizes and training paradigms, such as the self-supervised DINOv2-S (DINOv2)~\cite{DINOv2}, the standard supervised ViT-S (ViT)~\cite{dosovitskiy2021imageworth16x16words}, and more hierarchical or data-efficient designs like Swin-B (Swin)~\cite{swint}, DeiT-B (DeiT)~\cite{dosovitskiy2021imageworth16x16words}, ConvNeXt-B (CvNXt)~\cite{convnext}, and BiT-R50x1 (BiT)~\cite{BiT}. 

We analyze the performance of each method using four complementary metrics (AUROC, FPR@95, AUPR-IN, and AUPR-OUT) to evaluate different aspects of OOD detection performance.
Tables~\ref{tab:auroc_all}, \ref{tab:fpr_all}, \ref{tab:auprin_all}, and \ref{tab:auprout_all} summarize the performance of our post-hoc method across various backbone architectures, averaged over the five SP-OOD datasets studied in this work. Detailed results for each individual backbone across the different datasets are presented in the subsequent tables.

\input{backbone_tables/AUROC_all}
\input{backbone_tables/FPR_all}

\input{backbone_aupr_tables/AUPR_IN_all}
\input{backbone_aupr_tables/AUPR_OUT_all}

Overall, we observe that SPROD consistently achieves strong performance across all backbone architectures, often with a notable margin. KNN emerges as the second-best method, suggesting that simpler approaches can be counterintuitively effective; however, its performance is highly sensitive to the choice of hyperparameters. MDS is also among the top-performing methods; it is a metric-based approach similar to SPROD but employs a more complex model by estimating class-specific covariance matrices. This added complexity, while potentially beneficial, may increase the risk of overfitting, especially in OOD settings. In addition, output-based methods show notable drops in performance on several benchmarks, highlighting their susceptibility to distributional shifts.
Notably, SPROD is the only method that maintains stable performance across all evaluated settings, without experiencing major degradation under any backbone or dataset configuration.
Among the evaluated backbones, ConvNeXt-B and Swin-B stand out as frozen feature extractors with superior performance for post-hoc OOD detection.

\input{backbone_tables/ResNet-18}
\input{backbone_tables/ResNet-34}
\input{backbone_tables/ResNet-50}
\input{backbone_tables/ResNet-101}
\input{backbone_tables/DINOv2-S}
\input{backbone_tables/ViT-S}
\input{backbone_tables/Swin-B}
\input{backbone_tables/DeiT-B}
\input{backbone_tables/ConvNeXt-B}
\input{backbone_tables/BiT-R50x1}

\input{backbone_aupr_tables/ResNet-18}
\input{backbone_aupr_tables/ResNet-34}
\input{backbone_aupr_tables/ResNet-50}
\input{backbone_aupr_tables/ResNet-101}
\input{backbone_aupr_tables/DINOv2-S}
\input{backbone_aupr_tables/ViT-S}
\input{backbone_aupr_tables/Swin-B}
\input{backbone_aupr_tables/DeiT-B}
\input{backbone_aupr_tables/ConvNeXt-B}
\input{backbone_aupr_tables/BiT-R50x1}

\section{Analysis of Mixture of Prototypes}
\label{sec:mixture_of_prototypes}

To evaluate the role of prototype augmentation and refinement, here we analyze two new variants of SPROD, comparing them against our standard method (referred to as SPROD-Default in this section).
The first is SPROD-KMeans, a clustering baseline where embeddings of training samples within each class are clustered using K-Means. The number of centroids per class is set to match the number of group prototypes that SPROD-Default would derive. This tests whether the benefits of SPROD-Default stem solely from using multiple prototypes per class or from its proposed classification-driven refinement strategy. Recent works, such as Prototypical Learning with a Mixture of Prototypes (PALM) \cite{palm}, demonstrate the importance of multiple prototypes to capture intra-class diversity. SPROD-Kmeans serves as a simpler post-hoc method based on the K-Means algorithm in this context.

We also introduce SPROD-Converged, which iteratively refines the group prototypes derived from SPROD-Default's Stage 2 by repeatedly applying the Stage 3 reassignment and recalculation steps until centroid convergence, similar to the K-Means algorithm. The centroid initialization mechanism distinguishes SPROD-Converged from SPROD-Kmeans, where K-Means clustering is typically initialized using a method like K-Means++ directly on the raw class samples.

As shown in Tables~\ref{tab:clustering1} and~\ref{tab:clustering2}, SPROD-Default generally outperforms the two other variants when the amount of spurious correlation in the ID data is high (e.g., Waterbirds, CelebA, and UrbanCars); on other datasets, their performances are competitive.
Proto-KMeans is slightly less effective in separating SP-OOD samples, likely due to its reliance on purely geometric clustering, whereas SPROD with classification-aware refinement slightly enhances the robustness when the spurious correlation is high.

\input{appendix_tables/kmeans_tables}

\section{Reproducibility and Resources}

The data (cleaned AnimalsMetaCoCo dataset)  and code for our approach are available at the following GitHub repository: \url{https://github.com/ReihanehZohrabi/SPROD}. 
In our benchmarking, we utilized components of the OpenOOD v1.5~\cite{zhang2023openood,openood, survey_2} framework to obtain results for previously proposed OOD detection methods.

Our experiments were designed to be post-hoc and computationally efficient. All experiments were conducted on a single GeForce RTX 3090 Ti GPU, demonstrating the method's practicality in terms of resource requirements.

%% file: appendix_tables/related_comparisons.tex
\begin{table}[ht]
  \centering
  \caption{A comparative analysis of AUROC and FPR@95 performance metrics for different methods and models evaluated on the Waterbirds dataset. To ensure a fair comparison, we report our results using both the pretrained CLIP and ResNet-18 models, aligning with the settings of the compared methods.}
  \resizebox{\textwidth}{!}{%
  \begin{tabular}{l c c c c}
    \specialrule{1.5pt}{1pt}{1pt}
    \textbf{Method} & \textbf{Backbone} & \textbf{AUROC $\uparrow$} & \textbf{FPR@95 $\downarrow$} & \textbf{Notes} \\
    \specialrule{1.5pt}{1pt}{1pt}
    {Backmix~\cite{backmix} + Energy} & WideResNet40-4 & 80.6 & 81.7 & \\
    {ImOOD~\cite{imood}} & \multirow{3}{*}{ResNet-18} & 83.27 & 57.69 & \\
    {RW~\cite{rw} + MD} & & <90 & -- & Exact AUROC not clear from plots \\
    \rowcolor{gray!10} \textbf{SPROD} &  & \textbf{98.28} & \textbf{7.27} & \\
    \specialrule{1pt}{1pt}{1pt}
    CoVer~\cite{cover} & \multirow{6}{*}{CLIP-B/16} & 90.52 & 33.17 & Also reports MCM: 90.31 / 25.66 \\
    MCM~\cite{MCM} &  & 98.36 & 5.87 & \\
    Dai et al.~\cite{dai-etal-2023-exploring} &  & 98.62 & 4.56 & \\
    Neglabel~\cite{neglabel} &  & 94.67 & 9.5 & Also reports MCM: 93.30 / 14.45 \\
    CMA~\cite{CMA} &  & \textbf{99.01} & 3.22 & \\
    \rowcolor{gray!10} \textbf{SPROD} & & \textbf{99.01} & \textbf{2.94} &\\
    \specialrule{1pt}{1pt}{1pt}
  \end{tabular}}

  \label{tab:ood_comparisons}
\end{table}

%% file: appendix_tables/waterbirds_detail.tex
\begin{table}[!htbp]
  \centering
  \small
\caption{Group-wise distribution of the \textbf{Waterbirds}~\cite{DR} training set across land and water attributes at varying correlation levels. The distribution reflects the degree of alignment between bird classes (landbird or waterbird) and their respective backgrounds, with higher correlation level indicating a stronger dependence between the bird label and the background.}
      \label{tab:wb_data}
  \begin{tabular}{ccccc}
    \specialrule{1.5pt}{1pt}{1pt}
    \textbf{Correlation} & \textbf{SP-Feature} & \textbf{Landbird} & \textbf{Waterbird} & \textbf{Total (Row)} \\
    \specialrule{1.5pt}{1pt}{1pt}
    \multirow{2}{*}{50\%} & Land & 544 & 1853 & \textbf{2397} \\
    & Water & 545 & 1853 & \textbf{2398} \\
    \cmidrule{2-5}
    & \textbf{Total (Col)} & \textbf{1089} & \textbf{3706} & \textbf{4795} \\
    \midrule
    \multirow{2}{*}{90\%} & Land & 997 & 369 & \textbf{1366} \\
    & Water & 111 & 3318 & \textbf{3429} \\
    \cmidrule{2-5}
    & \textbf{Total (Col)} & \textbf{1108} & \textbf{3687} & \textbf{4795} \\
    \bottomrule
  \end{tabular}
\end{table}

%% file: appendix_tables/celeba_detail.tex
\begin{table}[!htbp]
  \centering
  \small
\caption{Distribution of the \textbf{CelebA}~\cite{CelebA} training set across male and female attributes at varying correlation levels. The correlation levels indicate the strength of the spurious relationship between gender and the presence of blond hair, with higher correlations reflecting a stronger association between these attributes in the dataset.}
  \label{tab:clb_data}
  \begin{tabular}{ccccc}
    \specialrule{1.5pt}{1pt}{1pt}
    \textbf{Correlation} & \textbf{SP-Feature} & \textbf{Blond} & \textbf{Non-blond} & \textbf{Total (Row)} \\
    \specialrule{1.5pt}{1pt}{1pt}
    \multirow{2}{*}{50\%} & Male & 1387 & 1387 & \textbf{2774} \\
    & Female & 1387 & 1387 & \textbf{2774} \\
    \cmidrule{2-5}
    & \textbf{Total (Col)} & \textbf{2774} & \textbf{2774} & \textbf{5548} \\
    \midrule
    \multirow{2}{*}{90\%} & Male & 296 & 2468 & \textbf{2764} \\
    & Female & 2474 & 310 & \textbf{2784} \\
    \cmidrule{2-5}
    & \textbf{Total (Col)} & \textbf{2770} & \textbf{2778} & \textbf{5548} \\
    \bottomrule
  \end{tabular}
\end{table}

%% file: appendix_tables/urbancars_detail.tex
\begin{table}[!htbp]
  \centering
  \small
  \caption{Distribution of \textbf{UrbanCars}~\cite{UrbanCars} training samples with 95\% correlation across groups, categorized by background and co-occurring object features within each class. This dataset features six minority groups out of eight possible combinations, highlighting its challenging nature due to the underrepresentation of most groups.}
  \label{tab:urban_cars_data}
  \begin{tabular}{cccc}
    \specialrule{1.5pt}{1pt}{1pt}
    \textbf{SP-Features} & \textbf{Country Car} & \textbf{Urban Car} & \textbf{Total (Row)} \\
    \specialrule{1.5pt}{1pt}{1pt}
    Country BG, Country CoObj & 3606 & 10 & \textbf{3616} \\
    Country BG, Urban CoObj & 190 & 190 & \textbf{380} \\
    Urban BG, Country CoObj & 190 & 189 & \textbf{379} \\
    Urban BG, Urban CoObj & 10 & 3605 & \textbf{3615} \\
    \cmidrule{1-4}
    \textbf{Total (Col)} & \textbf{3996} & \textbf{3994} & \textbf{7990} \\
    \bottomrule
  \end{tabular}
\end{table}

%% file: appendix_tables/animals_detail.tex
\begin{table}
{
\small
\caption{Group-wise distribution of 26 selected animal classes from the MetaCoCo~\cite{MetaCoCo} dataset called \textbf{Animals MetaCoCo} across various environments.}
\label{tab:animal_data}
\begin{tabular}{lccccccccc}
\specialrule{1.5pt}{1pt}{1pt}
Class \textbackslash  Attribute& at home & autumn & dim & grass & in cage & on snow & rock & water & Total \\
\specialrule{1.5pt}{1pt}{1pt}
bear & 0 & 109 & 76 & 219 & 0 & 98 & 98 & 362 & 962 \\
cat & 284 & 159 & 150 & 439 & 95 & 102 & 121 & 386 & 1736 \\
cow & 57 & 137 & 119 & 693 & 0 & 122 & 38 & 221 & 1387 \\
crab & 0 & 0 & 43 & 73 & 0 & 0 & 153 & 102 & 371 \\
dog & 106 & 221 & 174 & 561 & 111 & 208 & 145 & 440 & 1966 \\
dolphin & 0 & 0 & 113 & 0 & 0 & 0 & 15 & 425 & 553 \\
elephant & 108 & 98 & 247 & 434 & 0 & 59 & 38 & 375 & 1359 \\
fox & 0 & 195 & 62 & 367 & 66 & 160 & 136 & 145 & 1131 \\
frog & 0 & 232 & 3 & 530 & 0 & 0 & 322 & 342 & 1429 \\
giraffe & 0 & 479 & 150 & 397 & 0 & 0 & 78 & 227 & 1331 \\
goose & 0 & 105 & 135 & 378 & 54 & 0 & 83 & 263 & 1018 \\
horse & 57 & 366 & 128 & 672 & 0 & 129 & 59 & 457 & 1868 \\
kangaroo & 0 & 84 & 190 & 214 & 0 & 55 & 61 & 97 & 701 \\
lion & 0 & 560 & 58 & 537 & 36 & 79 & 275 & 171 & 1716 \\
lizard & 0 & 130 & 42 & 303 & 35 & 0 & 302 & 242 & 1054 \\
monkey & 0 & 183 & 84 & 592 & 76 & 100 & 463 & 255 & 1753 \\
ostrich & 0 & 164 & 125 & 235 & 159 & 76 & 73 & 144 & 976 \\
owl & 0 & 141 & 147 & 131 & 36 & 92 & 78 & 87 & 712 \\
rabbit & 0 & 147 & 31 & 637 & 105 & 134 & 91 & 73 & 1218 \\
rat & 110 & 0 & 0 & 123 & 52 & 41 & 0 & 66 & 392 \\
seal & 0 & 57 & 31 & 158 & 19 & 266 & 240 & 547 & 1318 \\
sheep & 0 & 626 & 30 & 865 & 0 & 99 & 207 & 237 & 2064 \\
squirrel & 0 & 212 & 32 & 418 & 0 & 132 & 188 & 118 & 1100 \\
tiger & 0 & 212 & 14 & 435 & 66 & 176 & 236 & 323 & 1462 \\
tortoise & 0 & 129 & 140 & 284 & 0 & 0 & 157 & 234 & 944 \\
wolf & 0 & 192 & 97 & 198 & 90 & 151 & 188 & 188 & 1104 \\
\midrule
Total & 722 & 4938 & 2421 & 9893 & 1000 & 2279 & 3845 & 6527 & \textbf{31625} \\
\bottomrule
\end{tabular}}
\end{table}

%% file: far_ood_tables/ResNet-50_AUROC.tex
\begin{table}[t]
    \centering
    \caption{Average NSP-OOD detection performance (AUROC and FPR@95) using a ResNet-50 backbone. Cell values are performance metrics averaged over SVHN, LSUN, and iSUN as OOD.}
    \label{tab:nspood_auroc_fpr}
    \begin{minipage}{0.49\textwidth}
      \centering
      {\scriptsize\textbf{AUROC}$\uparrow$} \\
      \resizebox{\linewidth}{!}{
\begin{tabular}{lcccccc}
\toprule
Method & WB & CA & UC & AMC & SpI & Avg. \\
\midrule
MSP & $86.2_{\textcolor{gray}{\pm1.0}}$ & $77.5_{\textcolor{gray}{\pm0.9}}$ & $52.6_{\textcolor{gray}{\pm4.0}}$ & $92.6_{\textcolor{gray}{\pm0.2}}$ & $94.9_{\textcolor{gray}{\pm0.3}}$ & $80.8$ \\
Energy & $85.0_{\textcolor{gray}{\pm6.4}}$ & $75.4_{\textcolor{gray}{\pm6.3}}$ & $53.2_{\textcolor{gray}{\pm8.0}}$ & $96.8_{\textcolor{gray}{\pm0.1}}$ & $97.7_{\textcolor{gray}{\pm0.3}}$ & $81.6$ \\
MLS & $85.7_{\textcolor{gray}{\pm4.2}}$ & $76.4_{\textcolor{gray}{\pm4.1}}$ & $53.0_{\textcolor{gray}{\pm6.3}}$ & $96.2_{\textcolor{gray}{\pm0.1}}$ & $96.3_{\textcolor{gray}{\pm0.3}}$ & $81.5$ \\
KLM & $57.7_{\textcolor{gray}{\pm2.0}}$ & $52.5_{\textcolor{gray}{\pm2.1}}$ & $47.5_{\textcolor{gray}{\pm1.6}}$ & $90.1_{\textcolor{gray}{\pm0.3}}$ & $96.1_{\textcolor{gray}{\pm0.5}}$ & $68.8$ \\
GNorm & $90.7_{\textcolor{gray}{\pm0.6}}$ & $70.8_{\textcolor{gray}{\pm0.9}}$ & $64.1_{\textcolor{gray}{\pm2.1}}$ & $97.3_{\textcolor{gray}{\pm0.1}}$ & $99.0_{\textcolor{gray}{\pm0.0}}$ & $84.4$ \\
ReAct & $86.8_{\textcolor{gray}{\pm5.2}}$ & $66.1_{\textcolor{gray}{\pm7.5}}$ & $50.0_{\textcolor{gray}{\pm8.7}}$ & $96.3_{\textcolor{gray}{\pm0.1}}$ & $99.0_{\textcolor{gray}{\pm0.3}}$ & $79.6$ \\
VIM & $\textbf{100.0}_{\textcolor{gray}{\pm0.0}}$ & $\textbf{100.0}_{\textcolor{gray}{\pm0.0}}$ & $99.6_{\textcolor{gray}{\pm0.4}}$ & $99.9_{\textcolor{gray}{\pm0.0}}$ & $99.6_{\textcolor{gray}{\pm0.0}}$ & $99.8$ \\
MDS & $\textbf{100.0}_{\textcolor{gray}{\pm0.0}}$ & $\textbf{100.0}_{\textcolor{gray}{\pm0.0}}$ & $\textbf{100.0}_{\textcolor{gray}{\pm0.0}}$ & $99.4_{\textcolor{gray}{\pm0.0}}$ & $97.5_{\textcolor{gray}{\pm0.0}}$ & $99.4$ \\
RMDS & $58.6_{\textcolor{gray}{\pm0.4}}$ & $68.6_{\textcolor{gray}{\pm2.9}}$ & $48.9_{\textcolor{gray}{\pm0.6}}$ & $93.0_{\textcolor{gray}{\pm0.2}}$ & $97.1_{\textcolor{gray}{\pm0.1}}$ & $73.2$ \\
KNN & $\textbf{100.0}_{\textcolor{gray}{\pm0.0}}$ & $\textbf{100.0}_{\textcolor{gray}{\pm0.0}}$ & $\textbf{100.0}_{\textcolor{gray}{\pm0.0}}$ & $\textbf{100.0}_{\textcolor{gray}{\pm0.0}}$ & $99.8_{\textcolor{gray}{\pm0.0}}$ & $\textbf{100.0}$ \\
SHE & $96.4_{\textcolor{gray}{\pm0.2}}$ & $98.6_{\textcolor{gray}{\pm0.2}}$ & $99.1_{\textcolor{gray}{\pm0.1}}$ & $94.3_{\textcolor{gray}{\pm0.2}}$ & $98.5_{\textcolor{gray}{\pm0.1}}$ & $97.4$ \\
\midrule
\textbf{SPROD} & $\textbf{100.0}_{\textcolor{gray}{\pm0.0}}$ & $\textbf{100.0}_{\textcolor{gray}{\pm0.0}}$ & $\textbf{100.0}_{\textcolor{gray}{\pm0.0}}$ & $\textbf{100.0}_{\textcolor{gray}{\pm0.0}}$ & $\textbf{99.9}_{\textcolor{gray}{\pm0.0}}$ & $\textbf{100.0}$ \\
\bottomrule
\end{tabular}
}
    \end{minipage}
    \hfill
    \begin{minipage}{0.49\textwidth}
      \centering
      {\scriptsize\textbf{FPR@95}$\downarrow$} \\
      \resizebox{\linewidth}{!}{
\begin{tabular}{lcccccc}
\toprule
Method & WB & CA & UC & AMC & SpI & Avg. \\
\midrule
MSP & $76.5_{\textcolor{gray}{\pm1.6}}$ & $85.2_{\textcolor{gray}{\pm2.7}}$ & $95.6_{\textcolor{gray}{\pm0.4}}$ & $46.1_{\textcolor{gray}{\pm1.5}}$ & $34.8_{\textcolor{gray}{\pm3.1}}$ & $67.6$ \\
Energy & $74.5_{\textcolor{gray}{\pm24.8}}$ & $84.4_{\textcolor{gray}{\pm14.5}}$ & $90.8_{\textcolor{gray}{\pm9.7}}$ & $19.9_{\textcolor{gray}{\pm1.1}}$ & $13.4_{\textcolor{gray}{\pm2.2}}$ & $56.6$ \\
MLS & $76.3_{\textcolor{gray}{\pm15.0}}$ & $86.9_{\textcolor{gray}{\pm9.6}}$ & $93.8_{\textcolor{gray}{\pm5.0}}$ & $26.3_{\textcolor{gray}{\pm1.3}}$ & $24.5_{\textcolor{gray}{\pm2.5}}$ & $61.6$ \\
KLM & $80.5_{\textcolor{gray}{\pm1.4}}$ & $84.9_{\textcolor{gray}{\pm2.6}}$ & $95.5_{\textcolor{gray}{\pm0.5}}$ & $57.8_{\textcolor{gray}{\pm1.4}}$ & $25.5_{\textcolor{gray}{\pm6.2}}$ & $68.8$ \\
GNorm & $66.4_{\textcolor{gray}{\pm2.7}}$ & $85.2_{\textcolor{gray}{\pm2.1}}$ & $93.2_{\textcolor{gray}{\pm0.6}}$ & $13.8_{\textcolor{gray}{\pm0.6}}$ & $4.8_{\textcolor{gray}{\pm0.3}}$ & $52.7$ \\
ReAct & $73.5_{\textcolor{gray}{\pm23.4}}$ & $88.0_{\textcolor{gray}{\pm10.7}}$ & $91.6_{\textcolor{gray}{\pm6.7}}$ & $23.3_{\textcolor{gray}{\pm0.8}}$ & $4.3_{\textcolor{gray}{\pm2.0}}$ & $56.1$ \\
VIM & $\textbf{0.0}_{\textcolor{gray}{\pm0.0}}$ & $\textbf{0.0}_{\textcolor{gray}{\pm0.0}}$ & $1.9_{\textcolor{gray}{\pm1.6}}$ & $\textbf{0.0}_{\textcolor{gray}{\pm0.0}}$ & $\textbf{0.0}_{\textcolor{gray}{\pm0.0}}$ & $0.4$ \\
MDS & $\textbf{0.0}_{\textcolor{gray}{\pm0.0}}$ & $\textbf{0.0}_{\textcolor{gray}{\pm0.0}}$ & $\textbf{0.0}_{\textcolor{gray}{\pm0.0}}$ & $0.7_{\textcolor{gray}{\pm0.0}}$ & $11.9_{\textcolor{gray}{\pm0.4}}$ & $2.5$ \\
RMDS & $90.5_{\textcolor{gray}{\pm0.3}}$ & $89.2_{\textcolor{gray}{\pm1.7}}$ & $95.4_{\textcolor{gray}{\pm0.4}}$ & $51.3_{\textcolor{gray}{\pm1.6}}$ & $17.7_{\textcolor{gray}{\pm0.4}}$ & $68.8$ \\
KNN & $\textbf{0.0}_{\textcolor{gray}{\pm0.0}}$ & $\textbf{0.0}_{\textcolor{gray}{\pm0.0}}$ & $\textbf{0.0}_{\textcolor{gray}{\pm0.0}}$ & $\textbf{0.0}_{\textcolor{gray}{\pm0.0}}$ & $0.6_{\textcolor{gray}{\pm0.0}}$ & $0.1$ \\
SHE & $15.9_{\textcolor{gray}{\pm0.6}}$ & $7.0_{\textcolor{gray}{\pm0.8}}$ & $4.9_{\textcolor{gray}{\pm0.4}}$ & $29.8_{\textcolor{gray}{\pm1.0}}$ & $7.7_{\textcolor{gray}{\pm0.8}}$ & $13.1$ \\
\midrule
\textbf{SPROD} & $\textbf{0.0}_{\textcolor{gray}{\pm0.0}}$ & $\textbf{0.0}_{\textcolor{gray}{\pm0.0}}$ & $\textbf{0.0}_{\textcolor{gray}{\pm0.0}}$ & $\textbf{0.0}_{\textcolor{gray}{\pm0.0}}$ & $\textbf{0.0}_{\textcolor{gray}{\pm0.0}}$ & $\textbf{0.0}$ \\
\bottomrule
\end{tabular}
}
    \end{minipage}
    \end{table}

%% file: far_ood_tables/ResNet-50_AUPR.tex
\begin{table}[t]
    \centering
    \caption{Average NSP-OOD detection performance (AUPR-In and AUPR-Out) using a ResNet-50 backbone. Cell values are performance metrics averaged over SVHN, LSUN, and iSUN as OOD.}
    \label{tab:nspood_aupr}
    \begin{minipage}{0.49\textwidth}
      \centering
      {\scriptsize\textbf{AUPR-IN}$\uparrow$} \\
      \resizebox{\linewidth}{!}{
\begin{tabular}{lcccccc}
\toprule
Method & WB & CA & UC & AMC & SpI & Avg. \\
\midrule
MSP & $95.8_{\textcolor{gray}{\pm0.3}}$ & $67.7_{\textcolor{gray}{\pm1.4}}$ & $43.4_{\textcolor{gray}{\pm3.9}}$ & $96.0_{\textcolor{gray}{\pm0.1}}$ & $95.6_{\textcolor{gray}{\pm0.2}}$ & $79.7$ \\
Energy & $95.5_{\textcolor{gray}{\pm1.8}}$ & $66.2_{\textcolor{gray}{\pm5.0}}$ & $44.2_{\textcolor{gray}{\pm6.2}}$ & $98.3_{\textcolor{gray}{\pm0.1}}$ & $97.9_{\textcolor{gray}{\pm0.2}}$ & $80.4$ \\
MLS & $95.7_{\textcolor{gray}{\pm1.3}}$ & $66.5_{\textcolor{gray}{\pm4.0}}$ & $43.9_{\textcolor{gray}{\pm5.6}}$ & $98.0_{\textcolor{gray}{\pm0.1}}$ & $96.8_{\textcolor{gray}{\pm0.2}}$ & $80.2$ \\
KLM & $77.1_{\textcolor{gray}{\pm1.1}}$ & $27.8_{\textcolor{gray}{\pm1.0}}$ & $34.4_{\textcolor{gray}{\pm1.3}}$ & $93.9_{\textcolor{gray}{\pm0.2}}$ & $96.6_{\textcolor{gray}{\pm0.4}}$ & $66.0$ \\
GNorm & $97.2_{\textcolor{gray}{\pm0.2}}$ & $49.7_{\textcolor{gray}{\pm1.1}}$ & $59.6_{\textcolor{gray}{\pm2.0}}$ & $98.4_{\textcolor{gray}{\pm0.1}}$ & $99.0_{\textcolor{gray}{\pm0.0}}$ & $80.8$ \\
ReAct & $96.1_{\textcolor{gray}{\pm1.5}}$ & $53.9_{\textcolor{gray}{\pm4.6}}$ & $40.6_{\textcolor{gray}{\pm5.9}}$ & $98.0_{\textcolor{gray}{\pm0.0}}$ & $99.0_{\textcolor{gray}{\pm0.3}}$ & $77.5$ \\
VIM & $\textbf{100.0}_{\textcolor{gray}{\pm0.0}}$ & $\textbf{100.0}_{\textcolor{gray}{\pm0.0}}$ & $99.2_{\textcolor{gray}{\pm0.7}}$ & $\textbf{100.0}_{\textcolor{gray}{\pm0.0}}$ & $99.7_{\textcolor{gray}{\pm0.0}}$ & $99.8$ \\
MDS & $\textbf{100.0}_{\textcolor{gray}{\pm0.0}}$ & $\textbf{100.0}_{\textcolor{gray}{\pm0.0}}$ & $\textbf{100.0}_{\textcolor{gray}{\pm0.0}}$ & $99.7_{\textcolor{gray}{\pm0.0}}$ & $98.0_{\textcolor{gray}{\pm0.0}}$ & $99.5$ \\
RMDS & $78.0_{\textcolor{gray}{\pm0.2}}$ & $46.1_{\textcolor{gray}{\pm3.9}}$ & $35.8_{\textcolor{gray}{\pm0.5}}$ & $96.4_{\textcolor{gray}{\pm0.1}}$ & $97.4_{\textcolor{gray}{\pm0.1}}$ & $70.7$ \\
KNN & $\textbf{100.0}_{\textcolor{gray}{\pm0.0}}$ & $\textbf{100.0}_{\textcolor{gray}{\pm0.0}}$ & $\textbf{100.0}_{\textcolor{gray}{\pm0.0}}$ & $\textbf{100.0}_{\textcolor{gray}{\pm0.0}}$ & $99.8_{\textcolor{gray}{\pm0.0}}$ & $\textbf{100.0}$ \\
SHE & $98.7_{\textcolor{gray}{\pm0.1}}$ & $96.6_{\textcolor{gray}{\pm0.4}}$ & $98.6_{\textcolor{gray}{\pm0.1}}$ & $89.6_{\textcolor{gray}{\pm0.5}}$ & $98.6_{\textcolor{gray}{\pm0.1}}$ & $96.4$ \\
\midrule
\textbf{SPROD} & $\textbf{100.0}_{\textcolor{gray}{\pm0.0}}$ & $\textbf{100.0}_{\textcolor{gray}{\pm0.0}}$ & $\textbf{100.0}_{\textcolor{gray}{\pm0.0}}$ & $\textbf{100.0}_{\textcolor{gray}{\pm0.0}}$ & $\textbf{99.9}_{\textcolor{gray}{\pm0.0}}$ & $\textbf{100.0}$ \\
\bottomrule
\end{tabular}
}
    \end{minipage}
    \hfill
    \begin{minipage}{0.49\textwidth}
      \centering
      {\scriptsize\textbf{AUPR-OUT}$\uparrow$} \\
      \resizebox{\linewidth}{!}{
\begin{tabular}{lcccccc}
\toprule
Method & WB & CA & UC & AMC & SpI & Avg. \\
\midrule
MSP & $55.8_{\textcolor{gray}{\pm1.9}}$ & $85.9_{\textcolor{gray}{\pm0.7}}$ & $65.0_{\textcolor{gray}{\pm2.2}}$ & $85.4_{\textcolor{gray}{\pm0.6}}$ & $93.9_{\textcolor{gray}{\pm0.5}}$ & $77.2$ \\
Energy & $56.4_{\textcolor{gray}{\pm18.0}}$ & $83.6_{\textcolor{gray}{\pm6.8}}$ & $66.3_{\textcolor{gray}{\pm7.9}}$ & $92.6_{\textcolor{gray}{\pm0.4}}$ & $97.5_{\textcolor{gray}{\pm0.3}}$ & $79.3$ \\
MLS & $56.3_{\textcolor{gray}{\pm12.1}}$ & $84.7_{\textcolor{gray}{\pm4.4}}$ & $65.7_{\textcolor{gray}{\pm5.5}}$ & $91.3_{\textcolor{gray}{\pm0.4}}$ & $95.7_{\textcolor{gray}{\pm0.4}}$ & $78.7$ \\
KLM & $38.4_{\textcolor{gray}{\pm1.7}}$ & $76.9_{\textcolor{gray}{\pm0.9}}$ & $62.7_{\textcolor{gray}{\pm0.6}}$ & $79.6_{\textcolor{gray}{\pm0.6}}$ & $94.9_{\textcolor{gray}{\pm0.6}}$ & $70.5$ \\
GNorm & $65.1_{\textcolor{gray}{\pm1.7}}$ & $83.8_{\textcolor{gray}{\pm0.7}}$ & $71.1_{\textcolor{gray}{\pm1.5}}$ & $94.2_{\textcolor{gray}{\pm0.2}}$ & $98.9_{\textcolor{gray}{\pm0.0}}$ & $82.6$ \\
ReAct & $57.9_{\textcolor{gray}{\pm15.9}}$ & $78.6_{\textcolor{gray}{\pm7.4}}$ & $65.0_{\textcolor{gray}{\pm8.0}}$ & $91.8_{\textcolor{gray}{\pm0.2}}$ & $99.0_{\textcolor{gray}{\pm0.3}}$ & $78.5$ \\
VIM & $\textbf{100.0}_{\textcolor{gray}{\pm0.0}}$ & $\textbf{100.0}_{\textcolor{gray}{\pm0.0}}$ & $99.8_{\textcolor{gray}{\pm0.2}}$ & $99.9_{\textcolor{gray}{\pm0.0}}$ & $99.2_{\textcolor{gray}{\pm0.0}}$ & $99.8$ \\
MDS & $\textbf{100.0}_{\textcolor{gray}{\pm0.0}}$ & $\textbf{100.0}_{\textcolor{gray}{\pm0.0}}$ & $\textbf{100.0}_{\textcolor{gray}{\pm0.0}}$ & $98.8_{\textcolor{gray}{\pm0.0}}$ & $96.5_{\textcolor{gray}{\pm0.0}}$ & $99.1$ \\
RMDS & $31.3_{\textcolor{gray}{\pm0.3}}$ & $81.8_{\textcolor{gray}{\pm1.8}}$ & $64.0_{\textcolor{gray}{\pm0.6}}$ & $83.5_{\textcolor{gray}{\pm0.5}}$ & $96.3_{\textcolor{gray}{\pm0.1}}$ & $71.4$ \\
KNN & $\textbf{100.0}_{\textcolor{gray}{\pm0.0}}$ & $\textbf{100.0}_{\textcolor{gray}{\pm0.0}}$ & $\textbf{100.0}_{\textcolor{gray}{\pm0.0}}$ & $\textbf{100.0}_{\textcolor{gray}{\pm0.0}}$ & $99.8_{\textcolor{gray}{\pm0.0}}$ & $\textbf{100.0}$ \\
SHE & $91.6_{\textcolor{gray}{\pm0.4}}$ & $99.4_{\textcolor{gray}{\pm0.1}}$ & $99.5_{\textcolor{gray}{\pm0.0}}$ & $97.0_{\textcolor{gray}{\pm0.1}}$ & $98.3_{\textcolor{gray}{\pm0.1}}$ & $97.2$ \\
\midrule
\textbf{SPROD} & $\textbf{100.0}_{\textcolor{gray}{\pm0.0}}$ & $\textbf{100.0}_{\textcolor{gray}{\pm0.0}}$ & $\textbf{100.0}_{\textcolor{gray}{\pm0.0}}$ & $99.9_{\textcolor{gray}{\pm0.0}}$ & $\textbf{100.0}_{\textcolor{gray}{\pm0.0}}$ & $\textbf{100.0}$ \\
\bottomrule
\end{tabular}
}
    \end{minipage}
    \end{table}

%% file: backbone_tables/AUROC_all.tex
\begin{table}[t]
\small
\centering
\caption{AUROC performance of all evaluated methods across various backbone architectures. Results are averaged over five SP-OOD datasets.}
\label{tab:auroc_all}
\begin{tabular}{lccccccccccc}
\toprule
Method & R18 & R34 & R50 & R101 & DINOv2 & ViT & Swin & DeiT & CvNXt & BiT & Avg. \\
\midrule
MSP & 61.1 & 61.9 & 61.9 & 63.3 & 70.2 & 65.4 & 70.8 & 69.7 & 72.0 & 65.0 & 66.1 \\
Energy & 61.5 & 61.6 & 61.3 & 62.7 & 70.5 & 65.8 & 70.4 & 70.0 & 71.5 & 64.7 & 66.0 \\
MLS & 61.7 & 61.9 & 61.6 & 63.1 & 70.5 & 65.8 & 70.7 & 70.1 & 71.3 & 64.8 & 66.2 \\
KLM & 56.6 & 58.8 & 60.7 & 61.5 & 70.2 & 64.5 & 68.6 & 62.4 & 69.3 & 60.6 & 63.3 \\
GNorm & 64.3 & 64.2 & 64.7 & 66.4 & 69.9 & 66.4 & 56.7 & 62.1 & 67.8 & 54.2 & 63.7 \\
ReAct & 62.4 & 62.1 & 64.7 & 64.1 & 70.0 & 67.3 & 72.1 & 69.8 & 70.7 & 60.5 & 66.4 \\
VIM & 66.1 & 68.0 & 69.3 & 72.1 & 78.2 & 73.5 & 77.6 & 72.9 & 80.0 & 77.8 & 73.5 \\
MDS & 65.6 & 70.5 & 72.2 & 74.2 & 83.1 & 82.9 & 80.0 & 71.3 & 84.8 & 84.2 & 76.9 \\
RMDS & 54.8 & 57.2 & 60.2 & 58.3 & 63.3 & 61.1 & 69.6 & 65.5 & 69.5 & 66.5 & 62.6 \\
KNN & 77.7 & 78.6 & 80.3 & 79.9 & 79.7 & 80.3 & 84.3 & 78.9 & 86.1 & 81.6 & 80.7 \\
SHE & 69.4 & 72.4 & 68.4 & 73.5 & 83.0 & 79.8 & 84.9 & 81.8 & 84.6 & 73.5 & 77.1 \\
\bottomrule
\textbf{SPROD} & \textbf{83.0} & \textbf{83.2} & \textbf{85.1} & \textbf{85.9} & \textbf{87.2} & \textbf{85.1} & \textbf{90.1} & \textbf{84.6} & \textbf{89.8} & \textbf{87.1} & \textbf{86.1} \\
\bottomrule
\end{tabular}
\end{table}

%% file: backbone_tables/FPR_all.tex
\begin{table}[t]
\small
\centering
\caption{FPR@95 performance of all evaluated methods across various backbone architectures. Results are averaged over five SP-OOD datasets.}
\label{tab:fpr_all}
\begin{tabular}{lccccccccccc}
\toprule
Method & R18 & R34 & R50 & R101 & DINOv2 & ViT & Swin & DeiT & CvNXt & BiT & Avg. \\
\midrule
MSP & 90.8 & 88.9 & 88.4 & 86.5 & 81.1 & 86.4 & 71.7 & 80.4 & 67.8 & 84.2 & 82.6 \\
Energy & 88.5 & 86.9 & 88.9 & 87.5 & 77.9 & 84.9 & 73.7 & 78.6 & 68.2 & 82.8 & 81.8 \\
MLS & 89.6 & 87.4 & 88.5 & 86.8 & 78.3 & 84.9 & 71.5 & 78.2 & 67.8 & 82.7 & 81.6 \\
KLM & 91.3 & 90.3 & 88.3 & 87.4 & 80.6 & 85.7 & 72.6 & 80.5 & 66.8 & 83.5 & 82.7 \\
GNorm & 86.1 & 84.0 & 83.8 & 82.9 & 80.2 & 85.3 & 79.2 & 82.3 & 70.7 & 87.7 & 82.2 \\
ReAct & 85.6 & 84.4 & 84.0 & 83.5 & 78.3 & 86.8 & 72.6 & 78.3 & 69.4 & 86.1 & 80.9 \\
VIM & 83.0 & 81.9 & 78.5 & 74.3 & 56.7 & 75.0 & 68.3 & 82.6 & 47.4 & 61.8 & 71.0 \\
MDS & 82.0 & 80.8 & 73.5 & 67.7 & 51.4 & 61.9 & 70.7 & 82.6 & 50.4 & 48.4 & 66.9 \\
RMDS & 93.2 & 92.2 & 91.6 & 89.4 & 81.0 & 83.8 & 72.9 & 85.2 & 67.7 & 82.2 & 83.9 \\
KNN & 64.7 & 62.4 & 58.4 & 57.3 & 51.8 & 66.7 & 53.4 & 71.9 & 46.9 & 51.0 & 58.5 \\
SHE & 68.2 & 65.7 & 70.5 & 67.0 & 54.2 & 63.1 & 46.4 & 66.9 & 50.5 & 74.0 & 62.6 \\
\bottomrule
\textbf{SPROD} & \textbf{53.1} & \textbf{54.6} & \textbf{49.0} & \textbf{46.6} & \textbf{43.0} & \textbf{51.1} & \textbf{35.3} & \textbf{55.1} & \textbf{35.2} & \textbf{42.9} & \textbf{46.6} \\
\bottomrule
\end{tabular}
\end{table}

%% file: backbone_aupr_tables/AUPR_IN_all.tex
\begin{table}[t]
\small
\centering
\caption{AUPR-IN performance of all evaluated methods across various backbone architectures. Results are averaged over five SP-OOD datasets.}
\label{tab:auprin_all}
\begin{tabular}{lccccccccccc}
\toprule
Method & R18 & R34 & R50 & R101 & DINOv2 & ViT & Swin & DeiT & CvNXt & BiT & Avg. \\
\midrule
MSP & 48.4 & 49.3 & 49.8 & 50.9 & 59.3 & 54.4 & 61.1 & 59.0 & 64.1 & 51.3 & 54.8 \\
Energy & 47.9 & 48.3 & 48.7 & 50.1 & 59.9 & 54.8 & 59.3 & 58.7 & 63.1 & 50.4 & 54.1 \\
MLS & 48.4 & 48.6 & 49.3 & 50.6 & 59.2 & 54.8 & 60.1 & 59.0 & 63.0 & 50.6 & 54.4 \\
KLM & 41.0 & 43.8 & 45.2 & 45.2 & 56.1 & 49.3 & 52.4 & 48.4 & 52.8 & 45.5 & 48.0 \\
GNorm & 50.3 & 50.2 & 51.0 & 51.3 & 58.5 & 55.5 & 38.8 & 42.9 & 51.5 & 40.5 & 49.0 \\
ReAct & 49.9 & 50.7 & 52.7 & 53.1 & 59.6 & 56.1 & 63.0 & 59.3 & 62.6 & 47.8 & 55.5 \\
VIM & 51.6 & 53.9 & 55.4 & 59.7 & 66.7 & 62.1 & 68.9 & 63.1 & 72.5 & 64.4 & 61.8 \\
MDS & 52.2 & 60.2 & 61.6 & 64.1 & 71.8 & 74.5 & 72.0 & 60.8 & 77.0 & 75.3 & 67.0 \\
RMDS & 42.2 & 45.3 & 45.1 & 45.5 & 50.2 & 49.5 & 59.0 & 53.7 & 59.2 & 48.8 & 49.9 \\
KNN & 63.6 & 66.0 & 68.0 & 67.7 & 64.6 & 70.8 & 74.6 & 69.2 & 76.5 & 69.8 & 69.1 \\
SHE & 52.8 & 57.2 & 52.5 & 56.6 & 75.1 & 71.5 & 75.0 & 73.5 & 80.1 & 61.2 & 65.5 \\
\bottomrule
\textbf{SPROD} & \textbf{72.7} & \textbf{74.0} & \textbf{76.5} & \textbf{77.1} & \textbf{78.9} & \textbf{77.1} & \textbf{84.1} & \textbf{76.5} & \textbf{84.3} & \textbf{78.3} & \textbf{78.0} \\
\bottomrule
\end{tabular}
\end{table}

%% file: backbone_aupr_tables/AUPR_OUT_all.tex
\begin{table}[t]
\small
\centering
\caption{AUPR-OUT performance of all evaluated methods across various backbone architectures. Results are averaged over five SP-OOD datasets.}
\label{tab:auprout_all}
\begin{tabular}{lccccccccccc}
\toprule
Method & R18 & R34 & R50 & R101 & DINOv2 & ViT & Swin & DeiT & CvNXt & BiT & Avg. \\
\midrule
MSP & 72.2 & 73.0 & 73.0 & 74.3 & 79.9 & 75.3 & 81.3 & 78.9 & 81.9 & 76.3 & 76.6 \\
Energy & 73.1 & 73.6 & 73.0 & 74.0 & 80.4 & 76.1 & 81.5 & 80.0 & 81.9 & 76.6 & 77.0 \\
MLS & 72.9 & 73.5 & 73.1 & 74.2 & 80.4 & 76.0 & 81.6 & 79.8 & 81.8 & 76.6 & 77.0 \\
KLM & 70.8 & 71.5 & 73.0 & 73.8 & 79.3 & 74.5 & 79.6 & 75.4 & 80.9 & 74.0 & 75.3 \\
GNorm & 74.3 & 74.8 & 75.0 & 76.1 & 79.5 & 76.0 & 76.1 & 76.6 & 79.2 & 71.8 & 75.9 \\
ReAct & 73.8 & 74.1 & 75.2 & 75.4 & 80.2 & 76.7 & 82.1 & 80.0 & 81.4 & 73.5 & 77.2 \\
VIM & 77.8 & 79.0 & 80.1 & 81.8 & 86.6 & 82.5 & 84.8 & 80.2 & 89.2 & 86.1 & 82.8 \\
MDS & 77.8 & 79.7 & 81.3 & 82.6 & 88.7 & 87.9 & 83.9 & 78.3 & 90.3 & 88.5 & 83.9 \\
RMDS & 68.9 & 70.4 & 71.8 & 72.4 & 76.4 & 74.0 & 79.8 & 75.7 & 80.7 & 74.9 & 74.5 \\
KNN & 85.6 & 85.9 & 87.4 & 87.7 & 89.4 & 86.9 & 90.7 & 85.8 & 91.1 & 89.5 & 88.0 \\
SHE & 80.1 & 82.6 & 79.1 & 82.6 & 88.3 & 86.4 & 90.7 & 84.6 & 87.6 & 82.4 & 84.4 \\
\bottomrule
\textbf{SPROD} & \textbf{88.8} & \textbf{88.7} & \textbf{90.1} & \textbf{90.7} & \textbf{92.9} & \textbf{90.3} & \textbf{93.9} & \textbf{90.7} & \textbf{93.8} & \textbf{92.7} & \textbf{91.3} \\
\bottomrule
\end{tabular}
\end{table}

%% file: backbone_tables/ResNet-18.tex
\begin{table}[t]
    \centering
    \caption{AUROC and FPR@95 performance of all methods using ResNet-18 as the feature backbone.}
    \label{tab:backbone_ResNet-18_aurocfpr}
    \begin{minipage}{0.49\textwidth}
      \centering
      {\scriptsize\textbf{AUROC}$\uparrow$} \\
      \resizebox{\linewidth}{!}{
\begin{tabular}{lcccccc}
\toprule
Method & WB & CA & UC & AMC & SpI & Avg. \\
\midrule
MSP & $68.1_{\textcolor{gray}{\pm0.7}}$ & $47.6_{\textcolor{gray}{\pm2.9}}$ & $40.2_{\textcolor{gray}{\pm0.5}}$ & $72.4_{\textcolor{gray}{\pm0.6}}$ & $77.3_{\textcolor{gray}{\pm0.2}}$ & $61.1$ \\
Energy & $69.4_{\textcolor{gray}{\pm4.3}}$ & $48.5_{\textcolor{gray}{\pm4.2}}$ & $43.2_{\textcolor{gray}{\pm5.0}}$ & $71.0_{\textcolor{gray}{\pm0.8}}$ & $75.6_{\textcolor{gray}{\pm0.3}}$ & $61.5$ \\
MLS & $69.2_{\textcolor{gray}{\pm3.6}}$ & $48.1_{\textcolor{gray}{\pm4.0}}$ & $42.7_{\textcolor{gray}{\pm3.8}}$ & $71.6_{\textcolor{gray}{\pm0.8}}$ & $76.8_{\textcolor{gray}{\pm0.3}}$ & $61.7$ \\
KLM & $43.9_{\textcolor{gray}{\pm0.5}}$ & $41.1_{\textcolor{gray}{\pm1.4}}$ & $57.7_{\textcolor{gray}{\pm0.4}}$ & $68.2_{\textcolor{gray}{\pm0.5}}$ & $72.3_{\textcolor{gray}{\pm0.4}}$ & $56.6$ \\
GNorm & $81.9_{\textcolor{gray}{\pm0.3}}$ & $42.0_{\textcolor{gray}{\pm2.4}}$ & $48.3_{\textcolor{gray}{\pm0.4}}$ & $67.5_{\textcolor{gray}{\pm0.6}}$ & $82.0_{\textcolor{gray}{\pm0.2}}$ & $64.3$ \\
ReAct & $75.9_{\textcolor{gray}{\pm4.1}}$ & $40.8_{\textcolor{gray}{\pm3.7}}$ & $44.0_{\textcolor{gray}{\pm5.5}}$ & $71.3_{\textcolor{gray}{\pm0.8}}$ & $79.8_{\textcolor{gray}{\pm0.4}}$ & $62.4$ \\
VIM & $77.2_{\textcolor{gray}{\pm3.0}}$ & $52.7_{\textcolor{gray}{\pm3.8}}$ & $63.9_{\textcolor{gray}{\pm3.8}}$ & $71.8_{\textcolor{gray}{\pm0.8}}$ & $64.8_{\textcolor{gray}{\pm0.2}}$ & $66.1$ \\
\bottomrule
\end{tabular}
}
    \end{minipage}
    \hfill
    \begin{minipage}{0.49\textwidth}
      \centering
      {\scriptsize\textbf{FPR@95}$\downarrow$} \\
      \resizebox{\linewidth}{!}{
\begin{tabular}{lcccccc}
\toprule
Method & WB & CA & UC & AMC & SpI & Avg. \\
\midrule
MSP & $89.7_{\textcolor{gray}{\pm0.6}}$ & $98.0_{\textcolor{gray}{\pm0.5}}$ & $95.5_{\textcolor{gray}{\pm0.6}}$ & $89.2_{\textcolor{gray}{\pm0.4}}$ & $81.7_{\textcolor{gray}{\pm0.8}}$ & $90.8$ \\
Energy & $85.7_{\textcolor{gray}{\pm8.0}}$ & $96.9_{\textcolor{gray}{\pm1.0}}$ & $91.3_{\textcolor{gray}{\pm5.1}}$ & $90.8_{\textcolor{gray}{\pm0.4}}$ & $77.9_{\textcolor{gray}{\pm1.3}}$ & $88.5$ \\
MLS & $86.7_{\textcolor{gray}{\pm6.2}}$ & $97.8_{\textcolor{gray}{\pm0.6}}$ & $93.0_{\textcolor{gray}{\pm3.6}}$ & $90.3_{\textcolor{gray}{\pm0.4}}$ & $80.3_{\textcolor{gray}{\pm0.6}}$ & $89.6$ \\
KLM & $91.2_{\textcolor{gray}{\pm0.5}}$ & $98.0_{\textcolor{gray}{\pm0.5}}$ & $95.5_{\textcolor{gray}{\pm0.6}}$ & $85.8_{\textcolor{gray}{\pm0.4}}$ & $86.0_{\textcolor{gray}{\pm1.6}}$ & $91.3$ \\
GNorm & $85.9_{\textcolor{gray}{\pm0.8}}$ & $98.1_{\textcolor{gray}{\pm0.4}}$ & $95.1_{\textcolor{gray}{\pm0.5}}$ & $91.4_{\textcolor{gray}{\pm0.4}}$ & $60.1_{\textcolor{gray}{\pm1.0}}$ & $86.1$ \\
ReAct & $82.0_{\textcolor{gray}{\pm9.1}}$ & $98.0_{\textcolor{gray}{\pm0.8}}$ & $92.1_{\textcolor{gray}{\pm4.5}}$ & $90.9_{\textcolor{gray}{\pm0.5}}$ & $64.8_{\textcolor{gray}{\pm1.5}}$ & $85.6$ \\
VIM & $69.0_{\textcolor{gray}{\pm5.6}}$ & $95.1_{\textcolor{gray}{\pm1.0}}$ & $71.4_{\textcolor{gray}{\pm5.7}}$ & $89.8_{\textcolor{gray}{\pm0.5}}$ & $89.8_{\textcolor{gray}{\pm0.5}}$ & $83.0$ \\
\bottomrule
\end{tabular}
}
    \end{minipage}
    \end{table}

%% file: backbone_tables/ResNet-34.tex
\begin{table}[t]
    \centering
    \caption{AUROC and FPR@95 performance of all methods using ResNet-34 as the feature backbone.}
    \label{tab:backbone_ResNet-34_aurocfpr}
    \begin{minipage}{0.49\textwidth}
      \centering
      {\scriptsize\textbf{AUROC}$\uparrow$} \\
      \resizebox{\linewidth}{!}{
\begin{tabular}{lcccccc}
\toprule
Method & WB & CA & UC & AMC & SpI & Avg. \\
\midrule
MSP & $66.0_{\textcolor{gray}{\pm0.7}}$ & $41.9_{\textcolor{gray}{\pm1.5}}$ & $42.7_{\textcolor{gray}{\pm0.6}}$ & $77.5_{\textcolor{gray}{\pm0.4}}$ & $81.6_{\textcolor{gray}{\pm0.2}}$ & $61.9$ \\
Energy & $64.9_{\textcolor{gray}{\pm6.8}}$ & $43.2_{\textcolor{gray}{\pm4.8}}$ & $42.5_{\textcolor{gray}{\pm4.4}}$ & $78.2_{\textcolor{gray}{\pm0.6}}$ & $79.4_{\textcolor{gray}{\pm0.2}}$ & $61.6$ \\
MLS & $65.0_{\textcolor{gray}{\pm5.7}}$ & $42.6_{\textcolor{gray}{\pm4.1}}$ & $42.5_{\textcolor{gray}{\pm3.4}}$ & $78.4_{\textcolor{gray}{\pm0.6}}$ & $80.8_{\textcolor{gray}{\pm0.2}}$ & $61.9$ \\
KLM & $42.2_{\textcolor{gray}{\pm0.8}}$ & $45.0_{\textcolor{gray}{\pm1.3}}$ & $56.2_{\textcolor{gray}{\pm0.5}}$ & $72.2_{\textcolor{gray}{\pm0.5}}$ & $78.2_{\textcolor{gray}{\pm0.8}}$ & $58.8$ \\
GNorm & $77.9_{\textcolor{gray}{\pm0.4}}$ & $35.4_{\textcolor{gray}{\pm0.9}}$ & $50.0_{\textcolor{gray}{\pm0.5}}$ & $73.7_{\textcolor{gray}{\pm0.4}}$ & $\textbf{84.1}_{\textcolor{gray}{\pm0.1}}$ & $64.2$ \\
ReAct & $72.3_{\textcolor{gray}{\pm5.9}}$ & $34.4_{\textcolor{gray}{\pm5.9}}$ & $43.5_{\textcolor{gray}{\pm5.4}}$ & $77.9_{\textcolor{gray}{\pm0.6}}$ & $82.4_{\textcolor{gray}{\pm0.4}}$ & $62.1$ \\
VIM & $80.8_{\textcolor{gray}{\pm4.3}}$ & $50.3_{\textcolor{gray}{\pm3.2}}$ & $62.4_{\textcolor{gray}{\pm3.3}}$ & $75.5_{\textcolor{gray}{\pm0.7}}$ & $70.9_{\textcolor{gray}{\pm0.2}}$ & $68.0$ \\
\bottomrule
\end{tabular}
}
    \end{minipage}
    \hfill
    \begin{minipage}{0.49\textwidth}
      \centering
      {\scriptsize\textbf{FPR@95}$\downarrow$} \\
      \resizebox{\linewidth}{!}{
\begin{tabular}{lcccccc}
\toprule
Method & WB & CA & UC & AMC & SpI & Avg. \\
\midrule
MSP & $91.2_{\textcolor{gray}{\pm0.4}}$ & $98.4_{\textcolor{gray}{\pm0.4}}$ & $95.5_{\textcolor{gray}{\pm0.7}}$ & $85.9_{\textcolor{gray}{\pm0.5}}$ & $73.7_{\textcolor{gray}{\pm2.5}}$ & $88.9$ \\
Energy & $87.1_{\textcolor{gray}{\pm9.2}}$ & $97.3_{\textcolor{gray}{\pm1.3}}$ & $93.1_{\textcolor{gray}{\pm4.8}}$ & $85.3_{\textcolor{gray}{\pm0.7}}$ & $71.6_{\textcolor{gray}{\pm2.0}}$ & $86.9$ \\
MLS & $88.6_{\textcolor{gray}{\pm7.0}}$ & $98.0_{\textcolor{gray}{\pm0.8}}$ & $94.1_{\textcolor{gray}{\pm3.7}}$ & $84.7_{\textcolor{gray}{\pm0.7}}$ & $71.8_{\textcolor{gray}{\pm1.3}}$ & $87.4$ \\
KLM & $92.9_{\textcolor{gray}{\pm0.5}}$ & $98.4_{\textcolor{gray}{\pm0.5}}$ & $95.5_{\textcolor{gray}{\pm0.8}}$ & $84.1_{\textcolor{gray}{\pm0.6}}$ & $80.6_{\textcolor{gray}{\pm2.7}}$ & $90.3$ \\
GNorm & $89.2_{\textcolor{gray}{\pm0.5}}$ & $98.5_{\textcolor{gray}{\pm0.4}}$ & $95.3_{\textcolor{gray}{\pm0.5}}$ & $85.0_{\textcolor{gray}{\pm0.5}}$ & $\textbf{52.1}_{\textcolor{gray}{\pm0.9}}$ & $84.0$ \\
ReAct & $86.2_{\textcolor{gray}{\pm9.8}}$ & $96.3_{\textcolor{gray}{\pm2.3}}$ & $94.5_{\textcolor{gray}{\pm4.0}}$ & $85.5_{\textcolor{gray}{\pm0.7}}$ & $59.6_{\textcolor{gray}{\pm1.7}}$ & $84.4$ \\
VIM & $64.6_{\textcolor{gray}{\pm8.5}}$ & $95.4_{\textcolor{gray}{\pm0.9}}$ & $72.2_{\textcolor{gray}{\pm4.7}}$ & $89.7_{\textcolor{gray}{\pm0.5}}$ & $87.6_{\textcolor{gray}{\pm0.1}}$ & $81.9$ \\
\bottomrule
\end{tabular}
}
    \end{minipage}
    \end{table}

%% file: backbone_tables/ResNet-50.tex
    \begin{table}[t]
    \centering
    \caption{AUROC and FPR@95 performance of all methods using ResNet-50 as the feature backbone.}
    \label{tab:backbone_ResNet-50_aurocfpr}
    \begin{minipage}{0.49\textwidth}
      \centering
      {\scriptsize\textbf{AUROC}$\uparrow$} \\
      \resizebox{\linewidth}{!}{
\begin{tabular}{lcccccc}
\toprule
Method & WB & CA & UC & AMC & SpI & Avg. \\
\midrule
MSP & $62.3_{\textcolor{gray}{\pm0.6}}$ & $46.0_{\textcolor{gray}{\pm1.4}}$ & $38.5_{\textcolor{gray}{\pm0.3}}$ & $79.7_{\textcolor{gray}{\pm0.4}}$ & $83.1_{\textcolor{gray}{\pm0.3}}$ & $61.9$ \\
Energy & $62.0_{\textcolor{gray}{\pm2.6}}$ & $45.4_{\textcolor{gray}{\pm3.4}}$ & $38.4_{\textcolor{gray}{\pm2.1}}$ & $79.9_{\textcolor{gray}{\pm0.6}}$ & $80.6_{\textcolor{gray}{\pm0.4}}$ & $61.3$ \\
MLS & $62.2_{\textcolor{gray}{\pm2.3}}$ & $45.3_{\textcolor{gray}{\pm3.2}}$ & $38.4_{\textcolor{gray}{\pm1.4}}$ & $80.2_{\textcolor{gray}{\pm0.6}}$ & $81.9_{\textcolor{gray}{\pm0.3}}$ & $61.6$ \\
KLM & $51.2_{\textcolor{gray}{\pm0.7}}$ & $41.7_{\textcolor{gray}{\pm2.5}}$ & $57.0_{\textcolor{gray}{\pm0.2}}$ & $74.2_{\textcolor{gray}{\pm0.6}}$ & $79.6_{\textcolor{gray}{\pm0.8}}$ & $60.7$ \\
GNorm & $79.5_{\textcolor{gray}{\pm0.4}}$ & $38.0_{\textcolor{gray}{\pm1.3}}$ & $46.6_{\textcolor{gray}{\pm0.4}}$ & $74.2_{\textcolor{gray}{\pm0.5}}$ & $85.2_{\textcolor{gray}{\pm0.2}}$ & $64.7$ \\
ReAct & $72.9_{\textcolor{gray}{\pm3.6}}$ & $45.6_{\textcolor{gray}{\pm5.3}}$ & $41.3_{\textcolor{gray}{\pm3.1}}$ & $80.1_{\textcolor{gray}{\pm0.6}}$ & $83.6_{\textcolor{gray}{\pm0.7}}$ & $64.7$ \\
VIM & $79.6_{\textcolor{gray}{\pm2.5}}$ & $50.4_{\textcolor{gray}{\pm3.1}}$ & $60.7_{\textcolor{gray}{\pm1.7}}$ & $78.6_{\textcolor{gray}{\pm0.6}}$ & $77.4_{\textcolor{gray}{\pm0.9}}$ & $69.3$ \\
\bottomrule
\end{tabular}
}
    \end{minipage}
    \hfill
    \begin{minipage}{0.49\textwidth}
      \centering
      {\scriptsize\textbf{FPR@95}$\downarrow$} \\
      \resizebox{\linewidth}{!}{
\begin{tabular}{lcccccc}
\toprule
Method & WB & CA & UC & AMC & SpI & Avg. \\
\midrule
MSP & $87.9_{\textcolor{gray}{\pm0.8}}$ & $98.7_{\textcolor{gray}{\pm0.5}}$ & $97.3_{\textcolor{gray}{\pm0.3}}$ & $83.8_{\textcolor{gray}{\pm0.7}}$ & $74.1_{\textcolor{gray}{\pm1.2}}$ & $88.4$ \\
Energy & $89.2_{\textcolor{gray}{\pm3.2}}$ & $98.6_{\textcolor{gray}{\pm0.7}}$ & $95.5_{\textcolor{gray}{\pm3.1}}$ & $84.8_{\textcolor{gray}{\pm0.8}}$ & $76.3_{\textcolor{gray}{\pm0.9}}$ & $88.9$ \\
MLS & $88.1_{\textcolor{gray}{\pm2.0}}$ & $98.8_{\textcolor{gray}{\pm0.6}}$ & $96.7_{\textcolor{gray}{\pm2.0}}$ & $84.4_{\textcolor{gray}{\pm0.8}}$ & $74.6_{\textcolor{gray}{\pm0.9}}$ & $88.5$ \\
KLM & $89.1_{\textcolor{gray}{\pm0.7}}$ & $98.7_{\textcolor{gray}{\pm0.5}}$ & $97.1_{\textcolor{gray}{\pm0.3}}$ & $80.5_{\textcolor{gray}{\pm0.8}}$ & $76.1_{\textcolor{gray}{\pm1.7}}$ & $88.3$ \\
GNorm & $84.2_{\textcolor{gray}{\pm0.7}}$ & $98.8_{\textcolor{gray}{\pm0.4}}$ & $97.1_{\textcolor{gray}{\pm0.1}}$ & $84.2_{\textcolor{gray}{\pm0.6}}$ & $54.7_{\textcolor{gray}{\pm0.6}}$ & $83.8$ \\
ReAct & $86.9_{\textcolor{gray}{\pm7.0}}$ & $96.3_{\textcolor{gray}{\pm2.4}}$ & $95.5_{\textcolor{gray}{\pm3.2}}$ & $83.9_{\textcolor{gray}{\pm0.8}}$ & $57.5_{\textcolor{gray}{\pm1.6}}$ & $84.0$ \\
VIM & $61.4_{\textcolor{gray}{\pm3.5}}$ & $96.2_{\textcolor{gray}{\pm0.4}}$ & $69.0_{\textcolor{gray}{\pm1.5}}$ & $86.6_{\textcolor{gray}{\pm0.7}}$ & $79.5_{\textcolor{gray}{\pm0.5}}$ & $78.5$ \\
\bottomrule
\end{tabular}
}
    \end{minipage}
    \end{table}
    

%% file: backbone_tables/ResNet-101.tex
\begin{table}[t]
    \centering
    \caption{AUROC and FPR@95 performance of all methods using ResNet-101 as the feature backbone.}
    \label{tab:backbone_ResNet-101_aurocfpr}
    \begin{minipage}{0.49\textwidth}
      \centering
      {\scriptsize\textbf{AUROC}$\uparrow$} \\
      \resizebox{\linewidth}{!}{
\begin{tabular}{lcccccc}
\toprule
Method & WB & CA & UC & AMC & SpI & Avg. \\
\midrule
MSP & $65.5_{\textcolor{gray}{\pm0.6}}$ & $44.9_{\textcolor{gray}{\pm3.1}}$ & $41.3_{\textcolor{gray}{\pm1.2}}$ & $81.2_{\textcolor{gray}{\pm0.5}}$ & $83.8_{\textcolor{gray}{\pm0.2}}$ & $63.3$ \\
Energy & $63.9_{\textcolor{gray}{\pm2.9}}$ & $45.5_{\textcolor{gray}{\pm3.9}}$ & $41.1_{\textcolor{gray}{\pm2.5}}$ & $81.8_{\textcolor{gray}{\pm0.7}}$ & $81.3_{\textcolor{gray}{\pm0.4}}$ & $62.7$ \\
MLS & $64.3_{\textcolor{gray}{\pm2.3}}$ & $45.3_{\textcolor{gray}{\pm3.7}}$ & $41.2_{\textcolor{gray}{\pm2.1}}$ & $82.0_{\textcolor{gray}{\pm0.7}}$ & $82.7_{\textcolor{gray}{\pm0.2}}$ & $63.1$ \\
KLM & $49.7_{\textcolor{gray}{\pm0.7}}$ & $45.0_{\textcolor{gray}{\pm2.5}}$ & $56.4_{\textcolor{gray}{\pm1.2}}$ & $76.1_{\textcolor{gray}{\pm0.6}}$ & $80.1_{\textcolor{gray}{\pm0.3}}$ & $61.5$ \\
GNorm & $76.5_{\textcolor{gray}{\pm0.5}}$ & $45.4_{\textcolor{gray}{\pm1.8}}$ & $50.0_{\textcolor{gray}{\pm0.9}}$ & $75.1_{\textcolor{gray}{\pm0.6}}$ & $85.0_{\textcolor{gray}{\pm0.2}}$ & $66.4$ \\
ReAct & $72.0_{\textcolor{gray}{\pm3.7}}$ & $39.0_{\textcolor{gray}{\pm7.9}}$ & $43.5_{\textcolor{gray}{\pm3.8}}$ & $81.3_{\textcolor{gray}{\pm0.9}}$ & $84.9_{\textcolor{gray}{\pm0.7}}$ & $64.1$ \\
VIM & $88.5_{\textcolor{gray}{\pm1.7}}$ & $48.3_{\textcolor{gray}{\pm3.7}}$ & $63.8_{\textcolor{gray}{\pm1.4}}$ & $79.4_{\textcolor{gray}{\pm0.7}}$ & $80.5_{\textcolor{gray}{\pm0.6}}$ & $72.1$ \\
\bottomrule
\end{tabular}
}
    \end{minipage}
    \hfill
    \begin{minipage}{0.49\textwidth}
      \centering
      {\scriptsize\textbf{FPR@95}$\downarrow$} \\
      \resizebox{\linewidth}{!}{
\begin{tabular}{lcccccc}
\toprule
Method & WB & CA & UC & AMC & SpI & Avg. \\
\midrule
MSP & $87.7_{\textcolor{gray}{\pm0.7}}$ & $98.9_{\textcolor{gray}{\pm0.3}}$ & $95.9_{\textcolor{gray}{\pm0.5}}$ & $81.0_{\textcolor{gray}{\pm0.9}}$ & $69.0_{\textcolor{gray}{\pm1.1}}$ & $86.5$ \\
Energy & $90.4_{\textcolor{gray}{\pm5.6}}$ & $98.2_{\textcolor{gray}{\pm0.9}}$ & $95.2_{\textcolor{gray}{\pm2.2}}$ & $79.2_{\textcolor{gray}{\pm1.2}}$ & $74.4_{\textcolor{gray}{\pm1.0}}$ & $87.5$ \\
MLS & $89.4_{\textcolor{gray}{\pm3.2}}$ & $98.8_{\textcolor{gray}{\pm0.4}}$ & $95.7_{\textcolor{gray}{\pm1.6}}$ & $79.2_{\textcolor{gray}{\pm1.1}}$ & $70.9_{\textcolor{gray}{\pm0.7}}$ & $86.8$ \\
KLM & $89.0_{\textcolor{gray}{\pm0.7}}$ & $98.9_{\textcolor{gray}{\pm0.3}}$ & $95.8_{\textcolor{gray}{\pm0.5}}$ & $78.6_{\textcolor{gray}{\pm0.9}}$ & $74.5_{\textcolor{gray}{\pm1.1}}$ & $87.4$ \\
GNorm & $85.7_{\textcolor{gray}{\pm0.9}}$ & $98.8_{\textcolor{gray}{\pm0.4}}$ & $95.7_{\textcolor{gray}{\pm0.6}}$ & $80.2_{\textcolor{gray}{\pm0.8}}$ & $53.9_{\textcolor{gray}{\pm1.3}}$ & $82.9$ \\
ReAct & $86.9_{\textcolor{gray}{\pm8.7}}$ & $97.2_{\textcolor{gray}{\pm1.7}}$ & $94.6_{\textcolor{gray}{\pm2.9}}$ & $79.9_{\textcolor{gray}{\pm1.1}}$ & $59.0_{\textcolor{gray}{\pm2.2}}$ & $83.5$ \\
VIM & $42.2_{\textcolor{gray}{\pm4.2}}$ & $97.2_{\textcolor{gray}{\pm0.6}}$ & $68.3_{\textcolor{gray}{\pm2.6}}$ & $87.0_{\textcolor{gray}{\pm0.8}}$ & $76.9_{\textcolor{gray}{\pm0.9}}$ & $74.3$ \\
\bottomrule
\end{tabular}
}
    \end{minipage}
    \end{table}

%% file: backbone_tables/DINOv2-S.tex
\begin{table}[t]
    \centering
    \caption{AUROC and FPR@95 performance of all methods using DINOv2-S as the feature backbone.}
    \label{tab:backbone_DINOv2-S_aurocfpr}
    \begin{minipage}{0.49\textwidth}
      \centering
      {\scriptsize\textbf{AUROC}$\uparrow$} \\
      \resizebox{\linewidth}{!}{
\begin{tabular}{lcccccc}
\toprule
Method & WB & CA & UC & AMC & SpI & Avg. \\
\midrule
MSP & $80.1_{\textcolor{gray}{\pm2.0}}$ & $34.0_{\textcolor{gray}{\pm3.3}}$ & $64.3_{\textcolor{gray}{\pm0.5}}$ & $87.6_{\textcolor{gray}{\pm0.4}}$ & $84.9_{\textcolor{gray}{\pm0.4}}$ & $70.2$ \\
Energy & $77.4_{\textcolor{gray}{\pm3.0}}$ & $36.5_{\textcolor{gray}{\pm9.2}}$ & $63.2_{\textcolor{gray}{\pm9.9}}$ & $88.8_{\textcolor{gray}{\pm0.4}}$ & $86.6_{\textcolor{gray}{\pm0.8}}$ & $70.5$ \\
MLS & $78.0_{\textcolor{gray}{\pm2.6}}$ & $35.9_{\textcolor{gray}{\pm8.9}}$ & $63.5_{\textcolor{gray}{\pm9.1}}$ & $88.8_{\textcolor{gray}{\pm0.4}}$ & $86.5_{\textcolor{gray}{\pm0.6}}$ & $70.5$ \\
KLM & $64.7_{\textcolor{gray}{\pm1.4}}$ & $54.1_{\textcolor{gray}{\pm3.6}}$ & $65.2_{\textcolor{gray}{\pm6.3}}$ & $82.8_{\textcolor{gray}{\pm0.7}}$ & $84.4_{\textcolor{gray}{\pm0.3}}$ & $70.2$ \\
GNorm & $82.7_{\textcolor{gray}{\pm1.5}}$ & $32.8_{\textcolor{gray}{\pm2.4}}$ & $63.5_{\textcolor{gray}{\pm0.5}}$ & $83.4_{\textcolor{gray}{\pm0.7}}$ & $\textbf{86.9}_{\textcolor{gray}{\pm0.6}}$ & $69.9$ \\
ReAct & $76.5_{\textcolor{gray}{\pm2.8}}$ & $35.2_{\textcolor{gray}{\pm8.2}}$ & $62.9_{\textcolor{gray}{\pm9.6}}$ & $88.7_{\textcolor{gray}{\pm0.4}}$ & $86.7_{\textcolor{gray}{\pm0.7}}$ & $70.0$ \\
VIM & $97.8_{\textcolor{gray}{\pm0.4}}$ & $39.0_{\textcolor{gray}{\pm8.2}}$ & $83.0_{\textcolor{gray}{\pm7.2}}$ & $86.5_{\textcolor{gray}{\pm0.7}}$ & $84.9_{\textcolor{gray}{\pm0.3}}$ & $78.2$ \\
\bottomrule
\end{tabular}
}
    \end{minipage}
    \hfill
    \begin{minipage}{0.49\textwidth}
      \centering
      {\scriptsize\textbf{FPR@95}$\downarrow$} \\
      \resizebox{\linewidth}{!}{
\begin{tabular}{lcccccc}
\toprule
Method & WB & CA & UC & AMC & SpI & Avg. \\
\midrule
MSP & $80.1_{\textcolor{gray}{\pm2.6}}$ & $99.0_{\textcolor{gray}{\pm0.2}}$ & $91.8_{\textcolor{gray}{\pm1.6}}$ & $71.6_{\textcolor{gray}{\pm1.2}}$ & $63.0_{\textcolor{gray}{\pm2.8}}$ & $81.1$ \\
Energy & $79.0_{\textcolor{gray}{\pm7.5}}$ & $97.9_{\textcolor{gray}{\pm1.5}}$ & $87.5_{\textcolor{gray}{\pm7.9}}$ & $69.3_{\textcolor{gray}{\pm1.7}}$ & $56.0_{\textcolor{gray}{\pm3.7}}$ & $77.9$ \\
MLS & $78.2_{\textcolor{gray}{\pm7.2}}$ & $98.4_{\textcolor{gray}{\pm1.0}}$ & $88.0_{\textcolor{gray}{\pm7.6}}$ & $69.3_{\textcolor{gray}{\pm1.6}}$ & $57.8_{\textcolor{gray}{\pm3.1}}$ & $78.3$ \\
KLM & $82.7_{\textcolor{gray}{\pm2.7}}$ & $99.0_{\textcolor{gray}{\pm0.2}}$ & $93.2_{\textcolor{gray}{\pm3.2}}$ & $63.7_{\textcolor{gray}{\pm1.2}}$ & $64.4_{\textcolor{gray}{\pm2.2}}$ & $80.6$ \\
GNorm & $79.6_{\textcolor{gray}{\pm2.7}}$ & $99.0_{\textcolor{gray}{\pm0.2}}$ & $91.9_{\textcolor{gray}{\pm1.6}}$ & $73.1_{\textcolor{gray}{\pm1.4}}$ & $57.3_{\textcolor{gray}{\pm2.3}}$ & $80.2$ \\
ReAct & $79.8_{\textcolor{gray}{\pm7.1}}$ & $97.8_{\textcolor{gray}{\pm1.5}}$ & $87.7_{\textcolor{gray}{\pm7.5}}$ & $68.9_{\textcolor{gray}{\pm1.7}}$ & $57.2_{\textcolor{gray}{\pm3.4}}$ & $78.3$ \\
VIM & $6.0_{\textcolor{gray}{\pm1.7}}$ & $97.4_{\textcolor{gray}{\pm1.1}}$ & $56.6_{\textcolor{gray}{\pm13.1}}$ & $67.1_{\textcolor{gray}{\pm2.2}}$ & $56.6_{\textcolor{gray}{\pm1.1}}$ & $56.7$ \\
\bottomrule
\end{tabular}
}
    \end{minipage}
    \end{table}

%% file: backbone_tables/ViT-S.tex
\begin{table}[t]
    \centering
    \caption{AUROC and FPR@95 performance of all methods using ViT-S as the feature backbone.}
    \label{tab:backbone_ViT-S_aurocfpr}
    \begin{minipage}{0.49\textwidth}
      \centering
      {\scriptsize\textbf{AUROC}$\uparrow$} \\
      \resizebox{\linewidth}{!}{
\begin{tabular}{lcccccc}
\toprule
Method & WB & CA & UC & AMC & SpI & Avg. \\
\midrule
MSP & $71.5_{\textcolor{gray}{\pm1.9}}$ & $39.2_{\textcolor{gray}{\pm1.2}}$ & $47.8_{\textcolor{gray}{\pm0.4}}$ & $83.2_{\textcolor{gray}{\pm0.2}}$ & $85.2_{\textcolor{gray}{\pm0.3}}$ & $65.4$ \\
Energy & $69.0_{\textcolor{gray}{\pm9.8}}$ & $40.3_{\textcolor{gray}{\pm10.5}}$ & $48.9_{\textcolor{gray}{\pm5.0}}$ & $85.4_{\textcolor{gray}{\pm0.4}}$ & $85.6_{\textcolor{gray}{\pm0.7}}$ & $65.8$ \\
MLS & $69.3_{\textcolor{gray}{\pm9.1}}$ & $39.8_{\textcolor{gray}{\pm9.9}}$ & $48.7_{\textcolor{gray}{\pm4.6}}$ & $85.3_{\textcolor{gray}{\pm0.4}}$ & $85.7_{\textcolor{gray}{\pm0.5}}$ & $65.8$ \\
KLM & $62.6_{\textcolor{gray}{\pm2.5}}$ & $50.6_{\textcolor{gray}{\pm4.7}}$ & $47.2_{\textcolor{gray}{\pm5.6}}$ & $77.9_{\textcolor{gray}{\pm0.5}}$ & $84.4_{\textcolor{gray}{\pm0.6}}$ & $64.5$ \\
GNorm & $76.0_{\textcolor{gray}{\pm1.4}}$ & $38.4_{\textcolor{gray}{\pm1.0}}$ & $47.9_{\textcolor{gray}{\pm0.6}}$ & $83.0_{\textcolor{gray}{\pm0.4}}$ & $\textbf{86.7}_{\textcolor{gray}{\pm0.3}}$ & $66.4$ \\
ReAct & $70.3_{\textcolor{gray}{\pm7.8}}$ & $49.1_{\textcolor{gray}{\pm9.6}}$ & $49.0_{\textcolor{gray}{\pm5.2}}$ & $84.9_{\textcolor{gray}{\pm0.4}}$ & $83.4_{\textcolor{gray}{\pm1.0}}$ & $67.3$ \\
VIM & $84.5_{\textcolor{gray}{\pm7.9}}$ & $44.0_{\textcolor{gray}{\pm9.7}}$ & $67.8_{\textcolor{gray}{\pm6.0}}$ & $85.9_{\textcolor{gray}{\pm0.4}}$ & $85.5_{\textcolor{gray}{\pm0.5}}$ & $73.5$ \\
\bottomrule
\end{tabular}
}
    \end{minipage}
    \hfill
    \begin{minipage}{0.49\textwidth}
      \centering
      {\scriptsize\textbf{FPR@95}$\downarrow$} \\
      \resizebox{\linewidth}{!}{
\begin{tabular}{lcccccc}
\toprule
Method & WB & CA & UC & AMC & SpI & Avg. \\
\midrule
MSP & $88.7_{\textcolor{gray}{\pm1.2}}$ & $99.0_{\textcolor{gray}{\pm0.2}}$ & $96.0_{\textcolor{gray}{\pm0.6}}$ & $81.3_{\textcolor{gray}{\pm0.6}}$ & $67.1_{\textcolor{gray}{\pm1.6}}$ & $86.4$ \\
Energy & $87.9_{\textcolor{gray}{\pm8.4}}$ & $98.3_{\textcolor{gray}{\pm1.5}}$ & $93.6_{\textcolor{gray}{\pm3.7}}$ & $78.2_{\textcolor{gray}{\pm1.2}}$ & $66.5_{\textcolor{gray}{\pm2.1}}$ & $84.9$ \\
MLS & $88.3_{\textcolor{gray}{\pm7.3}}$ & $98.6_{\textcolor{gray}{\pm1.0}}$ & $93.9_{\textcolor{gray}{\pm3.0}}$ & $78.6_{\textcolor{gray}{\pm1.1}}$ & $65.0_{\textcolor{gray}{\pm1.4}}$ & $84.9$ \\
KLM & $89.7_{\textcolor{gray}{\pm2.1}}$ & $99.1_{\textcolor{gray}{\pm0.2}}$ & $96.2_{\textcolor{gray}{\pm2.5}}$ & $78.3_{\textcolor{gray}{\pm0.7}}$ & $65.4_{\textcolor{gray}{\pm2.1}}$ & $85.7$ \\
GNorm & $88.4_{\textcolor{gray}{\pm1.2}}$ & $99.0_{\textcolor{gray}{\pm0.2}}$ & $95.9_{\textcolor{gray}{\pm0.5}}$ & $80.6_{\textcolor{gray}{\pm0.8}}$ & $62.5_{\textcolor{gray}{\pm0.7}}$ & $85.3$ \\
ReAct & $86.4_{\textcolor{gray}{\pm11.2}}$ & $97.0_{\textcolor{gray}{\pm2.9}}$ & $94.1_{\textcolor{gray}{\pm3.0}}$ & $79.6_{\textcolor{gray}{\pm1.1}}$ & $76.7_{\textcolor{gray}{\pm2.3}}$ & $86.8$ \\
VIM & $63.6_{\textcolor{gray}{\pm15.1}}$ & $97.5_{\textcolor{gray}{\pm1.4}}$ & $78.1_{\textcolor{gray}{\pm6.7}}$ & $72.1_{\textcolor{gray}{\pm1.2}}$ & $63.8_{\textcolor{gray}{\pm1.3}}$ & $75.0$ \\
\bottomrule
\end{tabular}
}
    \end{minipage}
    \end{table}

%% file: backbone_tables/Swin-B.tex
\begin{table}[t]
    \centering
    \caption{AUROC and FPR@95 performance of all methods using Swin-B as the feature backbone.}
    \label{tab:backbone_Swin-B_aurocfpr}
    \begin{minipage}{0.49\textwidth}
      \centering
      {\scriptsize\textbf{AUROC}$\uparrow$} \\
      \resizebox{\linewidth}{!}{
\begin{tabular}{lcccccc}
\toprule
Method & WB & CA & UC & AMC & SpI & Avg. \\
\midrule
MSP & $82.8_{\textcolor{gray}{\pm0.9}}$ & $30.9_{\textcolor{gray}{\pm2.0}}$ & $57.2_{\textcolor{gray}{\pm0.4}}$ & $92.7_{\textcolor{gray}{\pm0.2}}$ & $90.6_{\textcolor{gray}{\pm0.4}}$ & $70.8$ \\
Energy & $82.4_{\textcolor{gray}{\pm1.5}}$ & $31.0_{\textcolor{gray}{\pm2.9}}$ & $57.5_{\textcolor{gray}{\pm2.7}}$ & $\textbf{93.6}_{\textcolor{gray}{\pm0.3}}$ & $87.6_{\textcolor{gray}{\pm0.6}}$ & $70.4$ \\
MLS & $82.6_{\textcolor{gray}{\pm1.3}}$ & $30.9_{\textcolor{gray}{\pm2.8}}$ & $57.5_{\textcolor{gray}{\pm2.2}}$ & $92.9_{\textcolor{gray}{\pm0.3}}$ & $89.6_{\textcolor{gray}{\pm0.4}}$ & $70.7$ \\
KLM & $57.3_{\textcolor{gray}{\pm1.7}}$ & $52.8_{\textcolor{gray}{\pm3.5}}$ & $50.5_{\textcolor{gray}{\pm0.5}}$ & $92.5_{\textcolor{gray}{\pm0.3}}$ & $90.1_{\textcolor{gray}{\pm0.3}}$ & $68.6$ \\
GNorm & $63.4_{\textcolor{gray}{\pm1.1}}$ & $27.6_{\textcolor{gray}{\pm1.2}}$ & $52.4_{\textcolor{gray}{\pm0.4}}$ & $84.0_{\textcolor{gray}{\pm0.9}}$ & $56.0_{\textcolor{gray}{\pm0.8}}$ & $56.7$ \\
ReAct & $83.6_{\textcolor{gray}{\pm1.7}}$ & $34.3_{\textcolor{gray}{\pm2.9}}$ & $58.0_{\textcolor{gray}{\pm2.6}}$ & $\textbf{93.6}_{\textcolor{gray}{\pm0.3}}$ & $90.8_{\textcolor{gray}{\pm0.4}}$ & $72.1$ \\
VIM & $90.6_{\textcolor{gray}{\pm0.3}}$ & $39.8_{\textcolor{gray}{\pm3.2}}$ & $76.9_{\textcolor{gray}{\pm1.4}}$ & $88.1_{\textcolor{gray}{\pm0.3}}$ & $92.6_{\textcolor{gray}{\pm0.1}}$ & $77.6$ \\
\bottomrule
\end{tabular}
}
    \end{minipage}
    \hfill
    \begin{minipage}{0.49\textwidth}
      \centering
      {\scriptsize\textbf{FPR@95}$\downarrow$} \\
      \resizebox{\linewidth}{!}{
\begin{tabular}{lcccccc}
\toprule
Method & WB & CA & UC & AMC & SpI & Avg. \\
\midrule
MSP & $76.1_{\textcolor{gray}{\pm2.7}}$ & $99.7_{\textcolor{gray}{\pm0.1}}$ & $93.0_{\textcolor{gray}{\pm0.6}}$ & $51.8_{\textcolor{gray}{\pm2.0}}$ & $38.1_{\textcolor{gray}{\pm2.0}}$ & $71.7$ \\
Energy & $76.3_{\textcolor{gray}{\pm6.3}}$ & $99.4_{\textcolor{gray}{\pm0.3}}$ & $91.9_{\textcolor{gray}{\pm3.2}}$ & $47.3_{\textcolor{gray}{\pm2.3}}$ & $53.5_{\textcolor{gray}{\pm1.8}}$ & $73.7$ \\
MLS & $75.7_{\textcolor{gray}{\pm4.7}}$ & $99.6_{\textcolor{gray}{\pm0.2}}$ & $92.9_{\textcolor{gray}{\pm2.6}}$ & $48.5_{\textcolor{gray}{\pm2.2}}$ & $41.0_{\textcolor{gray}{\pm1.4}}$ & $71.5$ \\
KLM & $79.3_{\textcolor{gray}{\pm2.1}}$ & $99.7_{\textcolor{gray}{\pm0.1}}$ & $92.9_{\textcolor{gray}{\pm0.8}}$ & $\textbf{41.3}_{\textcolor{gray}{\pm1.7}}$ & $49.8_{\textcolor{gray}{\pm5.3}}$ & $72.6$ \\
GNorm & $75.9_{\textcolor{gray}{\pm2.4}}$ & $99.7_{\textcolor{gray}{\pm0.1}}$ & $93.1_{\textcolor{gray}{\pm0.6}}$ & $46.5_{\textcolor{gray}{\pm2.2}}$ & $80.9_{\textcolor{gray}{\pm2.2}}$ & $79.2$ \\
ReAct & $75.9_{\textcolor{gray}{\pm6.4}}$ & $99.0_{\textcolor{gray}{\pm0.4}}$ & $91.8_{\textcolor{gray}{\pm2.6}}$ & $48.0_{\textcolor{gray}{\pm2.2}}$ & $48.3_{\textcolor{gray}{\pm1.4}}$ & $72.6$ \\
VIM & $58.2_{\textcolor{gray}{\pm2.0}}$ & $97.7_{\textcolor{gray}{\pm0.6}}$ & $65.0_{\textcolor{gray}{\pm1.9}}$ & $82.7_{\textcolor{gray}{\pm1.2}}$ & $37.8_{\textcolor{gray}{\pm0.6}}$ & $68.3$ \\
\bottomrule
\end{tabular}
}
    \end{minipage}
    \end{table}

%% file: backbone_tables/DeiT-B.tex
\begin{table}[t]
    \centering
    \caption{AUROC and FPR@95 performance of all methods using DeiT-B as the feature backbone.}
    \label{tab:backbone_DeiT-B_aurocfpr}
    \begin{minipage}{0.49\textwidth}
      \centering
      {\scriptsize\textbf{AUROC}$\uparrow$} \\
      \resizebox{\linewidth}{!}{
\begin{tabular}{lcccccc}
\toprule
Method & WB & CA & UC & AMC & SpI & Avg. \\
\midrule
MSP & $75.6_{\textcolor{gray}{\pm0.9}}$ & $43.4_{\textcolor{gray}{\pm2.2}}$ & $53.3_{\textcolor{gray}{\pm0.1}}$ & $88.5_{\textcolor{gray}{\pm0.4}}$ & $87.8_{\textcolor{gray}{\pm0.4}}$ & $69.7$ \\
Energy & $77.2_{\textcolor{gray}{\pm3.1}}$ & $44.3_{\textcolor{gray}{\pm4.0}}$ & $52.4_{\textcolor{gray}{\pm3.1}}$ & $90.0_{\textcolor{gray}{\pm0.6}}$ & $86.2_{\textcolor{gray}{\pm1.1}}$ & $70.0$ \\
MLS & $76.9_{\textcolor{gray}{\pm2.5}}$ & $44.0_{\textcolor{gray}{\pm3.7}}$ & $52.6_{\textcolor{gray}{\pm2.5}}$ & $89.6_{\textcolor{gray}{\pm0.6}}$ & $87.4_{\textcolor{gray}{\pm0.4}}$ & $70.1$ \\
KLM & $47.0_{\textcolor{gray}{\pm0.6}}$ & $43.2_{\textcolor{gray}{\pm2.0}}$ & $49.6_{\textcolor{gray}{\pm0.3}}$ & $85.8_{\textcolor{gray}{\pm0.4}}$ & $86.6_{\textcolor{gray}{\pm0.4}}$ & $62.4$ \\
GNorm & $67.3_{\textcolor{gray}{\pm0.8}}$ & $43.2_{\textcolor{gray}{\pm1.6}}$ & $53.3_{\textcolor{gray}{\pm0.2}}$ & $76.5_{\textcolor{gray}{\pm1.2}}$ & $70.3_{\textcolor{gray}{\pm2.5}}$ & $62.1$ \\
ReAct & $77.1_{\textcolor{gray}{\pm3.0}}$ & $41.9_{\textcolor{gray}{\pm4.0}}$ & $52.7_{\textcolor{gray}{\pm3.3}}$ & $\textbf{90.2}_{\textcolor{gray}{\pm0.6}}$ & $87.3_{\textcolor{gray}{\pm1.0}}$ & $69.8$ \\
VIM & $82.7_{\textcolor{gray}{\pm1.7}}$ & $46.2_{\textcolor{gray}{\pm3.3}}$ & $65.6_{\textcolor{gray}{\pm2.4}}$ & $82.5_{\textcolor{gray}{\pm0.4}}$ & $87.4_{\textcolor{gray}{\pm0.2}}$ & $72.9$ \\
\bottomrule
\end{tabular}
}
    \end{minipage}
    \hfill
    \begin{minipage}{0.49\textwidth}
      \centering
      {\scriptsize\textbf{FPR@95}$\downarrow$} \\
      \resizebox{\linewidth}{!}{
\begin{tabular}{lcccccc}
\toprule
Method & WB & CA & UC & AMC & SpI & Avg. \\
\midrule
MSP & $86.7_{\textcolor{gray}{\pm0.6}}$ & $98.9_{\textcolor{gray}{\pm0.2}}$ & $94.9_{\textcolor{gray}{\pm0.6}}$ & $58.9_{\textcolor{gray}{\pm1.7}}$ & $62.5_{\textcolor{gray}{\pm2.3}}$ & $80.4$ \\
Energy & $80.3_{\textcolor{gray}{\pm8.6}}$ & $98.2_{\textcolor{gray}{\pm1.0}}$ & $94.3_{\textcolor{gray}{\pm3.3}}$ & $\textbf{51.5}_{\textcolor{gray}{\pm2.3}}$ & $68.6_{\textcolor{gray}{\pm4.8}}$ & $78.6$ \\
MLS & $82.0_{\textcolor{gray}{\pm5.9}}$ & $98.5_{\textcolor{gray}{\pm0.6}}$ & $94.7_{\textcolor{gray}{\pm2.6}}$ & $52.7_{\textcolor{gray}{\pm2.2}}$ & $63.2_{\textcolor{gray}{\pm2.9}}$ & $78.2$ \\
KLM & $87.3_{\textcolor{gray}{\pm0.8}}$ & $98.8_{\textcolor{gray}{\pm0.2}}$ & $94.9_{\textcolor{gray}{\pm0.4}}$ & $61.2_{\textcolor{gray}{\pm1.3}}$ & $60.2_{\textcolor{gray}{\pm4.9}}$ & $80.5$ \\
GNorm & $84.4_{\textcolor{gray}{\pm0.7}}$ & $98.9_{\textcolor{gray}{\pm0.2}}$ & $94.6_{\textcolor{gray}{\pm0.4}}$ & $52.5_{\textcolor{gray}{\pm2.2}}$ & $81.0_{\textcolor{gray}{\pm4.6}}$ & $82.3$ \\
ReAct & $80.6_{\textcolor{gray}{\pm7.9}}$ & $98.4_{\textcolor{gray}{\pm0.9}}$ & $93.5_{\textcolor{gray}{\pm3.7}}$ & $52.7_{\textcolor{gray}{\pm2.2}}$ & $66.2_{\textcolor{gray}{\pm4.8}}$ & $78.3$ \\
VIM & $79.3_{\textcolor{gray}{\pm2.4}}$ & $96.7_{\textcolor{gray}{\pm0.7}}$ & $83.0_{\textcolor{gray}{\pm4.0}}$ & $89.1_{\textcolor{gray}{\pm0.8}}$ & $64.8_{\textcolor{gray}{\pm0.9}}$ & $82.6$ \\
\bottomrule
\end{tabular}
}
    \end{minipage}
    \end{table}

%% file: backbone_tables/ConvNeXt-B.tex
\begin{table}[t]
    \centering
    \caption{AUROC and FPR@95 performance of all methods using ConvNeXt-B as the feature backbone.}
    \label{tab:backbone_ConvNeXt-B_aurocfpr}
    \begin{minipage}{0.49\textwidth}
      \centering
      {\scriptsize\textbf{AUROC}$\uparrow$} \\
      \resizebox{\linewidth}{!}{
\begin{tabular}{lcccccc}
\toprule
Method & WB & CA & UC & AMC & SpI & Avg. \\
\midrule
MSP & $88.1_{\textcolor{gray}{\pm0.7}}$ & $30.4_{\textcolor{gray}{\pm2.5}}$ & $57.5_{\textcolor{gray}{\pm0.4}}$ & $91.1_{\textcolor{gray}{\pm0.2}}$ & $93.1_{\textcolor{gray}{\pm0.2}}$ & $72.0$ \\
Energy & $86.1_{\textcolor{gray}{\pm4.0}}$ & $29.1_{\textcolor{gray}{\pm4.7}}$ & $55.7_{\textcolor{gray}{\pm5.7}}$ & $93.6_{\textcolor{gray}{\pm0.2}}$ & $92.9_{\textcolor{gray}{\pm0.3}}$ & $71.5$ \\
MLS & $86.6_{\textcolor{gray}{\pm3.8}}$ & $28.9_{\textcolor{gray}{\pm4.5}}$ & $56.0_{\textcolor{gray}{\pm5.0}}$ & $91.7_{\textcolor{gray}{\pm0.2}}$ & $93.1_{\textcolor{gray}{\pm0.3}}$ & $71.3$ \\
KLM & $65.4_{\textcolor{gray}{\pm1.8}}$ & $48.2_{\textcolor{gray}{\pm2.8}}$ & $50.0_{\textcolor{gray}{\pm0.4}}$ & $92.6_{\textcolor{gray}{\pm0.2}}$ & $90.4_{\textcolor{gray}{\pm0.4}}$ & $69.3$ \\
GNorm & $71.6_{\textcolor{gray}{\pm1.2}}$ & $47.0_{\textcolor{gray}{\pm1.8}}$ & $46.1_{\textcolor{gray}{\pm0.5}}$ & $85.0_{\textcolor{gray}{\pm0.7}}$ & $89.4_{\textcolor{gray}{\pm0.4}}$ & $67.8$ \\
ReAct & $85.4_{\textcolor{gray}{\pm4.2}}$ & $26.1_{\textcolor{gray}{\pm3.7}}$ & $55.7_{\textcolor{gray}{\pm5.5}}$ & $93.2_{\textcolor{gray}{\pm0.2}}$ & $92.9_{\textcolor{gray}{\pm0.2}}$ & $70.7$ \\
VIM & $97.6_{\textcolor{gray}{\pm0.5}}$ & $32.4_{\textcolor{gray}{\pm4.3}}$ & $82.6_{\textcolor{gray}{\pm3.2}}$ & $\textbf{94.7}_{\textcolor{gray}{\pm0.2}}$ & $92.7_{\textcolor{gray}{\pm0.1}}$ & $80.0$ \\
\bottomrule
\end{tabular}
}
    \end{minipage}
    \hfill
    \begin{minipage}{0.49\textwidth}
      \centering
      {\scriptsize\textbf{FPR@95}$\downarrow$} \\
      \resizebox{\linewidth}{!}{
\begin{tabular}{lcccccc}
\toprule
Method & WB & CA & UC & AMC & SpI & Avg. \\
\midrule
MSP & $59.0_{\textcolor{gray}{\pm3.4}}$ & $99.6_{\textcolor{gray}{\pm0.1}}$ & $93.7_{\textcolor{gray}{\pm1.1}}$ & $55.0_{\textcolor{gray}{\pm1.5}}$ & $31.5_{\textcolor{gray}{\pm1.1}}$ & $67.8$ \\
Energy & $65.1_{\textcolor{gray}{\pm10.2}}$ & $99.4_{\textcolor{gray}{\pm0.6}}$ & $91.1_{\textcolor{gray}{\pm7.3}}$ & $53.3_{\textcolor{gray}{\pm1.9}}$ & $32.3_{\textcolor{gray}{\pm1.8}}$ & $68.2$ \\
MLS & $62.4_{\textcolor{gray}{\pm9.2}}$ & $99.6_{\textcolor{gray}{\pm0.4}}$ & $92.1_{\textcolor{gray}{\pm5.8}}$ & $52.9_{\textcolor{gray}{\pm1.8}}$ & $32.0_{\textcolor{gray}{\pm2.5}}$ & $67.8$ \\
KLM & $59.7_{\textcolor{gray}{\pm3.7}}$ & $99.6_{\textcolor{gray}{\pm0.1}}$ & $93.6_{\textcolor{gray}{\pm0.9}}$ & $41.6_{\textcolor{gray}{\pm1.3}}$ & $39.7_{\textcolor{gray}{\pm5.4}}$ & $66.8$ \\
GNorm & $61.9_{\textcolor{gray}{\pm2.8}}$ & $99.6_{\textcolor{gray}{\pm0.1}}$ & $94.4_{\textcolor{gray}{\pm0.8}}$ & $60.5_{\textcolor{gray}{\pm1.7}}$ & $37.0_{\textcolor{gray}{\pm2.4}}$ & $70.7$ \\
ReAct & $67.2_{\textcolor{gray}{\pm8.4}}$ & $99.5_{\textcolor{gray}{\pm0.4}}$ & $91.5_{\textcolor{gray}{\pm6.7}}$ & $56.3_{\textcolor{gray}{\pm1.8}}$ & $32.4_{\textcolor{gray}{\pm2.3}}$ & $69.4$ \\
VIM & $9.4_{\textcolor{gray}{\pm4.3}}$ & $98.6_{\textcolor{gray}{\pm0.6}}$ & $51.5_{\textcolor{gray}{\pm6.7}}$ & $\textbf{37.5}_{\textcolor{gray}{\pm2.0}}$ & $39.8_{\textcolor{gray}{\pm2.8}}$ & $47.4$ \\
\bottomrule
\end{tabular}
}
    \end{minipage}
    \end{table}

%% file: backbone_tables/BiT-R50x1.tex
\begin{table}[t]
    \centering
    \caption{AUROC and FPR@95 performance of all methods using BiT-R50x1 as the feature backbone.}
    \label{tab:backbone_BiT-R50x1_aurocfpr}
    \begin{minipage}{0.49\textwidth}
      \centering
      {\scriptsize\textbf{AUROC}$\uparrow$} \\
      \resizebox{\linewidth}{!}{
\begin{tabular}{lcccccc}
\toprule
Method & WB & CA & UC & AMC & SpI & Avg. \\
\midrule
MSP & $69.5_{\textcolor{gray}{\pm1.7}}$ & $41.8_{\textcolor{gray}{\pm5.4}}$ & $46.8_{\textcolor{gray}{\pm1.0}}$ & $84.3_{\textcolor{gray}{\pm0.2}}$ & $82.5_{\textcolor{gray}{\pm0.3}}$ & $65.0$ \\
Energy & $66.8_{\textcolor{gray}{\pm2.3}}$ & $43.1_{\textcolor{gray}{\pm9.4}}$ & $44.2_{\textcolor{gray}{\pm7.0}}$ & $\textbf{87.6}_{\textcolor{gray}{\pm0.3}}$ & $81.6_{\textcolor{gray}{\pm0.6}}$ & $64.7$ \\
MLS & $67.0_{\textcolor{gray}{\pm2.1}}$ & $42.9_{\textcolor{gray}{\pm9.3}}$ & $44.4_{\textcolor{gray}{\pm6.3}}$ & $87.4_{\textcolor{gray}{\pm0.3}}$ & $82.2_{\textcolor{gray}{\pm0.5}}$ & $64.8$ \\
KLM & $48.4_{\textcolor{gray}{\pm1.0}}$ & $41.9_{\textcolor{gray}{\pm6.0}}$ & $51.1_{\textcolor{gray}{\pm1.1}}$ & $78.6_{\textcolor{gray}{\pm0.4}}$ & $83.0_{\textcolor{gray}{\pm0.4}}$ & $60.6$ \\
GNorm & $49.7_{\textcolor{gray}{\pm1.5}}$ & $35.7_{\textcolor{gray}{\pm2.8}}$ & $60.6_{\textcolor{gray}{\pm1.6}}$ & $71.3_{\textcolor{gray}{\pm0.6}}$ & $53.9_{\textcolor{gray}{\pm1.5}}$ & $54.2$ \\
ReAct & $57.0_{\textcolor{gray}{\pm2.3}}$ & $36.0_{\textcolor{gray}{\pm8.0}}$ & $46.0_{\textcolor{gray}{\pm7.5}}$ & $81.9_{\textcolor{gray}{\pm0.5}}$ & $81.4_{\textcolor{gray}{\pm1.0}}$ & $60.5$ \\
VIM & $93.8_{\textcolor{gray}{\pm1.6}}$ & $48.0_{\textcolor{gray}{\pm10.2}}$ & $73.3_{\textcolor{gray}{\pm7.5}}$ & $87.5_{\textcolor{gray}{\pm0.3}}$ & $\textbf{86.6}_{\textcolor{gray}{\pm0.2}}$ & $77.8$ \\
\bottomrule
\end{tabular}
}
    \end{minipage}
    \hfill
    \begin{minipage}{0.49\textwidth}
      \centering
      {\scriptsize\textbf{FPR@95}$\downarrow$} \\
      \resizebox{\linewidth}{!}{
\begin{tabular}{lcccccc}
\toprule
Method & WB & CA & UC & AMC & SpI & Avg. \\
\midrule
MSP & $84.7_{\textcolor{gray}{\pm1.4}}$ & $98.5_{\textcolor{gray}{\pm0.6}}$ & $95.2_{\textcolor{gray}{\pm0.6}}$ & $76.0_{\textcolor{gray}{\pm0.6}}$ & $66.7_{\textcolor{gray}{\pm1.9}}$ & $84.2$ \\
Energy & $89.2_{\textcolor{gray}{\pm2.8}}$ & $97.8_{\textcolor{gray}{\pm1.5}}$ & $94.9_{\textcolor{gray}{\pm5.4}}$ & $64.2_{\textcolor{gray}{\pm1.2}}$ & $68.1_{\textcolor{gray}{\pm2.7}}$ & $82.8$ \\
MLS & $88.2_{\textcolor{gray}{\pm2.8}}$ & $98.1_{\textcolor{gray}{\pm1.3}}$ & $95.1_{\textcolor{gray}{\pm4.8}}$ & $65.4_{\textcolor{gray}{\pm1.1}}$ & $66.6_{\textcolor{gray}{\pm2.4}}$ & $82.7$ \\
KLM & $85.2_{\textcolor{gray}{\pm1.2}}$ & $98.4_{\textcolor{gray}{\pm0.6}}$ & $95.1_{\textcolor{gray}{\pm0.5}}$ & $74.8_{\textcolor{gray}{\pm0.7}}$ & $63.8_{\textcolor{gray}{\pm2.8}}$ & $83.5$ \\
GNorm & $85.1_{\textcolor{gray}{\pm1.2}}$ & $98.6_{\textcolor{gray}{\pm0.6}}$ & $95.0_{\textcolor{gray}{\pm0.7}}$ & $70.1_{\textcolor{gray}{\pm0.9}}$ & $89.6_{\textcolor{gray}{\pm1.6}}$ & $87.7$ \\
ReAct & $94.3_{\textcolor{gray}{\pm1.4}}$ & $97.4_{\textcolor{gray}{\pm1.9}}$ & $95.2_{\textcolor{gray}{\pm4.7}}$ & $74.2_{\textcolor{gray}{\pm1.1}}$ & $69.4_{\textcolor{gray}{\pm3.4}}$ & $86.1$ \\
VIM & $27.9_{\textcolor{gray}{\pm5.2}}$ & $97.2_{\textcolor{gray}{\pm1.3}}$ & $61.0_{\textcolor{gray}{\pm12.0}}$ & $72.0_{\textcolor{gray}{\pm1.4}}$ & $\textbf{51.0}_{\textcolor{gray}{\pm0.5}}$ & $61.8$ \\
\bottomrule
\end{tabular}
}
    \end{minipage}
    \end{table}

%% file: backbone_aupr_tables/ResNet-18.tex
\begin{table}[t]
    \centering
    \caption{AUPR-IN and AUPR-OUT performance of all methods using ResNet-18 as the feature backbone.}
    \label{tab:backbone_ResNet-18_aupr}
    \begin{minipage}{0.49\textwidth}
      \centering
      {\scriptsize\textbf{AUPR-IN}$\uparrow$} \\
      \resizebox{\linewidth}{!}{
\begin{tabular}{lcccccc}
\toprule
Method & WB & CA & UC & AMC & SpI & Avg. \\
\midrule
MSP & $62.2_{\textcolor{gray}{\pm0.9}}$ & $23.1_{\textcolor{gray}{\pm2.4}}$ & $28.9_{\textcolor{gray}{\pm0.2}}$ & $80.8_{\textcolor{gray}{\pm0.5}}$ & $47.2_{\textcolor{gray}{\pm0.6}}$ & $48.4$ \\
Energy & $62.7_{\textcolor{gray}{\pm4.3}}$ & $23.6_{\textcolor{gray}{\pm3.7}}$ & $30.2_{\textcolor{gray}{\pm1.8}}$ & $79.8_{\textcolor{gray}{\pm0.6}}$ & $43.3_{\textcolor{gray}{\pm0.5}}$ & $47.9$ \\
MLS & $62.6_{\textcolor{gray}{\pm4.0}}$ & $23.5_{\textcolor{gray}{\pm3.6}}$ & $30.0_{\textcolor{gray}{\pm1.5}}$ & $80.2_{\textcolor{gray}{\pm0.6}}$ & $45.9_{\textcolor{gray}{\pm0.5}}$ & $48.4$ \\
KLM & $34.4_{\textcolor{gray}{\pm0.3}}$ & $16.8_{\textcolor{gray}{\pm0.4}}$ & $40.9_{\textcolor{gray}{\pm0.5}}$ & $70.9_{\textcolor{gray}{\pm0.6}}$ & $41.8_{\textcolor{gray}{\pm1.0}}$ & $41.0$ \\
GNorm & $76.0_{\textcolor{gray}{\pm0.4}}$ & $16.7_{\textcolor{gray}{\pm0.8}}$ & $35.1_{\textcolor{gray}{\pm0.2}}$ & $73.7_{\textcolor{gray}{\pm0.4}}$ & $50.0_{\textcolor{gray}{\pm0.5}}$ & $50.3$ \\
ReAct & $71.6_{\textcolor{gray}{\pm3.9}}$ & $18.1_{\textcolor{gray}{\pm1.6}}$ & $30.7_{\textcolor{gray}{\pm2.2}}$ & $80.4_{\textcolor{gray}{\pm0.6}}$ & $\textbf{48.7}_{\textcolor{gray}{\pm0.8}}$ & $49.9$ \\
VIM & $71.5_{\textcolor{gray}{\pm3.5}}$ & $28.3_{\textcolor{gray}{\pm4.1}}$ & $42.1_{\textcolor{gray}{\pm3.0}}$ & $80.6_{\textcolor{gray}{\pm0.5}}$ & $35.5_{\textcolor{gray}{\pm0.5}}$ & $51.6$ \\
\bottomrule
\end{tabular}
}
    \end{minipage}
    \hfill
    \begin{minipage}{0.49\textwidth}
      \centering
      {\scriptsize\textbf{AUPR-OUT}$\uparrow$} \\
      \resizebox{\linewidth}{!}{
\begin{tabular}{lcccccc}
\toprule
Method & WB & CA & UC & AMC & SpI & Avg. \\
\midrule
MSP & $72.7_{\textcolor{gray}{\pm0.4}}$ & $77.3_{\textcolor{gray}{\pm1.3}}$ & $59.7_{\textcolor{gray}{\pm0.5}}$ & $59.1_{\textcolor{gray}{\pm0.9}}$ & $92.4_{\textcolor{gray}{\pm0.2}}$ & $72.2$ \\
Energy & $74.5_{\textcolor{gray}{\pm5.4}}$ & $78.3_{\textcolor{gray}{\pm2.2}}$ & $63.0_{\textcolor{gray}{\pm5.2}}$ & $57.3_{\textcolor{gray}{\pm1.0}}$ & $92.4_{\textcolor{gray}{\pm0.1}}$ & $73.1$ \\
MLS & $74.3_{\textcolor{gray}{\pm4.2}}$ & $77.7_{\textcolor{gray}{\pm1.8}}$ & $62.2_{\textcolor{gray}{\pm4.0}}$ & $58.0_{\textcolor{gray}{\pm1.0}}$ & $92.4_{\textcolor{gray}{\pm0.1}}$ & $72.9$ \\
KLM & $60.6_{\textcolor{gray}{\pm0.5}}$ & $75.0_{\textcolor{gray}{\pm0.8}}$ & $68.6_{\textcolor{gray}{\pm0.4}}$ & $59.4_{\textcolor{gray}{\pm0.8}}$ & $90.4_{\textcolor{gray}{\pm0.3}}$ & $70.8$ \\
GNorm & $82.5_{\textcolor{gray}{\pm0.3}}$ & $75.8_{\textcolor{gray}{\pm1.3}}$ & $62.9_{\textcolor{gray}{\pm0.5}}$ & $55.3_{\textcolor{gray}{\pm1.0}}$ & $95.2_{\textcolor{gray}{\pm0.1}}$ & $74.3$ \\
ReAct & $79.2_{\textcolor{gray}{\pm5.0}}$ & $75.0_{\textcolor{gray}{\pm1.9}}$ & $63.1_{\textcolor{gray}{\pm5.5}}$ & $57.3_{\textcolor{gray}{\pm1.1}}$ & $94.3_{\textcolor{gray}{\pm0.1}}$ & $73.8$ \\
VIM & $83.3_{\textcolor{gray}{\pm2.5}}$ & $80.9_{\textcolor{gray}{\pm1.6}}$ & $79.4_{\textcolor{gray}{\pm2.8}}$ & $58.3_{\textcolor{gray}{\pm1.0}}$ & $87.3_{\textcolor{gray}{\pm0.1}}$ & $77.8$ \\
\bottomrule
\end{tabular}
}
    \end{minipage}
    \end{table}

%% file: backbone_aupr_tables/ResNet-34.tex
\begin{table}[t]
    \centering
    \caption{AUPR-IN and AUPR-OUT performance of all methods using ResNet-34 as the feature backbone.}
    \label{tab:backbone_ResNet-34_aupr}
    \begin{minipage}{0.49\textwidth}
      \centering
      {\scriptsize\textbf{AUPR-IN}$\uparrow$} \\
      \resizebox{\linewidth}{!}{
\begin{tabular}{lcccccc}
\toprule
Method & WB & CA & UC & AMC & SpI & Avg. \\
\midrule
MSP & $60.1_{\textcolor{gray}{\pm0.9}}$ & $18.6_{\textcolor{gray}{\pm1.1}}$ & $30.7_{\textcolor{gray}{\pm0.2}}$ & $84.0_{\textcolor{gray}{\pm0.3}}$ & $52.9_{\textcolor{gray}{\pm0.4}}$ & $49.3$ \\
Energy & $57.8_{\textcolor{gray}{\pm5.8}}$ & $19.0_{\textcolor{gray}{\pm2.9}}$ & $30.4_{\textcolor{gray}{\pm1.7}}$ & $85.4_{\textcolor{gray}{\pm0.4}}$ & $48.8_{\textcolor{gray}{\pm0.6}}$ & $48.3$ \\
MLS & $57.8_{\textcolor{gray}{\pm5.3}}$ & $18.8_{\textcolor{gray}{\pm2.8}}$ & $30.4_{\textcolor{gray}{\pm1.4}}$ & $84.8_{\textcolor{gray}{\pm0.4}}$ & $51.3_{\textcolor{gray}{\pm0.5}}$ & $48.6$ \\
KLM & $33.9_{\textcolor{gray}{\pm0.4}}$ & $18.3_{\textcolor{gray}{\pm0.4}}$ & $40.1_{\textcolor{gray}{\pm0.3}}$ & $74.3_{\textcolor{gray}{\pm0.6}}$ & $52.2_{\textcolor{gray}{\pm2.9}}$ & $43.8$ \\
GNorm & $69.5_{\textcolor{gray}{\pm0.3}}$ & $14.5_{\textcolor{gray}{\pm0.3}}$ & $36.8_{\textcolor{gray}{\pm0.3}}$ & $78.5_{\textcolor{gray}{\pm0.3}}$ & $51.7_{\textcolor{gray}{\pm0.2}}$ & $50.2$ \\
ReAct & $68.6_{\textcolor{gray}{\pm4.8}}$ & $14.3_{\textcolor{gray}{\pm1.2}}$ & $32.0_{\textcolor{gray}{\pm2.5}}$ & $85.4_{\textcolor{gray}{\pm0.4}}$ & $53.3_{\textcolor{gray}{\pm1.0}}$ & $50.7$ \\
VIM & $74.8_{\textcolor{gray}{\pm4.9}}$ & $24.7_{\textcolor{gray}{\pm4.0}}$ & $42.5_{\textcolor{gray}{\pm2.9}}$ & $83.9_{\textcolor{gray}{\pm0.5}}$ & $43.7_{\textcolor{gray}{\pm0.5}}$ & $53.9$ \\
\bottomrule
\end{tabular}
}
    \end{minipage}
    \hfill
    \begin{minipage}{0.49\textwidth}
      \centering
      {\scriptsize\textbf{AUPR-OUT}$\uparrow$} \\
      \resizebox{\linewidth}{!}{
\begin{tabular}{lcccccc}
\toprule
Method & WB & CA & UC & AMC & SpI & Avg. \\
\midrule
MSP & $70.9_{\textcolor{gray}{\pm0.6}}$ & $74.8_{\textcolor{gray}{\pm0.8}}$ & $60.8_{\textcolor{gray}{\pm0.6}}$ & $64.1_{\textcolor{gray}{\pm0.9}}$ & $94.3_{\textcolor{gray}{\pm0.1}}$ & $73.0$ \\
Energy & $71.4_{\textcolor{gray}{\pm7.5}}$ & $76.2_{\textcolor{gray}{\pm2.5}}$ & $61.7_{\textcolor{gray}{\pm5.3}}$ & $64.9_{\textcolor{gray}{\pm1.1}}$ & $93.8_{\textcolor{gray}{\pm0.1}}$ & $73.6$ \\
MLS & $71.3_{\textcolor{gray}{\pm6.0}}$ & $75.5_{\textcolor{gray}{\pm1.8}}$ & $61.4_{\textcolor{gray}{\pm4.0}}$ & $65.1_{\textcolor{gray}{\pm1.0}}$ & $94.1_{\textcolor{gray}{\pm0.1}}$ & $73.5$ \\
KLM & $58.5_{\textcolor{gray}{\pm0.6}}$ & $76.3_{\textcolor{gray}{\pm0.6}}$ & $67.3_{\textcolor{gray}{\pm0.7}}$ & $62.7_{\textcolor{gray}{\pm0.9}}$ & $92.6_{\textcolor{gray}{\pm0.4}}$ & $71.5$ \\
GNorm & $79.0_{\textcolor{gray}{\pm0.5}}$ & $73.0_{\textcolor{gray}{\pm0.6}}$ & $63.6_{\textcolor{gray}{\pm0.5}}$ & $62.6_{\textcolor{gray}{\pm1.0}}$ & $\textbf{95.8}_{\textcolor{gray}{\pm0.0}}$ & $74.8$ \\
ReAct & $75.5_{\textcolor{gray}{\pm6.7}}$ & $74.1_{\textcolor{gray}{\pm3.9}}$ & $61.3_{\textcolor{gray}{\pm5.4}}$ & $64.4_{\textcolor{gray}{\pm1.1}}$ & $95.2_{\textcolor{gray}{\pm0.2}}$ & $74.1$ \\
VIM & $85.7_{\textcolor{gray}{\pm3.3}}$ & $80.2_{\textcolor{gray}{\pm1.1}}$ & $78.7_{\textcolor{gray}{\pm2.3}}$ & $60.8_{\textcolor{gray}{\pm1.0}}$ & $89.6_{\textcolor{gray}{\pm0.1}}$ & $79.0$ \\
\bottomrule
\end{tabular}
}
    \end{minipage}
    \end{table}

%% file: backbone_aupr_tables/ResNet-50.tex
    \begin{table}[t]
    \centering
    \caption{AUPR-IN and AUPR-OUT performance of all methods using ResNet-50 as the feature backbone.}
    \label{tab:backbone_ResNet-50_aupr}
    \begin{minipage}{0.49\textwidth}
      \centering
      {\scriptsize\textbf{AUPR-IN}$\uparrow$} \\
      \resizebox{\linewidth}{!}{
\begin{tabular}{lcccccc}
\toprule
Method & WB & CA & UC & AMC & SpI & Avg. \\
\midrule
MSP & $52.4_{\textcolor{gray}{\pm1.1}}$ & $23.9_{\textcolor{gray}{\pm0.9}}$ & $28.3_{\textcolor{gray}{\pm0.1}}$ & $86.4_{\textcolor{gray}{\pm0.3}}$ & $58.2_{\textcolor{gray}{\pm0.6}}$ & $49.8$ \\
Energy & $52.1_{\textcolor{gray}{\pm3.4}}$ & $23.4_{\textcolor{gray}{\pm2.6}}$ & $28.2_{\textcolor{gray}{\pm0.7}}$ & $86.7_{\textcolor{gray}{\pm0.4}}$ & $52.9_{\textcolor{gray}{\pm0.9}}$ & $48.7$ \\
MLS & $52.1_{\textcolor{gray}{\pm3.3}}$ & $23.4_{\textcolor{gray}{\pm2.5}}$ & $28.2_{\textcolor{gray}{\pm0.5}}$ & $86.9_{\textcolor{gray}{\pm0.4}}$ & $55.8_{\textcolor{gray}{\pm0.5}}$ & $49.3$ \\
KLM & $39.5_{\textcolor{gray}{\pm0.5}}$ & $17.1_{\textcolor{gray}{\pm0.7}}$ & $41.0_{\textcolor{gray}{\pm0.7}}$ & $76.0_{\textcolor{gray}{\pm0.7}}$ & $52.3_{\textcolor{gray}{\pm1.2}}$ & $45.2$ \\
GNorm & $70.2_{\textcolor{gray}{\pm0.5}}$ & $15.8_{\textcolor{gray}{\pm0.2}}$ & $34.5_{\textcolor{gray}{\pm0.3}}$ & $78.4_{\textcolor{gray}{\pm0.5}}$ & $56.3_{\textcolor{gray}{\pm0.3}}$ & $51.0$ \\
ReAct & $67.8_{\textcolor{gray}{\pm3.2}}$ & $20.2_{\textcolor{gray}{\pm2.3}}$ & $29.7_{\textcolor{gray}{\pm1.1}}$ & $87.0_{\textcolor{gray}{\pm0.4}}$ & $\textbf{58.7}_{\textcolor{gray}{\pm1.5}}$ & $52.7$ \\
VIM & $72.7_{\textcolor{gray}{\pm3.6}}$ & $27.3_{\textcolor{gray}{\pm3.0}}$ & $39.2_{\textcolor{gray}{\pm1.2}}$ & $86.5_{\textcolor{gray}{\pm0.4}}$ & $51.5_{\textcolor{gray}{\pm1.2}}$ & $55.4$ \\
\bottomrule
\end{tabular}
}
    \end{minipage}
    \hfill
    \begin{minipage}{0.49\textwidth}
      \centering
      {\scriptsize\textbf{AUPR-OUT}$\uparrow$} \\
      \resizebox{\linewidth}{!}{
\begin{tabular}{lcccccc}
\toprule
Method & WB & CA & UC & AMC & SpI & Avg. \\
\midrule
MSP & $70.9_{\textcolor{gray}{\pm0.4}}$ & $75.8_{\textcolor{gray}{\pm1.0}}$ & $57.9_{\textcolor{gray}{\pm0.2}}$ & $66.5_{\textcolor{gray}{\pm0.9}}$ & $93.9_{\textcolor{gray}{\pm0.1}}$ & $73.0$ \\
Energy & $70.4_{\textcolor{gray}{\pm2.4}}$ & $75.8_{\textcolor{gray}{\pm1.9}}$ & $58.5_{\textcolor{gray}{\pm3.0}}$ & $66.5_{\textcolor{gray}{\pm1.0}}$ & $93.7_{\textcolor{gray}{\pm0.2}}$ & $73.0$ \\
MLS & $70.8_{\textcolor{gray}{\pm1.8}}$ & $75.6_{\textcolor{gray}{\pm1.7}}$ & $58.1_{\textcolor{gray}{\pm2.0}}$ & $66.9_{\textcolor{gray}{\pm1.0}}$ & $94.0_{\textcolor{gray}{\pm0.1}}$ & $73.1$ \\
KLM & $64.7_{\textcolor{gray}{\pm0.5}}$ & $74.6_{\textcolor{gray}{\pm1.5}}$ & $66.8_{\textcolor{gray}{\pm0.2}}$ & $65.7_{\textcolor{gray}{\pm0.9}}$ & $93.3_{\textcolor{gray}{\pm0.3}}$ & $73.0$ \\
GNorm & $81.5_{\textcolor{gray}{\pm0.5}}$ & $73.7_{\textcolor{gray}{\pm0.9}}$ & $60.7_{\textcolor{gray}{\pm0.3}}$ & $63.1_{\textcolor{gray}{\pm0.9}}$ & $\textbf{96.1}_{\textcolor{gray}{\pm0.1}}$ & $75.0$ \\
ReAct & $76.2_{\textcolor{gray}{\pm4.6}}$ & $77.7_{\textcolor{gray}{\pm3.5}}$ & $60.0_{\textcolor{gray}{\pm3.6}}$ & $66.5_{\textcolor{gray}{\pm1.0}}$ & $95.5_{\textcolor{gray}{\pm0.2}}$ & $75.2$ \\
VIM & $85.9_{\textcolor{gray}{\pm1.6}}$ & $79.5_{\textcolor{gray}{\pm1.3}}$ & $78.7_{\textcolor{gray}{\pm1.0}}$ & $64.2_{\textcolor{gray}{\pm1.0}}$ & $92.4_{\textcolor{gray}{\pm0.3}}$ & $80.1$ \\
\bottomrule
\end{tabular}
}
    \end{minipage}
    \end{table}
    

%% file: backbone_aupr_tables/ResNet-101.tex
\begin{table}[t]
    \centering
    \caption{AUPR-IN and AUPR-OUT performance of all methods using ResNet-101 as the feature backbone.}
    \label{tab:backbone_ResNet-101_aupr}
    \begin{minipage}{0.49\textwidth}
      \centering
      {\scriptsize\textbf{AUPR-IN}$\uparrow$} \\
      \resizebox{\linewidth}{!}{
\begin{tabular}{lcccccc}
\toprule
Method & WB & CA & UC & AMC & SpI & Avg. \\
\midrule
MSP & $55.8_{\textcolor{gray}{\pm0.9}}$ & $23.6_{\textcolor{gray}{\pm2.9}}$ & $29.6_{\textcolor{gray}{\pm0.6}}$ & $88.0_{\textcolor{gray}{\pm0.3}}$ & $57.4_{\textcolor{gray}{\pm0.3}}$ & $50.9$ \\
Energy & $53.9_{\textcolor{gray}{\pm3.2}}$ & $23.9_{\textcolor{gray}{\pm3.4}}$ & $29.6_{\textcolor{gray}{\pm1.1}}$ & $88.0_{\textcolor{gray}{\pm0.4}}$ & $55.0_{\textcolor{gray}{\pm0.7}}$ & $50.1$ \\
MLS & $54.1_{\textcolor{gray}{\pm2.9}}$ & $23.8_{\textcolor{gray}{\pm3.3}}$ & $29.7_{\textcolor{gray}{\pm1.0}}$ & $88.1_{\textcolor{gray}{\pm0.4}}$ & $57.5_{\textcolor{gray}{\pm0.4}}$ & $50.6$ \\
KLM & $38.1_{\textcolor{gray}{\pm0.4}}$ & $18.7_{\textcolor{gray}{\pm1.2}}$ & $40.2_{\textcolor{gray}{\pm0.6}}$ & $78.2_{\textcolor{gray}{\pm0.6}}$ & $50.6_{\textcolor{gray}{\pm0.8}}$ & $45.2$ \\
GNorm & $64.2_{\textcolor{gray}{\pm0.5}}$ & $19.0_{\textcolor{gray}{\pm0.6}}$ & $37.5_{\textcolor{gray}{\pm0.7}}$ & $78.8_{\textcolor{gray}{\pm0.4}}$ & $57.1_{\textcolor{gray}{\pm0.2}}$ & $51.3$ \\
ReAct & $67.2_{\textcolor{gray}{\pm3.0}}$ & $16.7_{\textcolor{gray}{\pm2.6}}$ & $31.4_{\textcolor{gray}{\pm1.8}}$ & $87.6_{\textcolor{gray}{\pm0.6}}$ & $62.8_{\textcolor{gray}{\pm1.9}}$ & $53.1$ \\
VIM & $83.7_{\textcolor{gray}{\pm2.2}}$ & $27.1_{\textcolor{gray}{\pm3.8}}$ & $43.4_{\textcolor{gray}{\pm1.8}}$ & $87.2_{\textcolor{gray}{\pm0.5}}$ & $56.9_{\textcolor{gray}{\pm1.1}}$ & $59.7$ \\
\bottomrule
\end{tabular}
}
    \end{minipage}
    \hfill
    \begin{minipage}{0.49\textwidth}
      \centering
      {\scriptsize\textbf{AUPR-OUT}$\uparrow$} \\
      \resizebox{\linewidth}{!}{
\begin{tabular}{lcccccc}
\toprule
Method & WB & CA & UC & AMC & SpI & Avg. \\
\midrule
MSP & $72.7_{\textcolor{gray}{\pm0.6}}$ & $75.3_{\textcolor{gray}{\pm1.4}}$ & $59.8_{\textcolor{gray}{\pm1.1}}$ & $68.8_{\textcolor{gray}{\pm1.0}}$ & $94.9_{\textcolor{gray}{\pm0.1}}$ & $74.3$ \\
Energy & $70.7_{\textcolor{gray}{\pm3.9}}$ & $76.0_{\textcolor{gray}{\pm2.0}}$ & $60.0_{\textcolor{gray}{\pm2.9}}$ & $69.6_{\textcolor{gray}{\pm1.2}}$ & $93.9_{\textcolor{gray}{\pm0.2}}$ & $74.0$ \\
MLS & $71.4_{\textcolor{gray}{\pm2.5}}$ & $75.6_{\textcolor{gray}{\pm1.7}}$ & $59.8_{\textcolor{gray}{\pm2.2}}$ & $69.9_{\textcolor{gray}{\pm1.2}}$ & $94.5_{\textcolor{gray}{\pm0.1}}$ & $74.2$ \\
KLM & $64.2_{\textcolor{gray}{\pm0.7}}$ & $75.8_{\textcolor{gray}{\pm1.1}}$ & $67.3_{\textcolor{gray}{\pm1.2}}$ & $67.9_{\textcolor{gray}{\pm1.0}}$ & $93.6_{\textcolor{gray}{\pm0.2}}$ & $73.8$ \\
GNorm & $79.8_{\textcolor{gray}{\pm0.6}}$ & $76.1_{\textcolor{gray}{\pm1.1}}$ & $63.0_{\textcolor{gray}{\pm1.0}}$ & $65.6_{\textcolor{gray}{\pm1.1}}$ & $96.0_{\textcolor{gray}{\pm0.1}}$ & $76.1$ \\
ReAct & $75.5_{\textcolor{gray}{\pm5.1}}$ & $75.0_{\textcolor{gray}{\pm4.2}}$ & $61.4_{\textcolor{gray}{\pm3.9}}$ & $69.2_{\textcolor{gray}{\pm1.2}}$ & $95.7_{\textcolor{gray}{\pm0.3}}$ & $75.4$ \\
VIM & $92.3_{\textcolor{gray}{\pm1.2}}$ & $78.4_{\textcolor{gray}{\pm1.6}}$ & $80.0_{\textcolor{gray}{\pm1.0}}$ & $64.8_{\textcolor{gray}{\pm1.1}}$ & $93.5_{\textcolor{gray}{\pm0.2}}$ & $81.8$ \\
\bottomrule
\end{tabular}
}
    \end{minipage}
    \end{table}

%% file: backbone_aupr_tables/DINOv2-S.tex
\begin{table}[t]
    \centering
    \caption{AUPR-IN and AUPR-OUT performance of all methods using DINOv2-S as the feature backbone.}
    \label{tab:backbone_DINOv2-S_aupr}
    \begin{minipage}{0.49\textwidth}
      \centering
      {\scriptsize\textbf{AUPR-IN}$\uparrow$} \\
      \resizebox{\linewidth}{!}{
\begin{tabular}{lcccccc}
\toprule
Method & WB & CA & UC & AMC & SpI & Avg. \\
\midrule
MSP & $75.8_{\textcolor{gray}{\pm2.4}}$ & $14.8_{\textcolor{gray}{\pm1.0}}$ & $54.0_{\textcolor{gray}{\pm0.6}}$ & $92.1_{\textcolor{gray}{\pm0.3}}$ & $60.0_{\textcolor{gray}{\pm1.1}}$ & $59.3$ \\
Energy & $71.7_{\textcolor{gray}{\pm3.3}}$ & $15.9_{\textcolor{gray}{\pm2.8}}$ & $53.1_{\textcolor{gray}{\pm9.3}}$ & $93.4_{\textcolor{gray}{\pm0.2}}$ & $65.6_{\textcolor{gray}{\pm1.5}}$ & $59.9$ \\
MLS & $72.2_{\textcolor{gray}{\pm3.0}}$ & $15.7_{\textcolor{gray}{\pm2.7}}$ & $53.3_{\textcolor{gray}{\pm8.8}}$ & $93.4_{\textcolor{gray}{\pm0.2}}$ & $61.4_{\textcolor{gray}{\pm1.5}}$ & $59.2$ \\
KLM & $64.7_{\textcolor{gray}{\pm1.6}}$ & $23.5_{\textcolor{gray}{\pm1.9}}$ & $47.2_{\textcolor{gray}{\pm6.2}}$ & $84.4_{\textcolor{gray}{\pm0.8}}$ & $60.8_{\textcolor{gray}{\pm1.6}}$ & $56.1$ \\
GNorm & $79.4_{\textcolor{gray}{\pm1.7}}$ & $13.8_{\textcolor{gray}{\pm0.5}}$ & $47.1_{\textcolor{gray}{\pm0.4}}$ & $86.3_{\textcolor{gray}{\pm0.7}}$ & $\textbf{66.0}_{\textcolor{gray}{\pm1.3}}$ & $58.5$ \\
ReAct & $70.7_{\textcolor{gray}{\pm2.9}}$ & $15.2_{\textcolor{gray}{\pm2.2}}$ & $52.9_{\textcolor{gray}{\pm9.2}}$ & $93.4_{\textcolor{gray}{\pm0.2}}$ & $65.6_{\textcolor{gray}{\pm1.3}}$ & $59.6$ \\
VIM & $95.8_{\textcolor{gray}{\pm0.9}}$ & $16.8_{\textcolor{gray}{\pm3.3}}$ & $74.4_{\textcolor{gray}{\pm9.4}}$ & $91.9_{\textcolor{gray}{\pm0.4}}$ & $54.7_{\textcolor{gray}{\pm1.0}}$ & $66.7$ \\
\bottomrule
\end{tabular}
}
    \end{minipage}
    \hfill
    \begin{minipage}{0.49\textwidth}
      \centering
      {\scriptsize\textbf{AUPR-OUT}$\uparrow$} \\
      \resizebox{\linewidth}{!}{
\begin{tabular}{lcccccc}
\toprule
Method & WB & CA & UC & AMC & SpI & Avg. \\
\midrule
MSP & $82.3_{\textcolor{gray}{\pm1.6}}$ & $71.9_{\textcolor{gray}{\pm1.3}}$ & $72.8_{\textcolor{gray}{\pm0.3}}$ & $77.1_{\textcolor{gray}{\pm0.8}}$ & $95.5_{\textcolor{gray}{\pm0.2}}$ & $79.9$ \\
Energy & $81.1_{\textcolor{gray}{\pm3.2}}$ & $73.7_{\textcolor{gray}{\pm4.3}}$ & $73.3_{\textcolor{gray}{\pm8.4}}$ & $78.0_{\textcolor{gray}{\pm1.1}}$ & $\textbf{96.1}_{\textcolor{gray}{\pm0.4}}$ & $80.4$ \\
MLS & $81.7_{\textcolor{gray}{\pm2.8}}$ & $73.1_{\textcolor{gray}{\pm4.0}}$ & $73.3_{\textcolor{gray}{\pm7.8}}$ & $78.2_{\textcolor{gray}{\pm1.0}}$ & $95.9_{\textcolor{gray}{\pm0.3}}$ & $80.4$ \\
KLM & $75.7_{\textcolor{gray}{\pm3.3}}$ & $79.2_{\textcolor{gray}{\pm1.8}}$ & $69.2_{\textcolor{gray}{\pm5.3}}$ & $77.4_{\textcolor{gray}{\pm0.9}}$ & $95.1_{\textcolor{gray}{\pm0.2}}$ & $79.3$ \\
GNorm & $83.8_{\textcolor{gray}{\pm1.4}}$ & $71.5_{\textcolor{gray}{\pm1.1}}$ & $72.7_{\textcolor{gray}{\pm0.3}}$ & $73.6_{\textcolor{gray}{\pm1.1}}$ & $\textbf{96.1}_{\textcolor{gray}{\pm0.2}}$ & $79.5$ \\
ReAct & $80.5_{\textcolor{gray}{\pm3.0}}$ & $73.3_{\textcolor{gray}{\pm4.0}}$ & $73.1_{\textcolor{gray}{\pm8.1}}$ & $78.0_{\textcolor{gray}{\pm1.1}}$ & $96.0_{\textcolor{gray}{\pm0.4}}$ & $80.2$ \\
VIM & $98.5_{\textcolor{gray}{\pm0.3}}$ & $75.1_{\textcolor{gray}{\pm3.6}}$ & $88.9_{\textcolor{gray}{\pm4.8}}$ & $74.7_{\textcolor{gray}{\pm1.2}}$ & $95.9_{\textcolor{gray}{\pm0.1}}$ & $86.6$ \\
\bottomrule
\end{tabular}
}
    \end{minipage}
    \end{table}

%% file: backbone_aupr_tables/ViT-S.tex
\begin{table}[t]
    \centering
    \caption{AUPR-IN and AUPR-OUT performance of all methods using ViT-S as the feature backbone.}
    \label{tab:backbone_ViT-S_aupr}
    \begin{minipage}{0.49\textwidth}
      \centering
      {\scriptsize\textbf{AUPR-IN}$\uparrow$} \\
      \resizebox{\linewidth}{!}{
\begin{tabular}{lcccccc}
\toprule
Method & WB & CA & UC & AMC & SpI & Avg. \\
\midrule
MSP & $67.1_{\textcolor{gray}{\pm2.2}}$ & $16.8_{\textcolor{gray}{\pm0.7}}$ & $35.0_{\textcolor{gray}{\pm0.3}}$ & $89.4_{\textcolor{gray}{\pm0.2}}$ & $63.9_{\textcolor{gray}{\pm1.2}}$ & $54.4$ \\
Energy & $63.2_{\textcolor{gray}{\pm10.1}}$ & $19.0_{\textcolor{gray}{\pm6.7}}$ & $35.5_{\textcolor{gray}{\pm2.8}}$ & $90.6_{\textcolor{gray}{\pm0.3}}$ & $65.6_{\textcolor{gray}{\pm1.7}}$ & $54.8$ \\
MLS & $63.3_{\textcolor{gray}{\pm9.8}}$ & $18.8_{\textcolor{gray}{\pm6.5}}$ & $35.4_{\textcolor{gray}{\pm2.6}}$ & $90.5_{\textcolor{gray}{\pm0.3}}$ & $66.1_{\textcolor{gray}{\pm1.5}}$ & $54.8$ \\
KLM & $53.6_{\textcolor{gray}{\pm2.9}}$ & $21.7_{\textcolor{gray}{\pm2.1}}$ & $31.2_{\textcolor{gray}{\pm3.6}}$ & $79.3_{\textcolor{gray}{\pm0.7}}$ & $60.5_{\textcolor{gray}{\pm2.1}}$ & $49.3$ \\
GNorm & $73.1_{\textcolor{gray}{\pm1.4}}$ & $16.4_{\textcolor{gray}{\pm0.5}}$ & $34.6_{\textcolor{gray}{\pm0.5}}$ & $87.8_{\textcolor{gray}{\pm0.4}}$ & $65.6_{\textcolor{gray}{\pm0.6}}$ & $55.5$ \\
ReAct & $64.0_{\textcolor{gray}{\pm8.1}}$ & $24.4_{\textcolor{gray}{\pm9.0}}$ & $35.5_{\textcolor{gray}{\pm2.9}}$ & $90.8_{\textcolor{gray}{\pm0.2}}$ & $\textbf{65.8}_{\textcolor{gray}{\pm2.0}}$ & $56.1$ \\
VIM & $81.0_{\textcolor{gray}{\pm8.8}}$ & $21.8_{\textcolor{gray}{\pm8.7}}$ & $51.7_{\textcolor{gray}{\pm5.8}}$ & $91.3_{\textcolor{gray}{\pm0.3}}$ & $64.6_{\textcolor{gray}{\pm1.3}}$ & $62.1$ \\
\bottomrule
\end{tabular}%
}
    \end{minipage}
    \end{table}

%% file: backbone_aupr_tables/Swin-B.tex
\begin{table}[t]
    \centering
    \caption{AUPR-IN and AUPR-OUT performance of all methods using Swin-B as the feature backbone.}
    \label{tab:backbone_Swin-B_aupr}
    \begin{minipage}{0.49\textwidth}
      \centering
      {\scriptsize\textbf{AUPR-IN}$\uparrow$} \\
      \resizebox{\linewidth}{!}{
%
}
    \end{minipage}
    \end{table}

%% file: backbone_aupr_tables/DeiT-B.tex
\begin{table}[t]
    \centering
    \caption{AUPR-IN and AUPR-OUT performance of all methods using DeiT-B as the feature backbone.}
    \label{tab:backbone_DeiT-B_aupr}
    \begin{minipage}{0.49\textwidth}
      \centering
      {\scriptsize\textbf{AUPR-IN}$\uparrow$} \\
      \resizebox{\linewidth}{!}{
%
}
    \end{minipage}
    \end{table}

%% file: backbone_aupr_tables/ConvNeXt-B.tex
\begin{table}[t]
    \centering
    \caption{AUPR-IN and AUPR-OUT performance of all methods using ConvNeXt-B as the feature backbone.}
    \label{tab:backbone_ConvNeXt-B_aupr}
    \begin{minipage}{0.49\textwidth}
      \centering
      {\scriptsize\textbf{AUPR-IN}$\uparrow$} \\
      \resizebox{\linewidth}{!}{
%
}
    \end{minipage}
    \end{table}

%% file: backbone_aupr_tables/BiT-R50x1.tex
\begin{table}[t]
    \centering
    \caption{AUPR-IN and AUPR-OUT performance of all methods using BiT-R50x1 as the feature backbone.}
    \label{tab:backbone_BiT-R50x1_aupr}
    \begin{minipage}{0.49\textwidth}
      \centering
      {\scriptsize\textbf{AUPR-IN}$\uparrow$} \\
      \resizebox{\linewidth}{!}{
%
}
    \end{minipage}
    \end{table}

%% file: appendix_tables/kmeans_tables.tex
\begin{table}[t]
\centering
\caption{Comparison of SPROD variants on SP-OOD datasets using AUROC and FPR@95 metrics. All methods utilize a pretrained ResNet-50 backbone. Values are averaged over five runs.}
\label{tab:clustering1}
\begin{minipage}{0.49\textwidth}
  \centering
  {\scriptsize\textbf{AUROC}$\uparrow$} \\
  \resizebox{\linewidth}{!}{
\begin{tabular}{lcccccc}
\toprule
Method & WB & CA & UC & AMC & SpI & Avg. \\
\midrule
SPROD-Default & $\textbf{98.8}_{\textcolor{gray}{\pm0.0}}$ & $\textbf{61.6}_{\textcolor{gray}{\pm0.9}}$ & $\textbf{97.4}_{\textcolor{gray}{\pm0.0}}$ & $82.1_{\textcolor{gray}{\pm0.0}}$ & $85.3_{\textcolor{gray}{\pm0.0}}$ & $\textbf{85.0}$ \\
SPROD-Converged & $98.5_{\textcolor{gray}{\pm0.0}}$ & $59.1_{\textcolor{gray}{\pm0.9}}$ & $97.1_{\textcolor{gray}{\pm0.1}}$ & $\textbf{83.0}_{\textcolor{gray}{\pm0.5}}$ & $85.3_{\textcolor{gray}{\pm0.0}}$ & $84.6$ \\
SPROD-KMeans & $98.3_{\textcolor{gray}{\pm0.1}}$ & $57.3_{\textcolor{gray}{\pm2.2}}$ & $96.8_{\textcolor{gray}{\pm0.1}}$ & $82.8_{\textcolor{gray}{\pm0.5}}$ & $\textbf{85.9}_{\textcolor{gray}{\pm0.0}}$ & $84.2$ \\
\bottomrule
\end{tabular}
}
\end{minipage}
\hfill
\begin{minipage}{0.49\textwidth}
  \centering
  {\scriptsize\textbf{FPR@95}$\downarrow$} \\
  \resizebox{\linewidth}{!}{
\begin{tabular}{lcccccc}
\toprule
Method & WB & CA & UC & AMC & SpI & Avg. \\
\midrule
SPROD-Default & $\textbf{4.7}_{\textcolor{gray}{\pm0.1}}$ & $\textbf{93.7}_{\textcolor{gray}{\pm0.9}}$ & $\textbf{19.0}_{\textcolor{gray}{\pm0.4}}$ & $\textbf{70.1}_{\textcolor{gray}{\pm0.0}}$ & $58.0_{\textcolor{gray}{\pm0.1}}$ & $\textbf{49.1}$ \\
SPROD-Converged & $6.4_{\textcolor{gray}{\pm0.1}}$ & $93.9_{\textcolor{gray}{\pm1.0}}$ & $19.1_{\textcolor{gray}{\pm0.4}}$ & $71.0_{\textcolor{gray}{\pm1.2}}$ & $58.1_{\textcolor{gray}{\pm0.1}}$ & $49.7$ \\
SPROD-KMeans & $6.0_{\textcolor{gray}{\pm0.5}}$ & $\textbf{93.7}_{\textcolor{gray}{\pm1.2}}$ & $19.3_{\textcolor{gray}{\pm0.4}}$ & $71.3_{\textcolor{gray}{\pm1.1}}$ & $\textbf{56.4}_{\textcolor{gray}{\pm0.1}}$ & $49.3$ \\
\bottomrule
\end{tabular}
}
\end{minipage}
\end{table}

\begin{table}[t]
\centering
\caption{Comparison of SPROD variants on SP-OOD datasets using AUPR-IN and AUPR-OUT metrics. All methods utilize a pretrained ResNet-50 backbone. Values are averaged over five runs.}
\label{tab:clustering2}
\begin{minipage}{0.49\textwidth}
  \centering
  {\scriptsize\textbf{AUPR-IN}$\uparrow$} \\
  \resizebox{\linewidth}{!}{
\begin{tabular}{lcccccc}
\toprule
Method & WB & CA & UC & AMC & SpI & Avg. \\
\midrule
SPROD-Default & $\textbf{98.1}_{\textcolor{gray}{\pm0.1}}$ & $\textbf{40.6}_{\textcolor{gray}{\pm1.2}}$ & $\textbf{96.7}_{\textcolor{gray}{\pm0.1}}$ & $\textbf{88.9}_{\textcolor{gray}{\pm0.0}}$ & $58.3_{\textcolor{gray}{\pm0.0}}$ & $\textbf{76.5}$ \\
SPROD-Converged & $97.5_{\textcolor{gray}{\pm0.1}}$ & $36.5_{\textcolor{gray}{\pm1.2}}$ & $96.0_{\textcolor{gray}{\pm0.1}}$ & $88.2_{\textcolor{gray}{\pm0.4}}$ & $55.8_{\textcolor{gray}{\pm0.2}}$ & $74.8$ \\
SPROD-KMeans & $97.6_{\textcolor{gray}{\pm0.1}}$ & $32.1_{\textcolor{gray}{\pm4.2}}$ & $96.0_{\textcolor{gray}{\pm0.1}}$ & $88.4_{\textcolor{gray}{\pm0.4}}$ & $\textbf{58.6}_{\textcolor{gray}{\pm0.3}}$ & $74.5$ \\
\bottomrule
\end{tabular}
}
\end{minipage}
\hfill
\begin{minipage}{0.49\textwidth}
  \centering
  {\scriptsize\textbf{AUPR-OUT}$\uparrow$} \\
  \resizebox{\linewidth}{!}{
\begin{tabular}{lcccccc}
\toprule
Method & WB & CA & UC & AMC & SpI & Avg. \\
\midrule
SPROD-Default & $\textbf{99.2}_{\textcolor{gray}{\pm0.0}}$ & $\textbf{84.3}_{\textcolor{gray}{\pm0.6}}$ & $\textbf{98.1}_{\textcolor{gray}{\pm0.0}}$ & $73.0_{\textcolor{gray}{\pm0.8}}$ & $96.0_{\textcolor{gray}{\pm0.0}}$ & $\textbf{90.1}$ \\
SPROD-Converged & $99.1_{\textcolor{gray}{\pm0.0}}$ & $83.6_{\textcolor{gray}{\pm0.6}}$ & $98.0_{\textcolor{gray}{\pm0.0}}$ & $73.2_{\textcolor{gray}{\pm0.8}}$ & $96.0_{\textcolor{gray}{\pm0.0}}$ & $90.0$ \\
SPROD-KMeans & $99.1_{\textcolor{gray}{\pm0.0}}$ & $83.4_{\textcolor{gray}{\pm0.6}}$ & $98.0_{\textcolor{gray}{\pm0.0}}$ & $\textbf{73.4}_{\textcolor{gray}{\pm0.8}}$ & $\textbf{96.3}_{\textcolor{gray}{\pm0.0}}$ & $90.0$ \\
\bottomrule
\end{tabular}
}
\end{minipage}
\end{table}